\documentclass{article} 
\usepackage{collas2023_conference,times}

\usepackage{hyperref}
\hypersetup{
    colorlinks=true,
    linkcolor=red,
    filecolor=magenta,
    urlcolor=blue,
    citecolor=purple,
    pdfpagemode=FullScreen,
    }
    
\usepackage[american]{babel}

\usepackage{mathtools} 
\usepackage{booktabs} 
\usepackage{tikz} 
\usepackage{wrapfig}

  \usepackage[ruled,vlined,linesnumbered]{algorithm2e}

    \usepackage{etoolbox}
\AtBeginEnvironment{algorithm}{\SetArgSty{textrm}}

\usepackage{xspace}
\usepackage{amsmath}
\usepackage{amssymb}
\usepackage{amsthm}
\usepackage{bm}
\usepackage{cleveref}
\usepackage{accents}

\usepackage{multirow}

\usepackage{caption}
\usepackage{subcaption}

\newcommand{\demosfiggen}[1]{Figure generated by \href{https://github.com/probml/rebayes/blob/main/demos/#1}{#1}}
\newcommand{\collasclffiggen}[1]{Figure generated by \href{https://github.com/probml/rebayes/blob/main/demos/collas/classification/outputs/#1}{#1}}
\newcommand{\collasregfiggen}[1]{Figure generated by \href{https://github.com/probml/rebayes/blob/main/demos/collas/regression/outputs/#1}{#1}}
\newcommand{\showdownfiggen}[1]{Figure generated by \href{https://github.com/probml/rebayes/blob/main/demos/showdown/#1}{#1}}

\newcommand{\banditfiggen}[1]{Figure generated by \href{https://github.com/probml/bandits/blob/main/demos/#1}{#1}}
\newcommand{\miscfiggen}[1]{Figure generated by \href{https://github.com/probml/rebayes/blob/main/demos/misc/#1}{#1}}

\newcommand{\eat}[1]{}

\newcommand{\sizeparam}[1]{\overline{#1}}
\newcommand{\sizeout}[1]{\tilde{#1}}
\newcommand{\discount}[1]{\acute{#1}}

\newcommand{\nparams}{P}
\newcommand{\nout}{C}
\newcommand{\nin}{D}
\newcommand{\memory}{L}
\newcommand{\replay}{B}
\newcommand{\buffer}{\replay}
\newcommand{\memoryout}{\sizeout{L}}

\newcommand{\T}{\intercal}

\newcommand{\data}{\calD}

\newcommand{\orfit}{ORFit\xspace}
\newcommand{\lofi}{LO-FI\xspace}
\newcommand{\LOFI}{\lofi}
\newcommand{\LRVGA}{LRVGA\xspace}
\newcommand{\SGDRB}{SGD-RB\xspace}

\newcommand{\myvec}[1]{\mathbf{#1}}
\newcommand{\myvecsym}[1]{\boldsymbol{#1}}

\newcommand{\lin}{\text{lin}}

\makeatletter
\newcommand{\algorithmfootnote}[2][\footnotesize]{%
  \let\old@algocf@finish\@algocf@finish
  \def\@algocf@finish{\old@algocf@finish
    \leavevmode\rlap{\begin{minipage}{\linewidth}
    #1#2
    \end{minipage}}%
  }%
}
\makeatother

\def\1{\bm{1}}

\makeatletter
\newcommand{\oset}[3][-0.3ex]{%
  \mathrel{\mathop{#3}\limits^{
    \vbox to#1{\kern-4\ex@
    \hbox{$\scriptstyle#2$}\vss}}}}
\makeatother

\newcommand{\vzero}{\myvecsym{0}}
\newcommand{\vone}{\myvecsym{1}}

\newcommand{\veta}{\myvecsym{\eta}}

\newcommand{\vmu}{\myvecsym{\mu}}

\newcommand{\vlambda}{\myvecsym{\lambda}}
\newcommand{\vLambda}{\myvecsym{\Lambda}}

\newcommand{\vpi}{\myvecsym{\pi}}

\newcommand{\vtheta}{\myvecsym{\theta}}

\newcommand{\vSigma}{\myvecsym{\Sigma}}

\newcommand{\vUpsilon}{\myvecsym{\Upsilon}}

\newcommand{\ve}{\myvec{e}}

\newcommand{\vg}{\myvec{g}}

\newcommand{\vp}{\myvec{p}}

\newcommand{\vs}{\myvec{s}}

\newcommand{\vv}{\myvec{v}}
\newcommand{\vw}{\myvec{w}}

\newcommand{\vx}{\myvec{x}}

\newcommand{\vy}{\myvec{y}}

\newcommand{\vz}{\myvec{z}}

\newcommand{\vA}{\myvec{A}}
\newcommand{\vB}{\myvec{B}}
\newcommand{\vC}{\myvec{C}}
\newcommand{\vD}{\myvec{D}}

\newcommand{\vF}{\myvec{F}}
\newcommand{\vG}{\myvec{G}}
\newcommand{\vH}{\myvec{H}}
\newcommand{\vI}{\myvec{I}}

\newcommand{\vK}{\myvec{K}}
\newcommand{\vL}{\myvec{L}}

\newcommand{\vQ}{\myvec{Q}}
\newcommand{\vR}{\myvec{R}}
\newcommand{\vS}{\myvec{S}}

\newcommand{\vU}{\myvec{U}}
\newcommand{\vV}{\myvec{V}}
\newcommand{\vW}{\myvec{W}}

\DeclareMathAlphabet{\mathsfit}{\encodingdefault}{\sfdefault}{m}{sl}
\SetMathAlphabet{\mathsfit}{bold}{\encodingdefault}{\sfdefault}{bx}{n}

\def\sR{{\mathbb{R}}}

\newcommand{\mymathcal}[1]{\mathcal{#1}}

\newcommand{\calD}{{\mymathcal{D}}}

\newcommand{\calL}{\mymathcal{L}}

\newcommand{\calX}{\mymathcal{X}}
\newcommand{\calY}{\mymathcal{Y}}

\newcommand{\gauss}{\mathcal{N}}

\newcommand{\softmax}{\mathrm{softmax}}

\newcommand{\unif}{\mathrm{Unif}}
\newcommand{\Unif}{\unif}

\newcommand{\cat}{\mathrm{Cat}}

\newcommand{\lr}{\varepsilon}

\newcommand{\argmax}{\operatornamewithlimits{argmax}}
\newcommand{\argmin}{\operatornamewithlimits{argmin}}

\newcommand{\real}{\sR}

\newcommand{\trans}{{\mkern-1.5mu\mathsf{T}}}

\newcommand{\chol}{\mathrm{chol}}

\newcommand{\KLpq}[2]{D_{\mathbb{KL}}\left({#1} \mrel{\|} {#2}\right)}

\newcommand{\expect}[1]{\mathbb{E}\left[{#1}\right]} 
\newcommand{\expectQ}[2]{\mathbb{E}_{{#2}}\left[ {#1} \right]} 
\newcommand{\Var}{\mathbb{V}}
\newcommand{\var}[1]{\mathbb{V}\left[ {#1}\right]}

\newcommand{\cov}[1]{\mathrm{Cov}\left[ {#1}\right] }

\newcommand{\diag}{\mathrm{diag}}

\newcommand{\loss}{\calL}

\newcommand{\obsVar}{R}

\newcommand{\be}{\begin{equation}}
\newcommand{\ee}{\end{equation}}

\newcommand{\bea}{\begin{eqnarray}}
\newcommand{\eea}{\end{eqnarray}}
\newcommand{\beaa}{\begin{eqnarray*}}
\newcommand{\eeaa}{\end{eqnarray*}}
\newcommand{\ba}{\begin{align*}}
\newcommand{\ea}{\end{align*}}

\newcommand{\mrel}[1]{\mathrel{#1}}

\def\vtheta{{\bm{\theta}}}

\def\ve{{\bm{e}}}

\def\vg{{\bm{g}}}

\def\vp{{\bm{p}}}

\def\vs{{\bm{s}}}

\def\vv{{\bm{v}}}
\def\vw{{\bm{w}}}
\def\vx{{\bm{x}}}
\def\vy{{\bm{y}}}
\def\vz{{\bm{z}}}

\title{Low-rank extended Kalman filtering for online learning of neural networks from streaming data}

\author{
 Peter G. Chang \\
 U. Chicago
 \And
 Gerardo Durán-Martín \\
 Queen Mary Univ.
 \And
 Alex Shestopaloff  \\
 Queen Mary Univ.
 \And
 Matt Jones \\
 U. Colorado, Boulder 
 \And
 Kevin Murphy\\
 Google DeepMind
}

\collasfinalcopy 

\begin{document}

\maketitle

\begin{abstract}
We propose an efficient online approximate Bayesian inference algorithm for estimating the parameters of a nonlinear function from a potentially non-stationary data stream. 
The method is based on the extended Kalman filter (EKF),
but uses a novel low-rank plus diagonal
decomposition of the posterior precision matrix,
which gives a cost per step
which is linear in the number of model parameters.
In contrast to methods based on stochastic variational inference, our method is fully deterministic,
and does not require step-size tuning.
We show experimentally that this results in much faster (more sample efficient) learning, which
results in more rapid adaptation to changing distributions,
and faster accumulation of reward when used as part of a contextual bandit algorithm.
    \end{abstract}

\section{Introduction}

Suppose we observe a stream of labeled observations,
$\data_t=
\{(\vx_t^n, \vy_t^n) \sim p_t(\vx,\vy): n=1{:}N_t\}$, where $\vx_t^n \in \calX = \real^{\nin}$,
$\vy_t^n \in \calY = \real^{\nout}$,
and $N_t$ is the number of examples at step $t$.
(In this paper, we assume
$N_t=1$, since we are interested in rapid learning
from individual data samples.)
Our goal is to fit a prediction model $\vy_t = h(\vx_t,\vtheta)$ in an online fashion,
 where $\vtheta \in \real^P$ are the parameters of the model.
(We focus on the case where $h$ is a deep neural network (DNN),
although in principle our methods can also be applied to other (differentiable) parametric models.)
In particular, we want to recursively estimate the posterior 
over the parameters
\begin{align}
p(\vtheta|\data_{1:t})
 \propto p(\vy_t|\vx_t,\vtheta) p(\vtheta|\data_{1:t-1})
 \label{eqn:post}
 \end{align}
 without having to store all the past data.
 Here $p(\vtheta|\data_{1:t-1})$ is the posterior belief
 state from the previous step, and 
$p(\vy_t|\vx_t,\vtheta)$ is the likelihood function given by
\begin{align}
    p(\vy_t|\vx_t,\vtheta) &= 
    \begin{cases}
    \gauss(\vy_t|h(\vx_t,\vtheta), \vR_t)
     & \mbox{regression} \\
     \cat(\vy_t|h(\vx_t,\vtheta)) 
     & \mbox{classification} 
    \end{cases}
    \label{eqn:likelihood}
\end{align}
For regression, we assume $h(\vx_t, \vtheta) \in \real^C$
returns the mean of the output, and $\vR_t = \obsVar\vI_{\nout}$ 
is the observation covariance, which we view as a hyper-parameter.
For classification,
$h(\vx_t,\vtheta)$ returns a $\nout$-dimensional vector
of class probabilities, which is the mean parameter
of the categorical distribution.
 
In many problem settings
(e.g., recommender systems  \citep{Huang2015},
robotics \citep{Wolczyk2021,Lesort2020},
and sensor networks \citep{Ditzler2015}),
 the data distribution $p_t(\vx,\vy)$ may change over time
 \citep{Gomes2019}.
Hence  we allow the model parameters $\vtheta_t$ to change
over time,
according to a simple Gaussian dynamics model:\footnote{
We do not assume access to any information
about if and when the distribution shifts
(sometimes called a ``task boundary''),
since such information is not usually available.
Furthermore,  the shifts may be gradual, which makes the concept of task boundary ill-defined.
} %
\begin{align}
p_t(\vtheta_t|\vtheta_{t-1}) &= 
\gauss(\vtheta_t  | \gamma_t \vtheta_{t-1}, \vQ_t).
\label{eqn:dynamics}
\end{align}
where we usually take $\vQ_t = q \vI$ and $\gamma_t=\gamma$,
where  $q \ge 0$
and $0 \leq \gamma \leq  1$.
Using $q>0$ injects some noise at each time step,
and ensures that the model does not lose
``plasticity", so it can continue to adapt to changes
\citep[cf.][]{Kurle2020,Ash2020,Dohare2021},
and using  $\gamma < 1$
ensures the variance of the unconditional stochastic process
does not blow up.
If we set $q=0$ and $\gamma=1$, this corresponds to a deterministic model in 
which the parameters do not change, i.e.,
\begin{align}
p_t(\vtheta_t|\vtheta_{t-1}) &= 
\delta(\vtheta_t  - \vtheta_{t-1})
\label{eq:stationary}
\end{align}
This is a useful special case for when we want to estimate the parameters from 
a stream of data coming from a static distribution.
(In practice we find this approach can also work well for the non-stationary setting.)

Recursively computing \cref{eqn:post}
corresponds to Bayesian inference (filtering)
in a state space model,
where the
dynamics model
in \cref{eqn:dynamics}
is linear Gaussian,
but the observation model
in \cref{eqn:likelihood}
is
non-linear and possibly non-Gaussian.
Many approximate algorithms have been proposed
for this task \citep[see e.g.][]{Sarkka13,pml2Book},
but in this paper, we focus on Gaussian approximations
to the posterior, $q(\vtheta_t|\data_{1:t}) =\gauss(\vtheta_t | \vmu_{t},\vSigma_{t})$,
since they strike a good balance between 
efficiency and expressivity.
In particular, we build on the extended
Kalman filter (EKF), which linearizes the observation model at each step, and then computes a closed form Gaussian update.
The EKF has been used for online training of neural networks
 in many papers
 \citep[see e.g.,][]{Singhal1988,Watanabe1990,Puskorius1991,
 Iiguni1992,Ruck1992,Haykin01}.
 It can be  thought of as an approximate
Bayesian inference method,
or as a natural gradient method
for MAP parameter estimation
 \citep{Ollivier2018},
 which leverages the posterior covariance
 as a preconditioning matrix for fast
 Newton-like updates \citep{Alessandri2007}.
The EKF was extended to exponential family likelihoods
in \citep{Ollivier2018,Tronarp2018},
which is necessary when fitting classification models.

The main drawback of the EKF is that it takes $O(\nparams^3)$ time per step,
where $\nparams = |\vtheta_t|$ is the number of parameters in the hidden  state vector, 
because we need to invert the posterior covariance matrix.
It is possible to derive diagonal approximations to the posterior covariance or precision,
by either minimizing 
$\KLpq{p(\vtheta_t|\data_{1:t})}{q(\vtheta_t)}$
or
$\KLpq{q(\vtheta_t)}{p(\vtheta_t|\data_{1:t})}$,
as discussed in \citep{Puskorius1991,Chang2022,Jones2023multiscale}.
These methods take $O(\nparams)$ time per step,
but can be much less statistically efficient
than full-covariance methods, since they ignore
joint uncertainty between the parameters.
This makes the method slower to learn,
and slower to adapt to changes in the data distribution,
as we show in \cref{sec:results}.

In this paper, we propose an efficient and deterministic method
to recursively
minimize $\KLpq{\gauss(\vtheta_t|\vmu_t, \vSigma_t)}{p(\vtheta_t|\data_{1:t})}$,
where we assume that the precision matrix is 
 diagonal plus low-rank, 
$\vSigma_t^{-1} = \vUpsilon_t + \vW_t\vW_t^\T$, where $\vUpsilon_t$ is diagonal and $\vW_t\in\real^{\nparams \times \memory}$ for some memory limit $\memory$.
The key insight
is that, if we linearize the observation model at each step, as in the EKF, we can use the resulting gradient vector or Jacobian
as ``pseudo-observation(s)" that we append to $\vW_{t-1}$, and then we can perform an efficient online SVD approximation to obtain $\vW_t$.
We therefore call our method \lofi,
which is short for low-rank extended Kalman filter.
Our code is available at https://github.com/probml/rebayes.

We use the posterior approximation $p(\vtheta_t|\data_{1:t})$
in two ways. First, under Bayesian updating the covariance matrix $\vSigma_t$
acts as a preconditioning matrix to yield a  deterministic second-order Newton-like update for the posterior mean (MAP estimate).
This update  does not have any step-size hyperparameters, in contrast to SGD.
Second, the posterior uncertainty in the parameters
can be propagated into the uncertainty of the predictive
distribution for observations,
which is crucial for online decision-making tasks,
such as
active learning \citep{Holzmuller2022},
Bayesian optimization \citep{Garnett2023},
contextual bandits \citep{Duran-Martin2022},
and reinforcement learning \citep{Khetarpal2022,Wang2021}.

In summary, our main contribution 
is a novel algorithm for efficiently (and deterministically) recursively updating a diagonal plus low-rank (DLR) approximation to the precision matrix of a Gaussian posterior for a special kind of  state space model, namely an SSM with an arbitrary non-linear (and possibly non-Gaussian) observation model, but with a simple linear Gaussian dynamics. This model family is ideally suited to online parameter learning for DNNs in potentially non-stationary environments (but the restricted form of the dynamics model excludes some other applications of SSMs). 
We show experimentally that our approach works better (in terms of accuracy for a given compute budget) than 
a variety of baseline algorithms --- including
online gradient descent,
online Laplace,
diagonal approximations to the EKF,
and a stochastic DLR VI method called L-RVGA
---  on a variety of stationary and non-stationary classification
and regression problems,
as well as a simple contextual bandit problem.


\section{Related work}
\label{sec:related}

Since exact Bayesian inference is intractable in our model family,
it is natural to compute an approximate posterior at
step $t$ using recursive variational inference (VI),
in which the prior for step $t$ is the approximate
posterior from step $t-1$
 \citep{Opper98,Broderick2013}.
 That is, at each step we minimize
 the ELBO (evidence lower bound),
 which is equal (up to a constant)
 to the reverse KL,
 given by
 \begin{align}
 \loss(\vmu_t, \vSigma_t) = 
  \KLpq {\gauss(\vtheta_t|\vmu_t,\vSigma_t)}
  {Z_t p(\vy_t|\vx_t,\vtheta_t) q_{t|t-1}(\vtheta_t|\data_{1:t-1})}
  \label{eqn:ELBO}
  \end{align}
 where $Z_t$ is a normalization constant
 and $q_t=\gauss(\vtheta_t|\vmu_t,\vSigma_t)$ 
 is the variational posterior which results
 from minimizing this expression.
The main challenge is how to efficiently optimize this objective.

One common approach is to assume the variational family consists of a diagonal Gaussian.
 By linearizing the likelihood, we can 
 solve the VI objective 
 in closed form, as shown in \citep{Chang2022};
 this is called the ``variational diagonal EKF" (VD-EKF).
They also propose a diagonal approximation
which minimizes 
the forwards KL,
$\KLpq{p(\vtheta_t|\data_{1:t})}{q(\vtheta_t)}$,
and  show that this is equivalent
to the ``fully decoupled EKF" (FD-EKF)
method of
 \citep{Puskorius1991}.
 Both of these methods are fully deterministic,
which avoids
the high variance that often plagues
stochastic VI methods \citep{Wu2019VB,Hausmann2020}.

It is also possible to derive diagonal approximations without linearizing the observation model.
In \citep{Kurle2020,Zeno2018} 
they propose  a diagonal approximation
to minimize the reverse KL,
$\KLpq{q(\vtheta_t)}{p(\vtheta_t|\data_{1:t})}$; this requires a
Monte Carlo approximation to the ELBO.
In \citep{Ghosh2016,Wagner2022},
they propose  a diagonal approximation
to minimize the forwards KL,
$\KLpq{p(\vtheta_t|\data_{1:t})}{q(\vtheta_t)}$; this  requires
approximating the first and second moments of the hidden
units at every layer of the model using numerical
integration.


 \citep{Farquhar2020} claims that, if one makes the model deep enough,
 one can get good performance using a diagonal approximation;
 however, this has not been our experience. 
 This motivates the need to go beyond a diagonal approximation.

One approach is to combine diagonal Gaussian approximations
with memory buffers,
such as the 
variational continual learning methd of
   \citep{Nguyen2018continual}
   and other works
 (see e.g.,  \citep{Kurle2020,Khan2021}).
However, we seek to find a richer approximation to the posterior that does not rely on memory buffers, which can be problematic in the non-stationary setting. 

\citep{Zeno2021} proposes the FOO-VB method,
which uses a Kronecker block structured
approximation to the posterior covariance.
However, this method requires 2 SVD decompositions
of the Kronecker factors for every layer of the model, in addition to  a large
number of Monte Carlo samples,
at each time step.
In \citep{Ong2018} they compute
a diagonal plus low-rank (DLR)
approximation to the posterior
covariance matrix using stochastic gradient
applied to the ELBO.
In \citep{Tomczak2020} they develop a version of the local reparameterization trick for the DLR posterior covariance,
to reduce the variance of the stochastic
gradient estimate.

In this paper we use a diagonal plus low-rank (DLR)
approximation to the posterior precision.
The same form of approximation has been used in several
prior papers.
In \citep{Mishkin2018}
they propose a technique
called ``SLANG'' 
(stochastic low-rank approximate natural-gradient),
which uses a stochastic
estimate of the natural gradient of the ELBO to update the posterior precision,
combined with a randomized eigenvalue solver to compute a DLR approximation.
Their NGD approximation enables the variational updates to be calculated solely from the loss gradients, whereas our approach requires the network Jacobian.
On the other hand, our EKF approach allows the posterior precision and the DLR approximation to be efficiently computed in closed form.

In \citep{LRVGA},
they propose a technique
called ``L-RVGA'' (low-rank recursive variational Gaussian approximation), 
which uses stochastic EM to optimize the ELBO
using  a DLR
approximation to the posterior precision.
Their method is a one-pass online method,
like ours, and also avoids the need to tune
the learning rate.
However, it is much slower, since it involves generating
multiple samples from the posterior
and multiple iterations of the EM algorithm
(see 
\cref{fig:energy-running-time}
for an experimental comparison of running time).

The GGT method of \citep{GGT} also computes a DLR approximation to the posterior precision, which they use as a preconditioner for computing the MAP estimate.
However, they bound the rank by simply using the most recent $\memory$ observations, whereas \lofi uses SVD to combine the past data in a more efficient way.

The \orfit method of \citep{ORFit}
is also an online low-rank MAP estimation method.
They 
use orthogonal projection to efficiently
compute a low rank representation 
of the precision at each step.
However, it is restricted to regression problems
with 1d, noiseless outputs
(i.e., they assume the likelihood has the
degenerate form $p(y_t|\vx_t,\vtheta_t) =\gauss(h(\vx_t,\vtheta_t), 0)$.)

The online Laplace method of \citep{Ritter2018online,Daxberger2021laplace} 
also computes a Gaussian approximation to the posterior,
but makes different approximations.
In particular, for ``task" $t$, it computes
the MAP estimate
$\vtheta_t = \argmax_{\vtheta} \log p(\data_t|\vtheta)
+ \log \gauss(\vtheta|\vmu_{t-1}, \vSigma_{t-1})$,
where $\vSigma_{t-1}=\vLambda_{t-1}^{-1}$ is the approximate
posterior covariance from the previous task.
(This optimization problem is solved using SGD applied to a replay buffer.)
This precision matrix is usually approximated as a block diagonal matrix, with one block per layer,
and the terms within each block may be additionally
approximated by a Kronecker product form,
as in KFAC \citep{Martens2015}.
By contrast, \lofi computes a posterior,
not just a point estimate, and approximates the
precision as diagonal plus low rank.
In the appendix, we show experimentally
that \lofi outperforms online Laplace in terms of NLPD on
various classification and regression tasks.

It is possible to go beyond Gaussian approximations
by using particle filtering 
(see e.g., \citep{Yang2023}).
However, we focus on faster 
deterministic inference methods,
since speed is important for many real time online decision making tasks
\citep{Ghunaim2023}.

There are many papers 
on continual learning,
which is related to online learning.
However the CL literature usually assumes the
task boundaries, corresponding to times when
the distribution shifts,
are given to the learner
(see e.g., \citep{Delange2021,DeLange2021iccv,Wang2022cvpr,Mai2022,Mundt2023,Wang2023CL}.)
By contrast, we are interested
in  the continual learning setting
where the distribution may change at unknown times,
in a continuous or discontinuous manner
(c.f., \citep{Gama2013});
this is sometimes called the
``task agnostic'' or ``streaming'' setting.
Furthermore,
our goal is accurate forecasting
of the future (which can be approximated
by our estimate of the ``current'' distribution),
so we are less concerned with performance
on ``past'' distributions that the agent may not encounter again;
thus ``catastrophic forgetting''
(see e.g., \citep{Parisi2019})
is not a focus of this work
(c.f., \citep{Dohare2021}).

\section{Methods}
\label{sec:methods}
\label{sec:lofi}

In \lofi,
we approximate the belief state by a Gaussian,
$p(\vtheta_t|\data_{1:t})=\gauss(\vmu_t,\vSigma_t)$,
where the posterior precision 
is diagonal plus low rank,
i.e., it has the form
$\vSigma_t^{-1} = \vUpsilon_t + \vW_t \vW_t^\trans$,
where $\vUpsilon_t$ is diagonal and $\vW_t$ is a $\nparams \times \memory$ matrix.
We denote this class of models by
$\text{DLR}(\memory)$,
where $\memory$ is the rank.
Below we show how to efficiently update this belief state
in a recursive (online) fashion.
This has two main steps --- 
predict
(see \cref{algo:LOFI-predict})
and update (see \cref{algo:LOFI-update}) ---
which are called repeatedly,
as shown in \cref{algo:LOFI}.
The predict step takes $O(\nparams \memory^2 + \memory^3)$ time,
and the update step takes $O(\nparams (\memory + \nout)^2)$ time,
where $\nout$ is the number of outputs.

\begin{algorithm}
def $\text{lofi}(\vmu_0, \vUpsilon_0, \vx_{1:T},
\vy_{1:T}, \gamma_{1:T}, q_{1:T}, \memory, h)$ \\
$\vW_0 = \vzero$\\
\ForEach{$t=1:T$}{
 $( \vmu_{t|t-1}, \vUpsilon_{t|t-1}, 
\vW_{t|t-1}, \hat{\vy}_{t})
= \text{predict}(
\vmu_{t-1}, \vUpsilon_{t-1},  \vW_{t-1}, \vx_t, \gamma_t, q_t, h)$ 
\\
 $(\vmu_t, \vUpsilon_t, \vW_t)
 = \text{update}(
\vmu_{t|t-1}, \vUpsilon_{t|t-1}, \vW_{t|t-1},
\vx_t, \vy_t, \hat{\vy}_{t}, h, \memory)$
\\
$\text{callback}(\hat{\vy}_{t}, \vy_t)$ 
}
\caption{LOFI main loop.}
\label{algo:LOFI}
\end{algorithm}

\subsection{Predict step}
\label{sec:predict-step}

\begin{algorithm}
def $\text{predict}(
\vmu_{t-1}, \vUpsilon_{t-1},  \vW_{t-1}, \vx_t, \gamma_t, q_t, h)$: 
\\
$\vmu_{t|t-1} = \gamma_t \vmu_{t-1}$
// Predict the mean of the next state
\\
$\vUpsilon_{t|t-1} = 
    \left(\gamma_t^{2} \vUpsilon_{t-1}^{-1}
    +q_t \vI_\nparams\right)^{-1}$
    // Predict the diagonal precision 
\\
$\vC_t = 
\left(
            \vI_{\memory}
            + q_t \vW_{t-1}^{\trans}  \vUpsilon_{t|t-1} \vUpsilon_{t-1}^{-1} \vW_{t-1}
        \right)^{-1}$
    \\
$\vW_{t|t-1} =
    \gamma_t  \vUpsilon_{t|t-1} \vUpsilon_{t-1}^{-1} \vW_{t-1} 
    {\rm chol}(\vC_t)$
    // Predict the low-rank precision
\\
$\hat{\vy}_t =
h\left(\vx_t, \vmu_{t|t-1} \right)$ 
// Predict the mean of the output 
\\
Return $( \vmu_{t|t-1}, \vUpsilon_{t|t-1},  \vW_{t|t-1}, \hat{\vy}_t)$ 
\caption{\lofi predict step.}
\label{algo:LOFI-predict}
\end{algorithm}

In the predict step, we go from the previous posterior,
$p(\vtheta_{t-1}|\data_{1:t-1})
= \gauss(\vtheta_{t-1}|\vmu_{t-1},\vSigma_{t-1})$,
to the one-step-ahead predictive distribution,
$p(\vtheta_{t}|\data_{1:t-1})
= \gauss(\vtheta_{t}|\vmu_{t|t-1}, \vSigma_{t|t-1})$.
To compute this predictive distribution, 
 we apply the dynamics in \cref{eqn:dynamics} with $\vQ_t=q_t\vI$ to get
    $\vmu_{t|t-1} = \gamma_t \vmu_{t-1}$
    and 
    $\vSigma_{t|t-1}  =\gamma_t^{2} \vSigma_{t-1} + q_t\vI_\nparams$.
However, this recursion is in terms of the covariance matrix,
but we need the corresponding result for a DLR precision matrix
in order to be computationally efficient.
In \cref{appx:predict-step}
we show how to use the matrix inversion lemma to efficiently
 compute  $\vSigma^{-1}_{t|t-1}=
\vUpsilon_{t|t-1} + \vW_{t|t-1} \vW_{t|t-1}^\trans$.
The result is
shown in the pseudocode in 
\cref{algo:LOFI-predict},
where 
$\vA={\rm chol}(\vB)$ denotes Cholesky decomposition
(i.e., $\vA\vA^\trans=\vB$).
The cost of computing
$\vUpsilon_{t|t-1}$ is $O(\nparams)$ since it is diagonal.
The cost of computing
$\vW_{t|t-1}$
is $O(\nparams\memory^2 + \memory^3)$.
If we use a full-rank approximation, $\memory = \nparams$,
we recover the standard EKF predict step.

\subsection{Update step}
\label{sec:update-step}

\begin{algorithm}
def $\text{update}(
\vmu_{t|t-1}, \vUpsilon_{t|t-1}, \vW_{t|t-1},
\vx_t, \vy_t, \hat{\vy}_t, h, \memory)$: 
\\
$\vR_t = h_V(\vx_t, \vmu_{t|t-1})$
// Covariance of predicted output  
\\
$\vL_t = \text{chol}(\vR_t)$ \\
$\vA_t = \vL_t^{-1}$ \\
$\vH_t = \text{jac}(h(\vx_t,\cdot))(\vmu_{t|t-1})$
// Jacobian of observation model 
\\
$\sizeout{\vW}_{t} = \left[\begin{array}{cc}
   \vW_{t|t-1}
    & \vH_{t}^{\trans} \vA_{t}^\trans
    \end{array}\right]$
    // Expand low-rank with new observation
\\
$\vG_t = \left(\vI_{\memoryout}+
\sizeout{\vW}_{t}^{\trans}\vUpsilon_{t|t-1}^{-1}
\sizeout{\vW}_{t}\right)^{-1}
$
\\
$\vC_t = \vH_{t}^{\trans} \vA_t^\trans \vA_t$
\\
$\vK_t = \vUpsilon_{t|t-1}^{-1} \vC_t
-\vUpsilon_{t|t-1}^{-1} \sizeout{\vW}_{t} \vG_t
\sizeout{\vW}_{t}^{\trans} \vUpsilon_{t|t-1}^{-1}
\vC_t$  // Kalman gain matrix
\\
$\vmu_t = \vmu_{t|t-1} + \vK_t(\vy_t - \hat{\vy}_t)$ 
// Mean update 
\\
$(\sizeout{\vLambda}_t,\sizeout{\vU}_t) = {\rm SVD}(\sizeout{\vW}_t)$ 
// Take SVD of the expanded low-rank \\
$(\vLambda_{t}, \vU_{t}) = 
    \left(\sizeout{\vLambda}_t, \sizeout{\vU}_t\right)[:,1{:}\memory]$
    // Keep top $L$ most important terms
\\
$\vW_t = \vU_t \vLambda_t$
// New low-rank approximation \\
$(\vLambda_t^\times, \vU_t^\times) = 
    \left(\sizeout{\vLambda}_t, \sizeout{\vU}_t\right)[:,(\memory+1){:}\memoryout] $
    // Extract remaining least important terms 
\\
$\vW_t^{\times} = \vU_t^{\times} \vLambda_t^{\times}$
// The low-rank part that is dropped \\
$\vUpsilon_{t} =
    \vUpsilon_{t|t-1} +
    \diag\left(\vW_t^{\times} (\vW_t^{\times})^{\trans} \right)$
    // Update diagonal to capture variance due to dropped terms
\\
Return $(\vmu_t, \vUpsilon_t, \vW_t)$ 
\caption{\lofi update step.
}
\label{algo:LOFI-update}
\end{algorithm}

In the update step, we go from the
prior predictive distribution,
$p(\vtheta_{t}|\data_{1:t-1})
= \gauss(\vtheta_{t}|\vmu_{t|t-1},\vSigma_{t|t-1})$,
to the posterior distribution,
$p(\vtheta_{t}|\data_{1:t})
= \gauss(\vtheta_{t}|\vmu_{t}, \vSigma_{t})$.
Unlike the predict step, this cannot be computed exactly.
Instead we will compute an approximate posterior $q_t$
by minimizing the  KL objective
in   \cref{eqn:ELBO}.
One can show
\citep[see e.g.,][]{Opper09,Kurle2020,RVGA}
that the optimum must satisfy the following fixed-point equations:
\begin{align}
    \vmu_t &= \vmu_{t|t-1} + \vSigma_{t-1} \nabla_{\vmu_t} 
    \expectQ{\log p(\vy_t|\vtheta_t)}{q_t} 
     = \vmu_{t|t-1} + \vSigma_{t-1} 
     \expectQ{\nabla_{\vtheta_t} \log p(\vy_t|\vtheta_t)}{q_t} 
     \\
        \vSigma_t^{-1} &= \vSigma_{t|t-1}^{-1} 
        -2  \nabla_{\vSigma_t} \expectQ{\log p(\vy_t|\vtheta_t)}{q_t} 
    = \vSigma_{t|t-1}^{-1} 
        -\expectQ{\nabla_{\vtheta_t}^2 \log p(\vy_t|\vtheta_t)}{q_t} 
\end{align}
Note that this is an implicit equation, since $q_t$ occurs
on the left and right hand sides.
A common approach to solving this optimization problem
(e.g., used in \citep{Mishkin2018,Kurle2020, RVGA})
is to approximate 
 the expectation  with  samples
from the prior predictive, $q_{t|t-1}$.
In addition, it is common to approximate
the Hessian matrix with the generalized Gauss Newton (GGN) matrix, which is derived from the Jacobian,
as we explain below.
In this paper we  replace the Monte Carlo expectations with analytic methods, by leveraging the same GGN approximation.
We then generalize 
 to the low-rank setting to make the method efficient.

In more detail,
we compute a linear-Gaussian approximation to the likelihood function,
after which the KL optimization problem can be solved exactly
by performing conjugate Bayesian updating.
To approximate the likelihood, we first 
linearize the observation model
about the prior predictive mean:
\begin{align}
    \hat{h}_t(\vtheta_t) =
    h(\vx_t,\vmu_{t|t-1}) + \vH_t(\vtheta_t-\vmu_{t|t-1})
    \label{eq:linear-obs-model}
\end{align}
where $\vH_t$ is the
$\nout \times \nparams$
Jacobian of $h(\vx_t,\cdot)$ evaluated at $\vmu_{t|t-1}$.
To handle non-Gaussian outputs,
we follow \cite{Ollivier2018} and \cite{Tronarp2018},
and approximate the output distribution using 
a Gaussian, whose  conditional moments are given by
\begin{align}
    \hat{\vy}_t &= \expect{\vy_t|\vx_t, \vtheta_t=\vmu_{t|t-1}} 
    = h(\vx_t, \vmu_{t|t-1}) 
    \label{eqn:pred-obs-mean}
    \\
    \vR_t &=  \cov{\vy_t|\vx_t, \vtheta_t=\vmu_{t|t-1}}
     = h_V(\vx_t, \vmu_{t|t-1}) 
         =\begin{cases}
      \obsVar_t\,\vI_{\nout} & \mbox{regression} \\
      \diag(\hat{\vy}_t) - \hat{\vy}_t \hat{\vy}_t^\trans & \mbox{classification}
\end{cases}
 \label{eqn:pred-obs-cov}
\end{align}
where $\hat{\vy}_t$ is a vector of $\nout$ probabilities
in the case of classification.\footnote{
In the classification case, $\vR_t$ has rank $\nout-1$,
due to the sum-to-one constraint on $\hat{\vy}_t$.
To avoid numerical problems when computing $\vR_t^{-1}$, we can either drop one of the dimensions,
or we can use a pseudoinverse.
The pseudoinverse works because the kernel of $\vR_t$ is contained in the kernel of $\vH_t^\trans$.
}

Under the above assumptions,  we can use the standard EKF update equations
\citep[see e.g.,][]{Sarkka13}.
In \cref{appx:update} we extend these equations
to the case where
the precision matrix is DLR;
this forms the core of our \LOFI method.
The basic idea is to compute the exact update
to get $\vSigma_{t}^{*-1} = \vUpsilon_t + \tilde{\vW}_t \tilde{\vW}_t^\trans$,
where $\tilde{\vW}_t$ extends $\vW_{t|t-1}$ with $\nout$ additional columns
coming from the Jacobian of the observation model,
and then to project $\tilde{\vW}_t$ back to rank $\memory$  using
SVD to get $\vSigma_{t}^{-1} = \vUpsilon_t + \vW_t \vW_t^\trans$,
where $\vUpsilon_t$ is chosen so as to satisfy 
$\diag(\vSigma_t^{-1}) = \diag(\vSigma_t^{*-1})$.
See  \cref{algo:LOFI-update} for the resulting pseudocode.
The cost is dominated by the $O(\nparams \memoryout^{2})$ time needed
for the SVD, where $\memoryout = \memory + \nout$.\footnote{
Computing the SVD takes $O(\nparams (\memory + \nout)^2)$ time in the update step (for both spherical and diagonal approximations),
which may be too expensive. 
In \cref{appx:orth-svd} we derive a modified update step
which takes $O(\nparams \memory \nout)$ time, but which is less accurate.
The approach is
based on the \orfit method \citep{ORFit},
which uses orthogonal projections to make the SVD fast to compute.
However, we have found its performance to be quite poor
(no better than diagonal approximations), so we have omitted
its results.
}

To gain some intuition for the method,
suppose the output is scalar,  with 
variance $R=1$.
Then 
we have $A_t=1$
and $\vH_t^\trans=\nabla_{\vtheta_t} h(\vx_t,\vtheta_t)=\vg_t$
as the approximate linear observation matrix.
(Note that, for a linear model, we have $\vg_t = \vx_t$.)
In this case, we have 
  $\sizeout{\vW}_{t} = \left[\begin{array}{cc}
    \vW_{t | t-1} & \vg_t \end{array} \right]$.
Thus $\sizeout{\vW}_t$ acts like a generalized memory buffer that stores data using a gradient embedding. 
This allows an interpretation of our method in terms of the neural tangent kernel \citep{jacot2018neural},
although we leave the details to future work.

\subsection{Predicting the observations}

So far we have just described how to recursively update
the belief state for the parameters.
To predict the output $\vy_t$ given a test input $\vx_t$,
we need to compute
the one-step-ahead predictive distribution
\begin{align}
    p(\vy_t |\vx_t,\data_{1:t-1})
     &= \int p(\vy_t|\vx_t,\vtheta_t)  p(\vtheta_t|\data_{1:t-1}) d\vtheta_t 
     \label{eqn:postPred}
\end{align}
The negative log of this,
$-\log p(\vy_t|\vx_t,\data_{1:t-1})$,
is called the negative log predictive density or NLPD.
If we ignore the posterior uncertainty,
this integral gives us the following  plugin approximation,
given by
\begin{align}
    p(\vy_t |\vx_t,\data_{1:t-1})
     &\approx \int p(\vy_t|\vx_t,\vtheta_t)
     \gauss(\vtheta_t | \vmu_{t|t-1}, 0 \vI) d\vtheta_t 
     = p(\vy_t|\vx_t,\vmu_{t|t-1}) 
\end{align}
The negative log of this,
$-\log p(\vy_t|\vx_t,\vmu_{t|t-1})$,
is called the negative log likelihood or NLL.
We report NLL results in the main paper, since they are easy to compute.

However, we can get better performance by using more accurate
approximations to the integral.
The simplest approach is to use Monte Carlo sampling;
alternatively we can use deterministic approximations,
as discussed in \cref{appx:predict-obs}.
We find that naively passing posterior samples through
the model can result in worse performance than using the plugin approximation,
which just uses the posterior mode.
However, if we pass the samples through the linearized
observation model,
as proposed in \citep{Immer2021linear},
we find that the NLPD can outperform the NLL,
as shown in 
\cref{appx:classification}
and
\cref{sec:mnist-reg-extra}
in the appendix.

\eat{
Surprisingly, we find that,
for our classification experiments,
this plugin approximation
yields better results (lower NLL and lower misclassificaton rates) than the Monte Carlo approximation,
even if we linearize the model before sampling,
as proposed in \citep{Immer2021linear}.
We believe this is a manifestation of the ``cold posterior'' effect
described in \citep{Wenzel2020bayes,Adlam2020},
since setting the temperature to 0 is equivalent to just
using the MAP estimate.
Indeed, 
we experimented with tuning the temperature parameter before drawing posterior samples,
but found that the optimal
setting was usually $T=0$,
corresponding to the plugin approximation.
In the case of our regression experiments, we did not observe any big difference between the plugin NLL and an MC
approximation to the NLPD
(see \cref{fig:rotated-mnist-iid} in the appendix),
presumably because the observation noise was low,
so the posterior becomes sharply concentrated on the mode.
However, in the case of bandits,
discussed in \cref{sec:bandits},
we find that there is useful information in the uncertainty captured by the posterior.
In particular, we show that we can combine the LOFI
posterior with Thompson sampling to get improved results over other methods.
}

\subsection{Initialization
and hyper-parameter tuning}
\label{sec:init}


The natural way to  initialize the belief state
is use a vague Gaussian prior of the form
$p(\vtheta_0) = \gauss(\vzero, \vUpsilon_0)$,
where $\vUpsilon_0=\eta_0\vI_\nparams$ and $\eta_0$ is a hyper-parameter
that controls the strength of the prior.
However, plugging in all 0s for the weights will result in a prediction of 0,
which will result in a zero gradient, and so no learning will take place. (With $\vmu_0=0$, no deterministic algorithm can ever break the network's inherent symmetry under permutation of the hidden units.)
So in practice we sample the initial
mean weights using a standard neural network initialization procedure, 
such as ``LeCun-Normal'', which has the form
$\vmu_0 \sim \gauss(\vzero, \vS_0)$,
where $\vS_0$ is diagonal and $S_{0}[j,j] = 1/F_j$ is the fan-in of weight $j$.
(The bias terms are initialized to 0.)
We then set
 $\vUpsilon_0=\eta_0\vI_\nparams$
 and $\vW_0 = [0]^{\nparams\times\memory}$.\footnote{
 To make the prior accord with the non-spherical distribution from which we sample $\vmu_0$,
 we can scale the parameters by the fan-in, to convert to a standardized
 coordinate frame. However we found this did not seem to make any difference in practice, at least for our classification experiments.
 }


The hyper-parameters of our method are
the initial prior precision $\eta_0$,
 the dynamics noise $q$,
the dynamics scaling factor $\gamma$,
and (for regression problems),
the observation variance $\obsVar$.
These play a role similar
to the hyper-parameters of a standard neural network,
such as degree of regularization and the learning rate.
We optimize these hyper-parameters using
Bayesian optimization, where the objective
is the validation set NLL for stationary problems,
or the average one-step-ahead NLL (aka prequential loss)
for non-stationary problems.
For details, see \cref{sec:hparams}.

\section{Experiments}
\label{sec:results}

In this section, 
we report experimental results 
on various classification and regression datasets.
using
the following approximate inference techniques:
\lofi (this paper);
FDEKF (fully decoupled diagonal EKF) \citep{Puskorius2003};
VDEKF (variational diagonal EKF) \citep{Chang2022};
SGD-RB (stochastic gradient descent with FIFO replay buffer),
with memory buffer of size $\replay$, using either sgd or adam as the 
optimizer;
online gradient descent (OGD),
which corresponds to SGD-RB with $\replay=1$;
the LRVGA method of \citep{LRVGA} (for the NLPD results in \cref{sec:UCI-app});
and the 
online Laplace approximation of \citep{Ritter2018online}
(for the NLPD results in
\cref{appx:classification}
and
\cref{sec:mnist-reg-extra}).
For additional results,
see \cref{sec:extra}.
For the source code to reproduce these results, see 
\url{https://github.com/probml/rebayes}.

\eat{
 For most of the
experiments presented, we train a two-layer MLP (with 500 hidden units each),
which has $648,010$ parameters. 
(See \cref{fig:fmnist-st-cnn} for results with a CNN
which has 421,641 parameters.)
However, we stress that our method works for any kind of DNN, and scales linearly with the number of parameters.
}

\subsection{Classification}

In this section, we report results on various image classification datasets.
We use a 2-layer MLP (with 500 hidden units each),
which has $648,010$ parameters.
(For results using a CNN, see \cref{appx:classification} in the appendix.)

\paragraph{Stationary distribution}
\label{sec:clfStationary}


\begin{figure}
\centering
\begin{subfigure}[b]{0.47\textwidth}
\centering
\includegraphics[height=2.3in]
{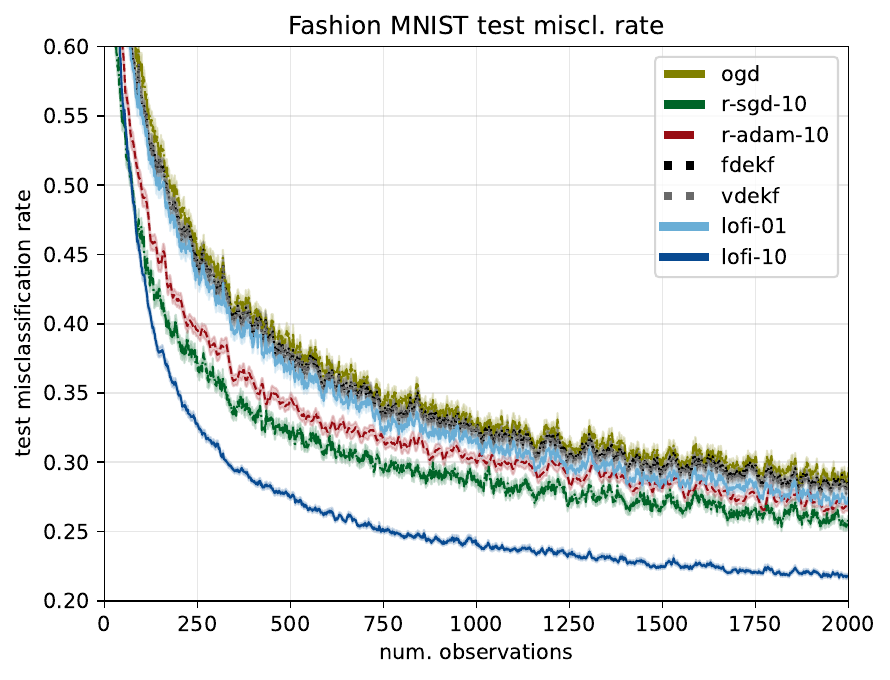}
\caption{ }
\label{fig:fmnist-st-clf-miscl}
\end{subfigure}
\begin{subfigure}[b]{0.47\textwidth}
\centering
\includegraphics[height=2.3in]{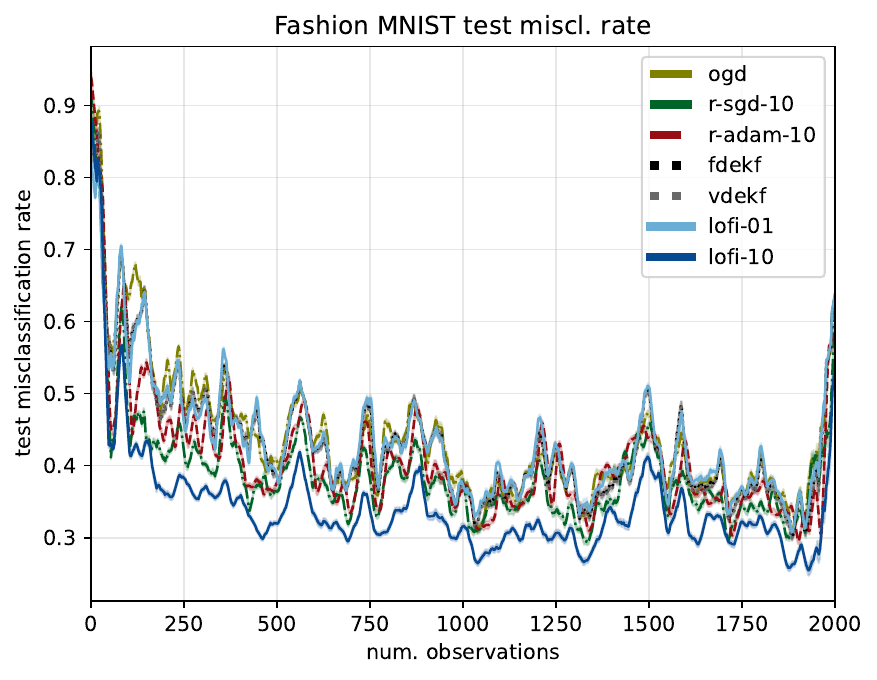}
\caption{ }
\label{fig:gr-clf}
\end{subfigure}
\begin{subfigure}[b]{\textwidth}
\centering
  \includegraphics[height=1.8in]
{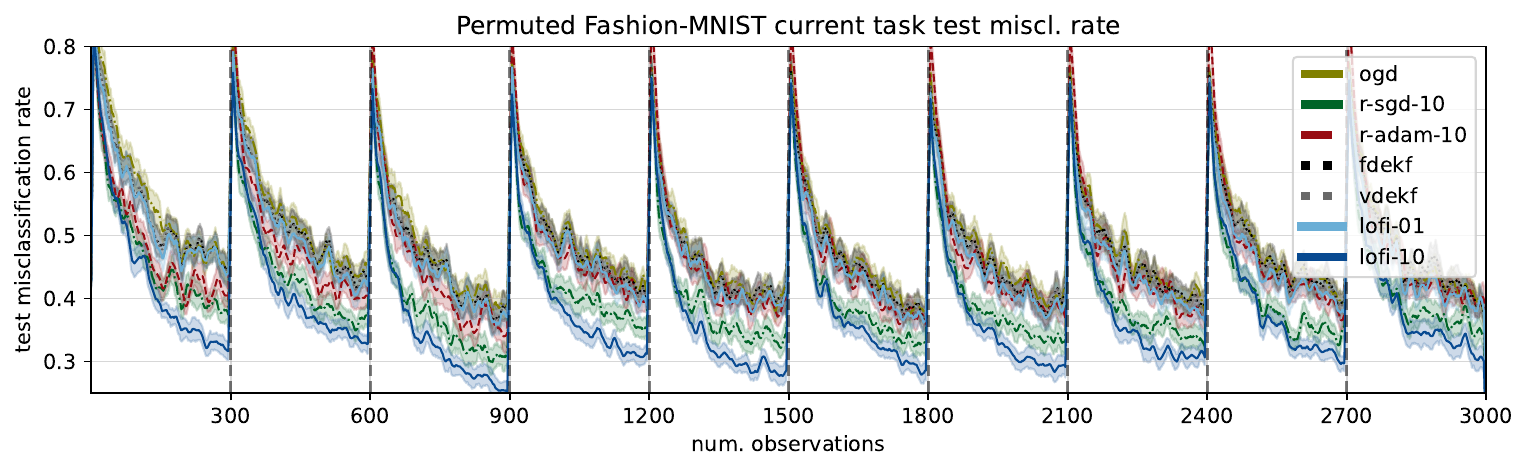}
\caption{ }
\label{fig:ns-pmnist-curr}
\end{subfigure}

\caption{
Test set misclassification rate vs number of observations on 
(a) the static fashion-MNIST dataset.
\collasclffiggen{generate\_stationary\_clf\_plots.ipynb}
(b) Gradually rotating fashion-MNIST.
\collasclffiggen{generate\_rotated\_clf\_plots.ipynb}
(c)
    Piecewise stationary permuted fashion-MNIST.
    The task boundaries are denoted by vertical lines.
    We show performance on the current task.
    \collasclffiggen{generate\_permuted\_clf\_plots.ipynb}
}
\end{figure}

We start by considering the fashion-MNIST image classification dataset (\cite{Xiao2017}).
For replay-SGD, we
use a replay buffer of size $10$ and tune the learning rate.
In \cref{fig:fmnist-st-clf-miscl} we plot the misclassification rate on the test set vs number of training samples using the MLP.
(We show the mean and
standard error over 100 random trials.)
We see that LOFI (with $\memory=10$)
is the most sample efficient learner,
then replay SGD (with $\buffer=10$), 
then replay Adam;
the diagonal EKF versions and OGD are the least 
sample efficient learners. 

In the appendix we show the following additional results.
In \cref{fig:fmnist-st-nll}
we show the results
using NLL as the evaluation metric;
in this case, the gap between LOFI and the other methods
is similarly noticeable.
In \cref{fig:fmnist-st-clf-nlpd} we show the results using
NLPD under the generalized probit approximation; the performance
gap reduces but \lofi is still the best method (see \cref{appx:predict-obs}
for discussion on analytical approximations to the NLPD).
In \cref{fig:fmnist-st-cnn} we show results using a CNN
(a LeNet-style architecture
with 3 hidden layers and 421,641 parameters);
trends are similar to the MLP case.
In \cref{fig:fmnist-st-cnn-vs-rank} we show how
changing the rank $\memory$ 
of \lofi affects performance within the range 1 to 50.
We see that for both NLL and misclassification rate,
larger $\memory$ is better, with gains plateauing at around $\memory\approx10$.
We also show that a spherical approximation to \lofi,
discussed in \cref{sec:spherical} in the appendix,
gives worse results.

\paragraph{Piecewise stationary distribution}
\label{sec:clfPiecewise}

To evaluate model performance in the non-stationary classification setting, we perform inference
under the incremental domain learning scenario
using the 
permuted-fashion-MNIST dataset  \citep{Hsu2018}.
After every $300$ training examples, the images are permuted randomly
and we compare performances across $10$ consecutive tasks.

\eat{
\footnote{
In a real non-stationary problem, where there are no task boundaries,
we can use the error in the one-step-ahead predictive distribution
as an evaluation metric,
as proposed in \citep{Gama2013}.
This is equivalent to the prequential loss
discussed in \citep{Dawid99,Bornschein2022}.
The results using this metric are
qualitatively similar to the ones shown above,
but are much noisier.
} %
}

In \cref{fig:ns-pmnist-curr} we plot the
performance over the current test set for each task
(each test size has  size $500$)
as a function of the number of training samples.
(We show mean and standard 
error across $20$ random initializations of the dataset).
The task boundaries are denoted
by vertical dotted lines (this boundary information is not available to the 
learning agents, and is only used for evaluation).
We see that LO-FI rapidly adapts to each new distribution and outperforms
all other methods.

In the appendix we show the following additional results.
In \cref{fig:ns-pmnist-curr-nll} we show the results
using NLL as the evaluation metric;
in this case, the gap between LOFI and the other methods
is even larger.
In \cref{fig:ns-pmnist-curr-rank},
we show misclassification
for the current task as a function of \lofi rank;
as before, performance increases with rank, and plateaus
at $\memory=10$.
%
In \cref{fig:smnist}, 
we show results on {\em split} fashion MNIST  \citep{Hsu2018},
in which each task corresponds to a new pair of classes.
However, since this is such an easy task
that all methods are effectively indistinguishable.

\paragraph{Slowly changing distribution}
\label{sec:clfDrifting}

The above experiments simulate an unusual form of non-stationarity,
corresponding to a sudden change in the task.
In this section, we consider a slowly changing distribution,
where the task is to classify the images as they slowly rotate.
The angle of rotation $\alpha_t$ gradually drifts according to an 
Ornstein-Uhlenbeck process,
so $d \alpha_t = -\theta (\mu- \alpha_t) dt + \sigma d W_t$,
where $W_t$ is a white noise process,
$\mu=45$, $\sigma=15$, $\theta=10$ and
$dt  = 1/N$, where $N=2000$ is the number of examples.
The test-set is modified using the same rotation at each step, perturbed
by a Gaussian noise with standard deviation of $5$ degrees.
To evaluate performance we use  a sliding window of size $200$
around the current time point.
%
The misclassification results are shown in \cref{fig:gr-clf}.
\lofi adapts to the continuously changing environment quickly 
and outperforms the other methods.
In \cref{fig:gr-NLL-NLPD}  in the appendix we show the NLL and NLPD,
which shows a similar trend.

\subsection{Regression}
\label{sec:regression}

In this section, we consider regression tasks using variants
of the fashion-MNIST dataset (images from class 2),
where we artificially rotate the images, and seek to predict the 
angle of rotation. As in the classification setting, we use a 2-hidden layer
MLP with 500 units per layer.

\eat{
To evaluate performance, we compute both the RMSE
(which is equal, up to a constant) to the NLL with a plugin MAP estimate ,
and a Monte-Carlo approximation to the 
 the negative log posterior predictive density (NLPD)
(see \cref{appx:predict-obs} for details).
We add the \textit{-map} suffix for methods evaluated using a plugin estimate of the posterior predictive,
and  \textit{-mc} suffix for methods that use a Monte Carlo approximation.
In the case of SGD and Adam, 
we use the online Laplace method of \citep{Ritter2018online} to compute the posterior,
using  a diagonal approximation to the Hessian.
The NLPD results are in the appendix, to save space.
}

\paragraph{Stationary distribution}
\label{sec:regStatic}

\begin{figure}[h]
\centering
\begin{subfigure}[b]{0.47\textwidth}
    \centering
    \includegraphics[height=2.3in]{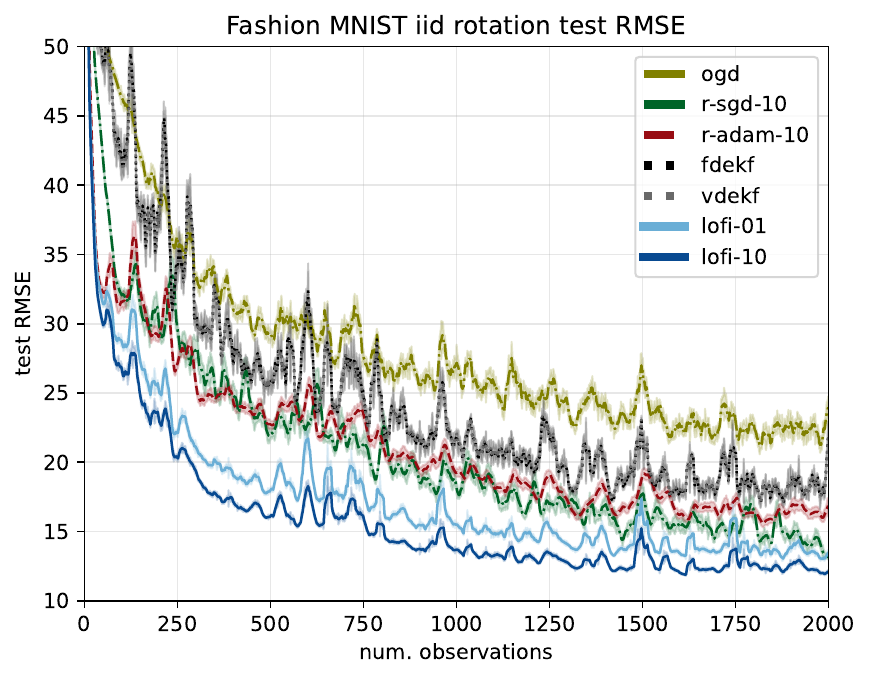}
    \caption{ }
    \label{fig:rotated-mnist-iid-rmse}
\end{subfigure}
\begin{subfigure}[b]{0.47\textwidth}
    \centering
    \includegraphics[height=2.3in]{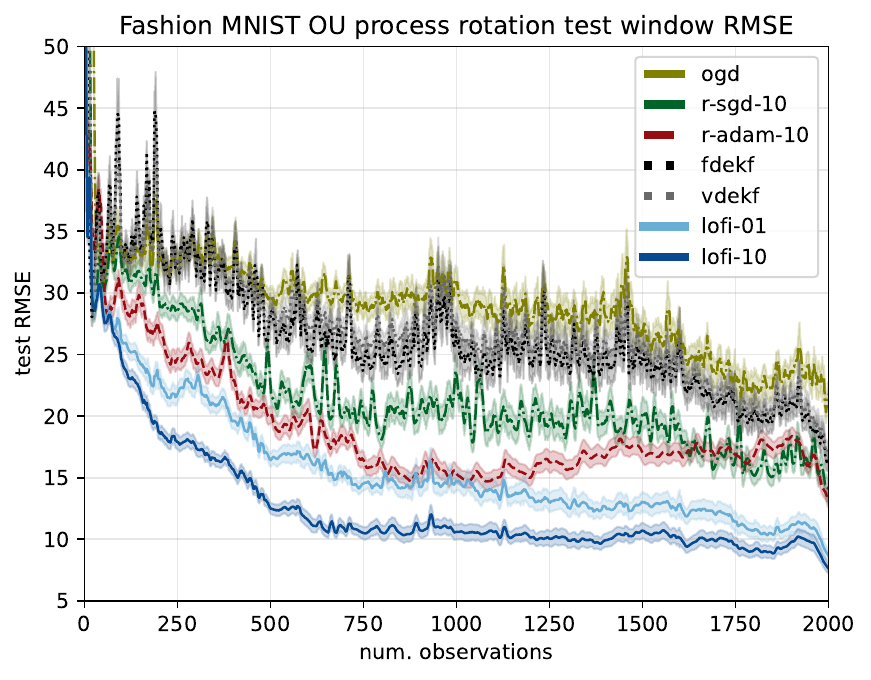}
    \caption{ }
    \label{fig:rotated-mnist-damped-rmse-all}
\end{subfigure}

\begin{subfigure}[b]{\textwidth}
    \centering
    \includegraphics[height=1.8in]{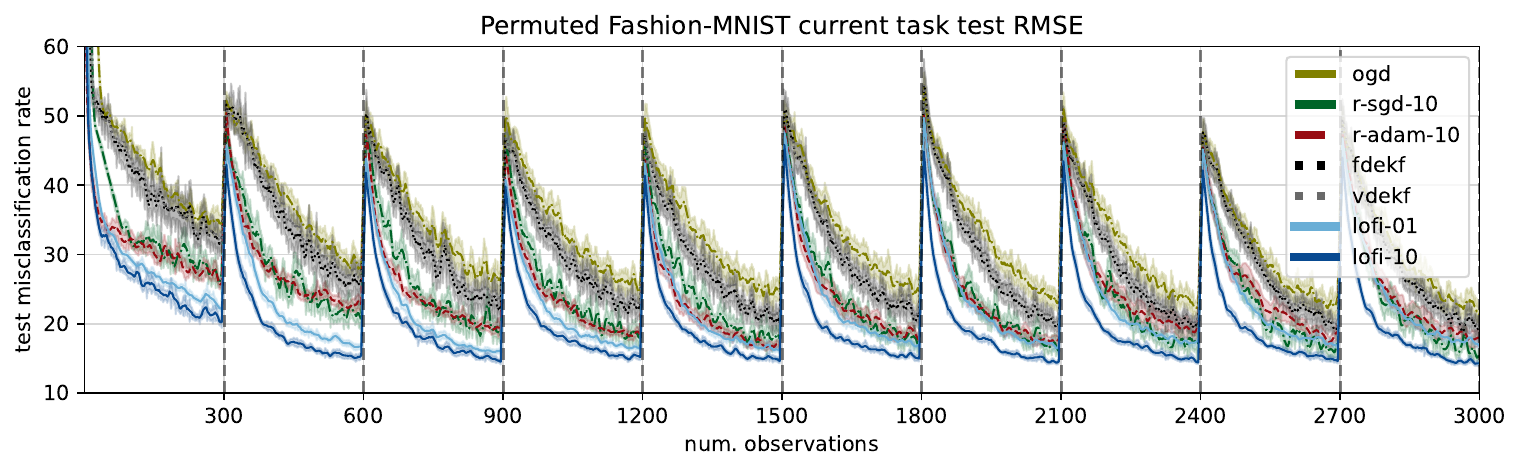}
    \caption{ }
    \label{fig:prmnist-reg}
\end{subfigure}

\caption{
  Test set regression error (measured using RMSE),
  computed using plugin approximation  on various datasets.
(a)  Static iid distribution of rotated MNIST images.
\collasregfiggen{generate\_iid\_reg\_plots.ipynb}
(b) Slowly changing version of rotated MNIST.
\collasregfiggen{generate\_rw\_reg\_plots.ipynb}
(c) Piecewise stationary permuted roated  MNIST.
    The task boundaries are denoted by vertical lines.
We show performance on the current task.
\collasregfiggen{generate\_permuted\_reg\_plots.ipynb}
}
\end{figure}

We start by sampling an iid dataset of images,
where the angle of rotation at time $t$ is sampled from a uniform ${\cal U}[0, 180]$ distribution.
In Figure \cref{fig:rotated-mnist-iid-rmse}, we show the RMSE over the test
set as a function of the number of trained examples;
we see that LOFI outperforms the other methods by a healthy margin.
(The NLL and NLPD results
in \cref{fig:rotated-mnist-iid} show a similar trend.)

\eat{
and in Figure \cref{fig:rotated-mnist-iid-nlpd}, we show the corresponding NLPD.
We see that the LOFI methods have the best performance, and have the lowest variance.
(Note that the NLPD is computed via an analytic
approximation to the posterior predictive under the linearized observation model,
and in the case of the GD methods, a diagonal version of the 
online Laplace approximation is used to capture the parameter uncertainty. For additional
discussion of these approximations, see \cref{appx:predict-obs}.)
}

\eat{
\begin{figure}
\centering
\includegraphics[height=1.3in]{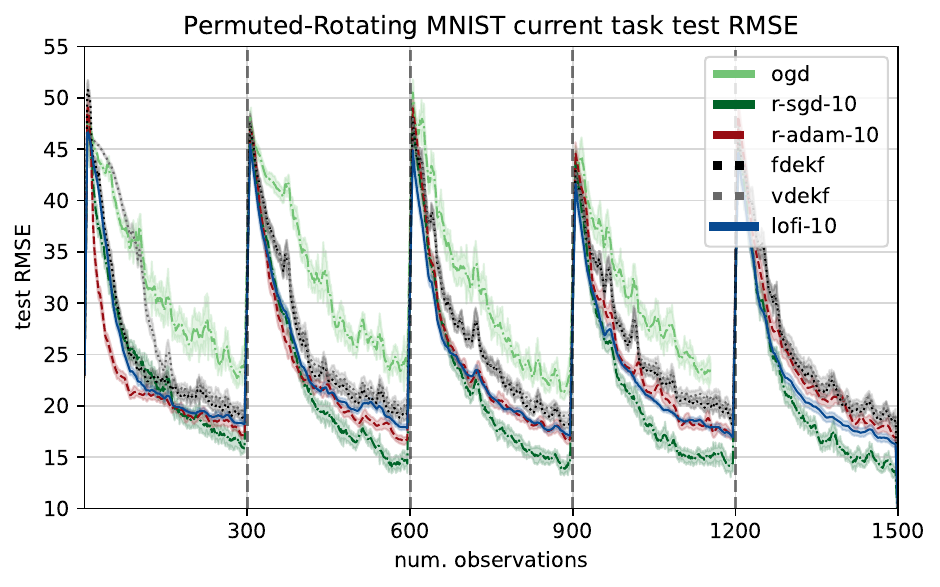}
\caption{
Permuted rotated fashion-MNIST regression problem.
We show RMSE on the current task.
}
\end{figure}
}

\paragraph{Piecewise stationary distribution}
\label{sec:regPiecewise}

We introduce nonstationarity through discrete task changes:
we randomly permute the fashion-MNIST dataset after every $300$
training examples, for a total of $10$ tasks.
This is similar to the classification
setting of \cref{sec:regPiecewise}, except
the prediction target is the angle, which is 
randomly sampled from $(0, 180)$ degrees.
The goal is to predict the rotation angle of test-set images
with the same permutation as the current task.
The results are shown in 
\cref{fig:prmnist-reg}.
We see that \lofi outperforms all other methods.

\paragraph{Slowly changing distribution}
\label{sec:regDrifting}


To simulate an arguably more realistic kind of change,
we consider the case where the rotation angle slowly changes,
generated via an Ornstein-Uhlenbeck process as in \cref{sec:clfDrifting},
except with parameters $\mu=90, \sigma=30$.
To evaluate performance we use a sliding window of size $200$,
applied to the test set whose rotations are generated by the
same rotations as the training set, except perturbed by a Gaussian
noise with standard deviation of $5$ degrees.
We show the results in 
\cref{fig:rotated-mnist-damped-rmse-all}.
We see that \LOFI outperforms the baseline methods.

\paragraph{Results on stationary UCI regression benchmark}
\label{sec:UCI}

\begin{figure}
\centering
\begin{subfigure}[b]{0.47\textwidth}
\centering
\includegraphics[height=2.0in]{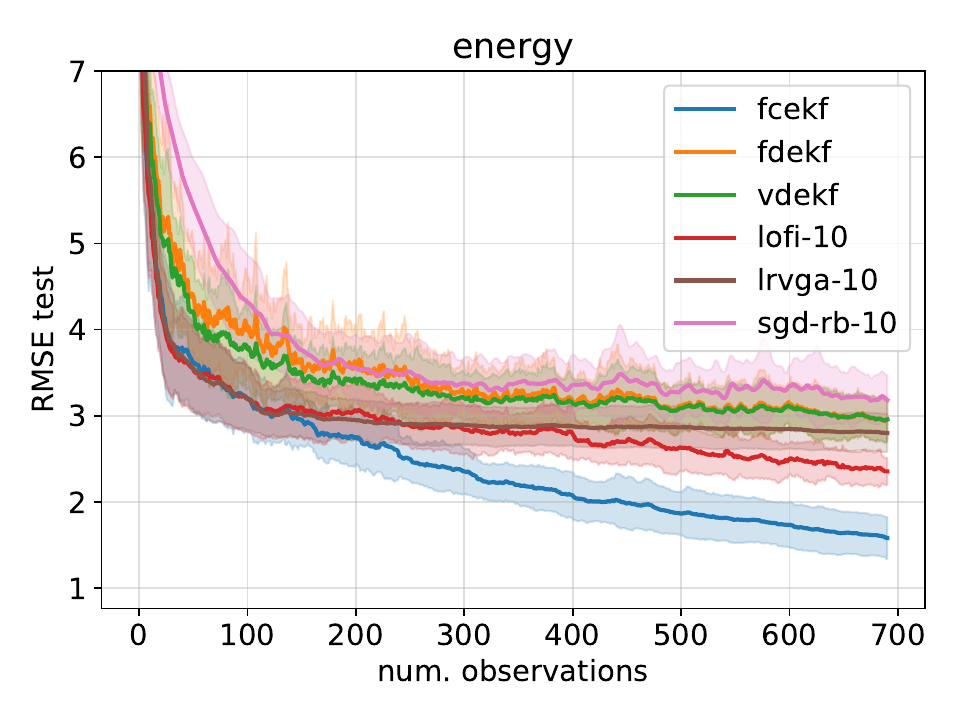}
\caption{ }
\label{fig:energy}
\end{subfigure}
\begin{subfigure}[b]{0.47\textwidth}
\centering
\includegraphics[height=1.5in]{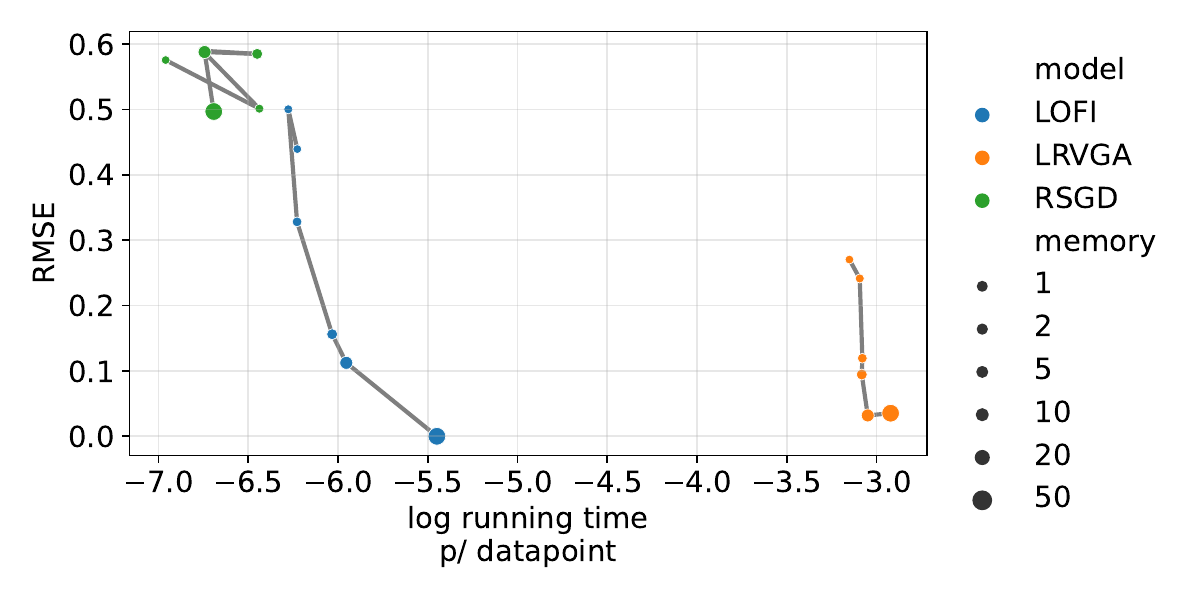}
\caption{ }
\label{fig:pareto}
\end{subfigure}
\caption{
(a) RMSE vs number of examples on the UCI energy dataset.
We show the mean and standard error across 20 partitions.
\showdownfiggen{plots-xval.ipynb}
(b) RMSE vs log running time  per data point averaged over
multiple UCI regression datasets.
The speedup of LOFI compared to LRVGA is
about $e^3 \approx 20$.
\showdownfiggen{time-analysis.ipynb}
}
\end{figure}

In this section, we evaluate various methods
on the UCI tabular regression benchmarks
used in  several other BNN papers
(e.g., \citep{HernandezLobato2015icml,Gal2016,Mishkin2018}).
We use the same splits as in \citep{Gal2016}.
As in these prior works,
we consider an MLP with 1 hidden layer of $H=50$ units
using RELU activation,
so the number of parameters is $\nparams = (D+2)H + 1$,
where $D$ is the number of input features.
In Table \ref{tab:uci-regression-description} in the appendix, we show the number of features in each dataset, as well
as the number of training and testing examples in each of the 20 partitions.

We use these small datasets to compare \LOFI with \LRVGA,
as well as the other baselines.
We show the RMSE vs number of training examples for the Energy dataset in  \cref{fig:energy}.
In this case, we see that \LOFI (rank 10)  outperforms \LRVGA (rank 10),
and both outperform diagonal EKF and SGD-RB (buffer size 10).
However, full covariance EKF is the most sample efficient learner.
On other UCI datasets, \LRVGA can slightly outperform \LOFI
(see \cref{sec:UCI-app} for details).
However, it is about 20 times slower than LOFI.
This is visualized in \cref{fig:pareto},
which shows RMSE vs 
 compute time,
 averaged over the 8 UCI datasets listed in 
 \cref{tab:uci-regression-description}.
 This shows that, controlling for compute costs,
 \LOFI is a more efficient estimator,
 and both outperform replay SGD.

\subsection{Contextual bandits}
\label{sec:bandits}

\begin{wrapfigure}{R}{0.5\textwidth}
    \centering
    \includegraphics[width=1.0\linewidth]{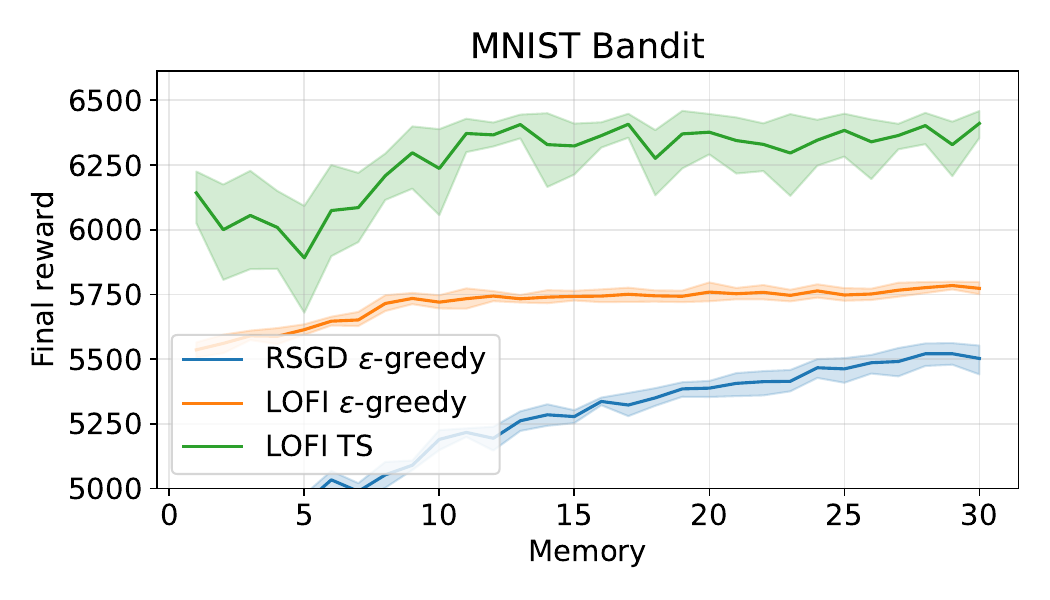}
    \label{fig:bandits}
    \caption{
        Total reward on MNIST bandit problem after 8000 steps
    vs memory of the posterior approximation. 
    We show results (averaged over 5 trials)
    using Thompson sampling or $\epsilon$-greedy with $\epsilon=0.1$.
    See text for details. \banditfiggen{bandit-vs-memory.ipynb}}
    \vspace{-30pt}
\end{wrapfigure}

In this section, we illustrate the utility of an online Bayesian inference method by applying it to a contextual bandit problem.
Following prior work 
(e.g., \citep{Duran-Martin2022}),
we convert the MNIST classification problem into a bandit problem by defining the action space as a label from 0 to 9, and defining the reward to be 1 if the correct label is predicted, and 0 otherwise.
For simplicity, we model this using a nonlinear Gaussian regression model, rather than a nonlinear Bernoulli classification model.
To tackle the exploration-exploration tradeoff, we 
either use Thompson sampling (TS) or the simpler $\epsilon$-greedy baseline.
In TS, we sample a parameter from the posterior,
$\tilde{\vtheta}_t \sim p(\vtheta_t|a_{1:t-1}, \vx_{1:t-1}), r_{1:t-1})$
and then take the greedy action with this value plugged in,
$a_t = \argmax_a E[r|\vx_t, \tilde{\vtheta}_t]$.
This method is known to obtain optimal regret \citep{Russo2018},
although the guarantees are weaker
when using approximate inference
\citep{Phan2019}.
Of course, TS requires access to a posterior distribution
to sample from.
To compare to methods (such as SGD) that just compute a point estimate,
we also use $\epsilon$-greedy;
in this approach, with probability $\epsilon=0.1$
we try a random action (to encourage exploration),
and with probability $1-\epsilon$
we pick the best action, as predicted by plugging in the MAP parameters into the reward model.


In \cref{fig:bandits}, we compare these algorithms
on the MNIST bandit problem,
where the regression model is a simple MLP 
with the same architecture as shown in Figure 1b 
of \citep{Duran-Martin2022}.
For the $\epsilon$-greedy exploration policy
we use $\epsilon=0.1$, where the MAP parameter estimate is
either  computed using \lofi
(where the rank is on the $x$-axis)
or  using SGD with replay buffer
(where the buffer size
is on the $x$-axis).
We also show results of using TS with \lofi.
We see see that TS is much better than $\epsilon$-greedy with LOFI MAP estimate,
which in turn is better than $\epsilon$-greedy with SGD MAP estimate.
In \cref{fig:bandits-vs-time-app} in the appendix, we plot
reward vs time for these methods.

\section{Conclusion and future work}
\label{sec:concl}

We have presented an efficient new method of 
fitting neural networks online to streaming datasets, 
using a diagonal plus low-rank Gaussian approximation.
In the future, we are interested in developing online methods for
estimating the hyper-parameters, perhaps by
extending the variational Bayes approach of
\citep{Huang2020,viking},
or the gradient based method of \citep{Greenberg2021}.
\eat{
hierarchical Bayesian
approach, in which the neural network parameters get updated quickly,
and the SSM parameters get updated more slowly
\citep[c.f.,][]{deFreitas00ekf}
}
We would also like to further explore the predictive uncertainty
created by our posterior approximation, to see if it can be used
for sequential decision making tasks, such as Bayesian optimization
or active learning.
This may require the use of (online) deep Bayesian ensembles,
to capture functional as well as parametric uncertainty.

\clearpage

\bibliography{refs}
\bibliographystyle{collas2023_conference}

\appendix

\clearpage

\section{Derivations}

\subsection{Predict step}
\label{appx:predict-step}

We begin with the posterior from the previous time step
\begin{align}
    p(\vtheta_{t-1} | \data_{1:t-1}) 
    = \gauss \left(
        \vtheta_{t-1} | \vmu_{t-1}, 
        \left(
            \vUpsilon_{t-1}
            + \vW_{t-1} \vW_{t-1}^\trans
        \right)^{-1}
    \right)
\end{align}
and the dynamic assumption
\begin{align}
    p(\vtheta_t|\vtheta_{t-1}) = 
    \gauss(\vtheta_t | \gamma_t \vtheta_{t-1}, q_t \vI_\nparams)
    \label{eq:predict-appx-dynamics}
\end{align}
These imply the prior on the current time step is $p(\vtheta_{t}|\data_{1:t-1}) = \gauss(\vtheta_{t}|\vmu_{t|t-1},\vSigma_{t|t-1})$ with
\begin{align}
    \vmu_{t|t-1} &= \gamma_t \vmu_{t-1} 
    \label{eq:mean-predict-appx}\\
    \vSigma_{t|t-1} &=
        \gamma_t^2
        \left(
            \vUpsilon_{t-1}
            + \vW_{t-1} \vW_{t-1}^\trans
        \right)^{-1}
        + q_t \vI_\nparams
        \label{eq:predict-step-covariance-unsolved}
\end{align}
Applying the Woodbury identity to \cref{eq:predict-step-covariance-unsolved} gives this expression for the prior covariance:
\begin{align}
    \vSigma_{t|t-1} &=
        \gamma_t^2
        \left(
            \vUpsilon_{t-1}^{-1}
            - \vUpsilon_{t-1}^{-1} \vW_{t-1} 
            \left(
                \vI_{\memory}
                + \vW_{t-1}^{\trans} \vUpsilon_{t-1}^{-1} \vW_{t-1}
            \right)^{-1}
            \vW_{t-1}^{\trans} \vUpsilon_{t-1}^{-1}
        \right)
        + q_t \vI_\nparams \\
        &= \vUpsilon_{t|t-1}^{-1}
        - \vUpsilon_{t-1}^{-1} \vW_{t-1} \vB_{t|t-1} \vW_{t-1}^{\trans} \vUpsilon_{t-1}^{-1}
\end{align}
where
\begin{align}
    \vUpsilon_{t|t-1} &=
    \left(\gamma_t^{2} \vUpsilon_{t-1}^{-1}+q_t\vI_\nparams\right)^{-1} 
    \label{eq:diagonal-predict-appx}\\
    \vB_{t|t-1} &=
    \gamma_t^{2}
    \left(
        \vI_{\memory}
        + \vW_{t-1}^{\trans} \vUpsilon_{t-1}^{-1} \vW_{t-1}
    \right)^{-1}
\end{align}
Applying Woodbury again yields this expression for the prior precision:
\begin{align}
    \vSigma_{t|t-1}^{-1} &=
    \left( 
        \vUpsilon_{t|t-1}^{-1}
        - \vUpsilon_{t-1}^{-1} \vW_{t-1} \vB_{t|t-1} \vW_{t-1}^{\trans} \vUpsilon_{t-1}^{-1} 
    \right)^{-1} \\
    &= \vUpsilon_{t|t-1}
    + \vUpsilon_{t|t-1} \vUpsilon_{t-1}^{-1} \vW_{t-1}
    \left(
        \vB_{t|t-1}^{-1}
        - \vW_{t-1}^{\trans} \vUpsilon_{t-1}^{-1} \vUpsilon_{t|t-1} \vUpsilon_{t-1}^{-1} \vW_{t-1}
    \right)^{-1}
    \vW_{t-1}^{\trans} \vUpsilon_{t-1}^{-1} \vUpsilon_{t|t-1} \\
    &= \vUpsilon_{t|t-1} + \vW_{t|t-1} \vW_{t|t-1}^{\trans}
\end{align}
where
\begin{align}
    \vW_{t|t-1} &=
    \vUpsilon_{t|t-1} \vUpsilon_{t-1}^{-1} \vW_{t-1}
    {\rm chol} \left( \left(
        \vB_{t|t-1}^{-1}
        - \vW_{t-1}^{\trans} \vUpsilon_{t-1}^{-1} \vUpsilon_{t|t-1} \vUpsilon_{t-1}^{-1} \vW_{t-1}
    \right)^{-1} \right) \\
    &= \gamma_t  \vUpsilon_{t|t-1} \vUpsilon_{t-1}^{-1} \vW_{t-1} \,
        {\rm chol} \left( \left(
            \vI_{\memory}
            + q_t \vW_{t-1}^{\trans}  \vUpsilon_{t|t-1} \vUpsilon_{t-1}^{-1} \vW_{t-1}
        \right)^{-1} \right)
        \label{eq:lowrank-predict-appx}
\end{align}
 Calculating $\vUpsilon_{t|t-1}$ and $\vW_{t|t-1}$ respectively take $O(\nparams)$ and $O(\nparams\memory^2+\memory^3)$ time.
See  \cref{algo:LOFI-predict-UL} for the pseudocode;
this is the same as \cref{algo:LOFI-predict}
except we replace $\vW_t$ with $\vU_t \vLambda_t$,
as a stepping stone to 
the spherical version in 
\cref{sec:spherical}.

\begin{algorithm}
def $\text{predict}(
\vmu_{t-1}, \vUpsilon_{t-1}, \vLambda_{t-1}, \vU_{t-1}, \vx_t, \gamma_t, q_t)$: 
\\
$\vW_{t-1} = \vU_{t-1} \vLambda_{t-1}$
// Recreate the low-rank precision \\
$\vmu_{t|t-1} = \gamma \vmu_{t-1}$
// Predict the mean of the next state
\\
$\vUpsilon_{t|t-1} = 
    \left(\gamma_t^{2} \vUpsilon_{t-1}^{-1}
    +q_t\vI_\nparams\right)^{-1}$
    // Predict the diagonal precision 
\\
$\vC_t = 
\left(
            \vI_{\memory}
            + q_t \vW_{t-1}^{\trans}  \vUpsilon_{t|t-1} \vUpsilon_{t-1}^{-1} \vW_{t-1}
        \right)^{-1}$
    \\
$\vW_{t|t-1} =
    \gamma_t  \vUpsilon_{t|t-1} \vUpsilon_{t-1}^{-1} \vW_{t-1} 
    {\rm chol}(\vC_t)$
    // Predict the low-rank precision
\\
$\vU_{t|t-1}=\vW_{t|t-1}$
// For compatibility with spherical \lofi \\
$\vLambda_{t|t-1}=\vone$ // Arbitrary scaling
\label{line:SVD-predict-step}
\\
$\hat{\vy}_t =
h\left(\vx_t, \vmu_{t|t-1} \right)$ 
// Predict the mean of the output 
\\
Return $( \vmu_{t|t-1}, \vUpsilon_{t|t-1}, 
\vLambda_{t|t-1}, \vU_{t|t-1}, \hat{\vy}_t)$ 
\caption{\lofi predict step.}
\label{algo:LOFI-predict-UL}
\end{algorithm}

\subsection{Update step}
\label{appx:update}

After creating a linear-Gaussian approximation
to the likelihood (as explained in the main text),
standard  results 
\citep[see e.g.,][]{Sarkka23}
imply the 
exact posterior can be written as
$p(\vtheta_{t}|\data_{1:t}) =
\gauss(\vtheta_t  |     \vmu_{t}, \vSigma_{t}^{*})$,
where
\begin{align}
      \vSigma_{t}^{*-1} &=  \vSigma_{t|t-1}^{-1} + \vH_t^\trans \vR_t^{-1} \vH_t 
    \label{eq:EKF-var-update} \\
    \vK_t &= \vSigma_{t}^{*} \vH_t^\trans \vR_t^{-1} 
    \label{eq:kalman-gain}\\
    \ve_t &= \vy_t - \hat{\vy}_t \label{eq:innovation}\\
    \vmu_{t} &= \vmu_{t|t-1} + \vK_t  \ve_t
    \label{eq:EKF-mu-update}
\end{align}
where $\vK_t$ is known as the Kalman gain matrix,
and $\ve_t$ is the innovation vector (i.e., error in the prediction).

We now derive a low-rank version of the above update equations.
Because $\vR_t$ is positive-definite, we can write $\vR_t^{-1} = \vA_t^\trans\vA_t$.
We then define the  matrix
\begin{align}
    \sizeout{\vW}_{t} = \left[\begin{array}{cc}
    \vW_{t | t-1} 
    & \vH_{t}^{\trans} \vA_{t}^{\trans}
\end{array}\right]
\label{eqn:Wtilde}
\end{align}
This has size $\nparams \times \memoryout$,
where $\memoryout = \memory + \nout$.
Note that if the output is scalar,  with 
variance $R=\sigma^2$,
we have $\vH_t=\nabla_{\vtheta_t} h(\vx_t,\vtheta_t)$.
For a linear model, 
$h(\vx_t,\vtheta_t) = \vtheta_t^\trans \vx_t$,
the gradient equals the data vector $\vx_t$.
In this case, we have 
\begin{align}
  \sizeout{\vW}_{t} = \left[\begin{array}{cc}
    \vW_{t | t-1} & \frac{1}{\sigma} \vx_t \end{array} \right]
    \end{align}
Thus $\sizeout{\vW}_t$ acts like a generalized memory buffer that stores data using a gradient embedding.

From \cref{eq:EKF-var-update},
\eat{and \cref{eqn:post-pred-prec}}
the exact Bayesian inference step for the precision is
\begin{align}
    \vSigma_{t}^{*-1} 
    &= \vSigma_{t|t-1}^{-1}
    + \vH_{t}^{\trans} \vA_{t}^{\trans} \vA_{t} \vH_{t} \\
    &= \vUpsilon_{t|t-1}
    + \vW_{t|t-1} \vW_{t|t-1}^\trans 
   + \vH_{t}^{\trans} \vA_{t}^{\trans} \vA_{t} \vH_{t} \\
    &= \vUpsilon_{t|t-1}
    + \sizeout{\vW}_{t} \sizeout{\vW}_{t}^{\trans}
    \label{eq:posterior-precision-exact}
\end{align}

From \cref{eq:kalman-gain,eq:innovation,eq:EKF-mu-update}, the exact mean update is given by
\begin{align}
    \vmu_{t}=\vmu_{t|t-1}+\vSigma_{t}^{*}\vH_{t}^{\trans}\vR_{t}^{-1}\ve_{t}
    \label{eq:mean-update-unsolved}
\end{align}
Applying the Woodbury identity to \cref{eq:posterior-precision-exact} and substituting into \cref{eq:mean-update-unsolved}, we obtain an expression that can be computed in
$O(\nparams\memoryout^{2})$ time:
\begin{equation}
    \vmu_{t}=\vmu_{t|t-1}+\left(\vUpsilon_{t|t-1}^{-1}-\vUpsilon_{t|t-1}^{-1}\sizeout{\vW}_{t}\left(\vI_{\memoryout}+\sizeout{\vW}_{t}^{\trans}\vUpsilon_{t|t-1}^{-1}\sizeout{\vW}_{t}\right)^{-1}\sizeout{\vW}_{t}^{\trans}\vUpsilon_{t|t-1}^{-1}\right)\vH_{t}^{\trans}\vR_{t}^{-1}\ve_{t}
    \label{eq:mean-update} 
\end{equation}

\Cref{eq:posterior-precision-exact,eq:mean-update} give the exact posterior,
given the DLR($\memory$) prior.
However, to propagate this posterior to the next step,
we need to project $\vSigma_{t}^{*-1}$
from $\text{DLR}(\memoryout)$ back to $\text{DLR}(\memory)$.
To do this, we first 
perform an SVD of $\tilde{\vW}_t$
to get the new basis:
\begin{align}
    (\sizeout{\vLambda}_t,\sizeout{\vU}_t) &= {\rm SVD}(\sizeout{\vW}_t)
        \label{eq:SVD-update-step}\\
    \vW_{t} &= 
    \left(\sizeout{\vU}_t \sizeout{\vLambda}_t\right)[:,1{:}\memory]
        \label{eq:SVD-update-step-topL}
 \end{align}
 Here, $\tilde{\vLambda}_t$ and $\tilde{\vU}_t$ are respectively the singular values and left singular vectors of $\tilde{\vW}_t$, assumed to be ordered in decreasing value of $\tilde{\vLambda}_t$
 (so $\tilde{\vLambda}_t$ is diagonal
 of size $\memoryout \times \memoryout$,
 and  $\tilde{\vU}_t$
 is of size $\nparams \times \memoryout$). 
 \eat{
Retaining the top $\memory$ of these as $\vW_t$ achieves the Frobenius projection:  
\begin{align}
  \vW_t &= 
  \argmin_{\vW\in\real^{\nparams\times\memory}} ||\tilde{\vW}_t \tilde{\vW}_t^\trans 
   - \vW \vW^\trans||_F
   \label{eq:frobenius-projection}
\end{align}
}
 Finally, we update the diagonal term as follows:
 \begin{align}
     \vUpsilon_{t} & =\vUpsilon_{t|t-1} +
    \diag\left(\vW_t^\times \vW_t^{\times\trans}\right)
    \label{eq:diagonal-update}
    \\
    \vW_t^\times &= 
    \left(\sizeout{\vU}_t \sizeout{\vLambda}_t\right)[:,(\memory+1){:}\memoryout]
\end{align}
Adding the diagonal contribution from the remaining $\nout$ singular vectors to $\vUpsilon_{t}$ ensures the diagonal portion of the DLR approximation is exact, i.e.,  
\begin{align}
    \diag(\vSigma_t^{-1}) = \diag(\vSigma_t^{*-1}) 
    \label{eq:exact-diagonal}.
\end{align}

See  \cref{algo:LOFI-update-UL} for the pseudocode.
This is the same as \cref{algo:LOFI-update}
except we replace $\vW_t$ with $\vU_t \vLambda_t$.
This procedure takes $O(\nparams\memoryout^{2})$ time for the SVD,
and
$O(\nparams\nout)$ for calculating $\diag\left(\vW_{t}^{\times}\vW_{t}^{\times\trans} \right)$.\footnote{
Suppose $\vA \in \real^{n \times m}$
and $\vB \in \real^{m \times n}$.
Then we can efficiently compute
$\diag(\vA \vB)$ in $O(m n)$ time
using
$(\vA \vB)_{ii} = \sum_{j=1}^M
A_{ij} B_{ji}$.
}

\begin{algorithm}
def $\text{update}(
\vmu_{t|t-1}, \vUpsilon_{t|t-1}, \vLambda_{t|t-1}, \vU_{t|t-1},
\vx_t, \vy_t, \hat{\vy}_t, h, \memory)$: 
\\
$\vR_t = h_V(\vx_t, \vmu_{t|t-1})$
// Covariance of predicted output  
\\
$\vL_t = \text{chol}(\vR_t)$ \\
$\vA_t = \vL_t^{-1}$ \\
$\vH_t = \text{jac}(h(\vx_t,\cdot))(\vmu_{t|t-1})$
// Jacobian of observation model 
\\
$\vW_{t|t-1} =  \vU_{t | t-1} \vLambda_{t|t-1}$
// Predicted low-rank precision
\\
$\sizeout{\vW}_{t} = \left[\begin{array}{cc}
   \vW_{t|t-1}
    & \vH_{t}^{\trans} \vA_{t}^{\trans}
    \end{array}\right]$
    // Expand low-rank with new observation
\\
$\vG_t = \left(\vI_{\memoryout}+
\sizeout{\vW}_{t}^{\trans}\vUpsilon_{t|t-1}^{-1}
\sizeout{\vW}_{t}\right)^{-1}
$
\\
$\vC_t = \vH_{t}^{\trans} \vA_t^\trans \vA_t$
\\
$\vK_t = \vUpsilon_{t|t-1}^{-1} \vC_t
-\vUpsilon_{t|t-1}^{-1} \sizeout{\vW}_{t} \vG_t
\sizeout{\vW}_{t}^{\trans} \vUpsilon_{t|t-1}^{-1}
\vC_t$  // Kalman gain matrix
\\
$\vmu_t = \vmu_{t|t-1} + \vK_t(\vy_t - \hat{\vy}_t)$ 
// Mean update 
\\
$(\sizeout{\vLambda}_t,\sizeout{\vU}_t) = {\rm SVD}(\sizeout{\vW}_t)$ 
// Take SVD of the expanded low-rank \\
$(\vLambda_{t}, \vU_{t}) = 
    \left(\sizeout{\vLambda}_t, \sizeout{\vU}_t\right)[:,1{:}\memory]$
    // Keep top $L$ most important terms
\\
$(\vLambda_t^\times, \vU_t^\times) = 
    \left(\sizeout{\vLambda}_t, \sizeout{\vU}_t\right)[:,(\memory+1){:}\memoryout] $
    // Extra least important terms 
\\
$\vW_t^{\times} = \vU_t^{\times} \vLambda_t^{\times}$
// The low-rank part that is dropped \\
$\vUpsilon_{t} =
    \vUpsilon_{t|t-1} +
    \diag\left(\vW_t^{\times} (\vW_t^{\times})^{\trans} \right)$
    // Update diagonal to capture variance due to dropped terms
\\
Return $(\vmu_t, \vUpsilon_t, \vLambda_t, \vU_t)$ 
\caption{\lofi update step.
}
\label{algo:LOFI-update-UL}
\end{algorithm}

\subsection{Alternative diagonal update}

Instead of updating $\vUpsilon_t$ to achieve $\diag(\vSigma_t^{-1}) = \diag(\vSigma_t^{*-1})$,
we can minimize the KL divergence. If we define 
\begin{align}
    \vUpsilon_t = \argmin_{\vUpsilon} \KLpq {\gauss\left(\vmu_t,\left(\vUpsilon+\vW_t\vW_t^\trans\right)^{-1}\right)}
    {\gauss(\vmu_t,\vSigma_t^*)}
\end{align}
then we get the condition
\begin{align}
    \diag(\vSigma_t - \vSigma_t \vSigma_t^{*-1}\vSigma_t) = 0
\end{align}

If instead we use forward KL,
\begin{align}
    \vUpsilon_t = \argmin_{\vUpsilon} \KLpq
    {\gauss(\vmu_t,\vSigma_t^*)}
    {\gauss\left(\vmu_t,\left(\vUpsilon+\vW_t\vW_t^\trans\right)^{-1}\right)}
\end{align}
then we get the condition
\begin{align}
    \diag(\vSigma_t) = \diag(\vSigma_t^*)
\end{align}
We leave exploration of possible efficient implementations of these updates to future work.

\subsection{Zero-rank \lofi}

When $\memory=0$, \lofi approximates the covariance simply as
\begin{align}
    \vSigma_t = \vUpsilon_t^{-1}
\end{align}

Consequently, the predict step comprises only \cref{eq:mean-predict-appx,eq:diagonal-predict-appx}, repeated here:
\begin{align}
    \vmu_{t|t-1} &= \gamma_t {\vmu}_{t-1} \\
    \vUpsilon_{t|t-1} &=
    \left(\gamma_t^{2} {\vUpsilon}_{t-1}^{-1}+q_t\vI_\nparams\right)^{-1} 
\end{align}

In the update step, $\vW_{t|t-1}$ is empty, so $\vW_t^\times = \sizeout{\vW}_t = \vH_t^\trans \vA_t^\trans$. 
Therefore \cref{eq:mean-update,eq:diagonal-update} become
\begin{align}
    \vmu_{t} &= \vmu_{t|t-1}
    + \vUpsilon_{t|t-1}^{-1} \vH_t^\trans
    \left(
        \vH_t \vUpsilon_{t|t-1}^{-1} \vH_t^\trans
        + \vR_t
    \right)^{-1}
    \ve_{t} \\
    \vUpsilon_t &= \vUpsilon_{t|t-1} + \diag(\vH_t^\trans \vR_t^{-1} \vH_t)
\end{align}

Finally, in the predictive distribution for the observation, the variance in \cref{eqn:pred-obs-var-woodbury} simplifies:
\begin{align}
    \hat{\vy}_t &= h(\vx_t, \vmu_{t|t-1}) \\
    \vV_{t} &= \vH_{t} \vUpsilon_{t|t-1}^{-1} \vH_{t}^{\trans} + \vR_{t}
\end{align}

These equations match those of the VD-EKF \citep{Chang2022}, confirming that \lofi reduces to VD-EKF when $\memory=0$.

\clearpage
\section{Posterior predictive distribution for the observations}
\label{appx:predict-obs}

In this section, we discuss how to use the posterior
over parameters to approximate the posterior predictive distribution for the observations:
\begin{align}
    p(\vy_t |\vx_t,\data_{1:t-1})
     &= \int p(\vy_t|\vx_t,\vtheta_t)  p(\vtheta_t|\data_{1:t-1}) d\vtheta_t 
\end{align}
A simple approach is to use a plugin approximation,
which arises when we assume the posterior is a point estimate:
\begin{align}
    p(\vy_t |\vx_t,\data_{1:t-1})
     &\approx \int p(\vy_t|\vx_t,\vtheta_t)  \delta(\vtheta_t  - \hat{\vtheta}_t) d\vtheta_t \\
     &=
     \begin{cases}
      \gauss(\vy_t|h(\vx_t, \hat{\vtheta}_t), \vR_t)  & \mbox{regression} \\ \cat(\vy_t|\softmax(h(\vx_t, \hat{\vtheta}_t))  
     & \mbox{classification}
     \end{cases}
\end{align}
We can capture more uncertainty by sampling parameters from the (Gaussian) posterior,
$\vtheta_t^s \sim \gauss(\vmu_{t|t-1}, \vSigma_{t|t-1})$,
which results in the following Monte Carlo approximation:
\begin{align}
    p(\vy_t |\vx_t,\data_{1:t-1})
     &\approx 
     \begin{cases}
     \frac{1}{S} 
     \sum_{s=1}^S \gauss(\vy_t|h(\vx_t, \vtheta_t^s), \vR_t)  & \mbox{regression} \\
    \frac{1}{S} 
     \sum_{s=1}^S \cat(\vy_t|\softmax(h(\vx_t, \vtheta_t^s))  
     & \mbox{classification}
     \end{cases}
\end{align}
If we have a DLR approximation to the precision matrix, we can use the 
importance sampling method of Section 6.2 of \citep{LRVGA} to draw samples in 
$O(\nparams S)$ time, without needing to create or invert the full precision matrix.
 
However, as argued in \citep{Immer2021linear}, it can sometimes be better to 
approximate the predictive distribution by first linearizing the 
observation model, and then passing the samples through the linearized model, 
to avoid evaluating the nonlinear function with parameter values that are far 
from the posterior mode.
Once we have linearized the model, we can further replace the Monte Carlo 
approximation with a deterministic integral, as we explain below.

\subsection{Deterministic approximation for regression}

If we linearize the observation model,
and assume a Gaussian output,
we can compute the  posterior predictive distribution analytically, as follows: 
\begin{align}
    p(\vy_t |\vx_t,\data_{1:t-1})
     &= \int p_{\lin}(\vy_t|\vx_t,\vtheta_t)  p(\vtheta_t|\data_{1:t-1}) d\vtheta_t \\
     &= \int \gauss(\vy_t|\hat{h}_t(\vtheta_t), \vR_t) 
     \gauss(\vtheta_t|\vmu_{t|t-1}, \vSigma_{t|t-1}) d\vtheta_{t} 
\end{align}
Hence
\begin{align}
    \hat{\vy}_t &= \expect{\vy_t|\vx_t,\data_{1:t-1}}
      = h(\vx_t, \vmu_{t|t-1}) \\
\vV_{t} &= \cov{\vy_t | \vx_t, \data_{1:t-1}} 
= \vH_t \vSigma_{t|t-1} \vH_t^{\trans} + \vR_t
\label{eqn:pred-obs-var}
\end{align}
We can rewrite $\vV_t$
using Woodbury in a form that can be computed in $O(\nparams\memory^{2})$ time:
\begin{equation}
    \vV_{t}=
    \vH_{t}
    \left(\vUpsilon_{t|t-1}^{-1}
    -\vUpsilon_{t|t-1}^{-1}\vW_{t|t-1}
    \left(\vI_{\memory}+\vW_{t|t-1}^\trans\vUpsilon_{t|t-1}^{-1}\vW_{t|t-1}\right)^{-1}
    \vW_{t|t-1}^\trans\vUpsilon_{t|t-1}^{-1}\right)
    \vH_{t}^{\trans}+\vR_{t}
    \label{eqn:pred-obs-var-woodbury}
\end{equation}

\subsection{Deterministic approximation for classification}

In this section, we consider a classification model:
$h(\vx,\vtheta) = \softmax(f(\vx,\vtheta))$,
where $f$ is a neural network that outputs
a vector of $\nout$ logits.
Following \citep{Immer2021linear},
suppose we linearize $f$:
\begin{align}
    \hat{f}_t(\vtheta) =
    f(\vx_t, \vmu_{t|t-1}) + \vF_t (\vtheta-\vmu_{t|t-1})
\end{align}
 where $\vF_t$ is the Jacobian of $f(\vx_t,\cdot)$
 at $\vmu_{t|t-1}$.
 (This is the analog of $\hat{h}_t$ and $\vH_t$,
 except we omit the final softmax layer.)
 Let $\vz_t = \hat{f}_t(\vtheta)$ be the
 predicted logits.
We can now deterministically
approximate the predicted probabilities
by using the generalized probit 
approximation \citep{Gibbs1997,Daunizeau2017}:
\begin{align}
    \vp_t &= \int \softmax(\vz_t) 
    \gauss(\vz_t | \hat{\vz}_t, 
    \vF_t \vSigma_{t|t-1} \vF_t^\trans) d \vz_t \\
    &\approx \softmax\left(
    \left\{ \frac{\hat{z}_{t,c}}
    {\sqrt{1+\frac{\pi}{8} v_c}} \right\} \right)
\end{align}
where $v_c = [\vF_t \vSigma_{t|t-1} \vF_t^\trans]_{cc}$
is the marginal variance for class $c$.
This makes the probabilities ``less extreme''
(closer to uniform) when the parameters are uncertain.
Alternatively, we can use the 
``Laplace bridge'' method of
\citep{Hobbhahn2022},
which has been shown to be more accurate
than the generalized probit approximation.

\clearpage
\section{Tuning the hyper-parameters}
\label{sec:hparams}

In this section, we discuss how to estimate the SSM hyper-parameters,
namely the system noise $q$, the system dynamics $\gamma$,
and (for regression) the observation noise $\obsVar$. 
We also need to specify the initial belief state
$\vmu_0$ (which we sample from a zero-mean Gaussian prior)
and $\vSigma_0=(1/\eta_0) \vI$.

\subsection{Bayesian optimization}

We optimize the hyper-parameters using
black-box Bayesian optimization, using performance on a
validation set as the metric for static datasets,
and  the (averaged) one-step-ahead error  as the metric for non-stationary datasets.

\subsection{Online adaptation of the hyper-parameters}
\label{sec:adaptive-estimation}

Offline hyper-parameter tuning using
a validation set cannot be applied to non-stationary problems.
To tackle this, we can estimate  the SSM parameters
online; this approach is called 
adaptive Kalman filtering.
As a simple example, we implemented a
recursive estimate
for $\vR_t$,
based on a running average of the empirical prediction errors,
as proposed in \cite{Ljung1983} and \cite{Iiguni1992}:
\begin{align}
    \hat{\vR}_t &=
    (1-\lr_t) \hat{\vR}_{t-1} +
    \lr_t (\vy_t - \hat{\vy}_t)
    (\vy_t - \hat{\vy}_t)^\trans
\end{align}
where $\lr_t>0$ is a learning rate
(e.g., $\lr_t=\max(\lr_{\min}, 1/t)$),
and $\hat{\vy}_t = h(\vx_t,\vmu_{t|t-1})$.
If $\vR_t = r_t \vI$, this becomes
\begin{align}
    \hat{r}_t &=
    (1-\lr_t) \hat{r}_{t-1} +
    \lr_t (\vy_t - \hat{\vy}_t)^\trans
    (\vy_t - \hat{\vy}_t)
\end{align}

To estimate the other hyper-parameters, such as $Q$, in an online way,
we may be able to 
extend the variational Bayes approach of
\citep{Huang2020,viking},
or the gradient based method of \citep{Greenberg2021}.
However we leave this to future work.

\clearpage
\section{Additional experimental results}
\label{sec:extra}

\subsection{UCI regression}
\label{sec:UCI-app}

\begin{table}[h!]
\centering
\begin{tabular}{l|rrrrr}
& Num. features & Num. train & Num. test & Num. obs. & Num. parameters\\
 \noalign{\vskip1mm}
 \hline
 \noalign{\vskip-1mm}
 \noalign{\vskip 2mm}
Boston & 13 & 455 & 51 & 506 & 751 \\
Concrete & 8 & 927 & 103 & 1030 & 501 \\
Energy & 8 & 691 & 77 & 768 & 501 \\
Kin8nm & 8 & 7373 & 819 & 8192 & 501 \\
Naval & 16 & 10741 & 1193 & 11934 & 901 \\
Power & 4 & 8611 & 957 & 9568 & 301 \\
Wine & 11 & 1439 & 160 & 1599 & 651 \\
Yacht & 6 & 277 & 31 & 308 & 401 \\
\end{tabular}
\caption{UCI regression dataset summary,
and the corresponding number of parameters in a 
single-layered MLP with 50 hidden units.}
\label{tab:uci-regression-description}
\end{table}

\begin{figure*}
\centering
\begin{tabular}{ccc}
\includegraphics[height=2in]{figs/experiments/energy-rank10-test-set.pdf}
&
\includegraphics[height=2in]{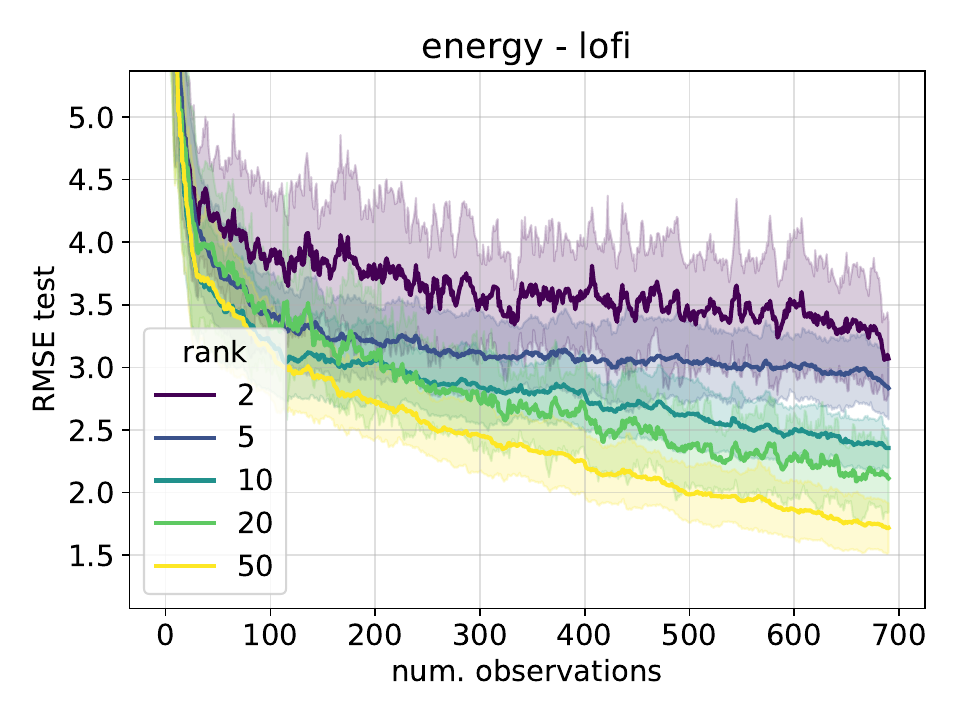}
\end{tabular}
\caption{
Error vs number of observations on the energy dataset.
We show the mean and standard error across 20 partitions.
  (a) Curves correspind to the following methods:
  for 
  FCEKF,
  FDEKF (similar to VDEKF),
  \LOFI-$10$,
  \LRVGA-$10$,
   \SGDRB-$10$.
   (b) Curves correspond to \LOFI with different ranks.
   \showdownfiggen{plots-xval.ipynb}.
}
\label{fig:energy-xval}
\end{figure*}


\begin{figure}[h!]
\centering
\includegraphics[width=0.9\linewidth]{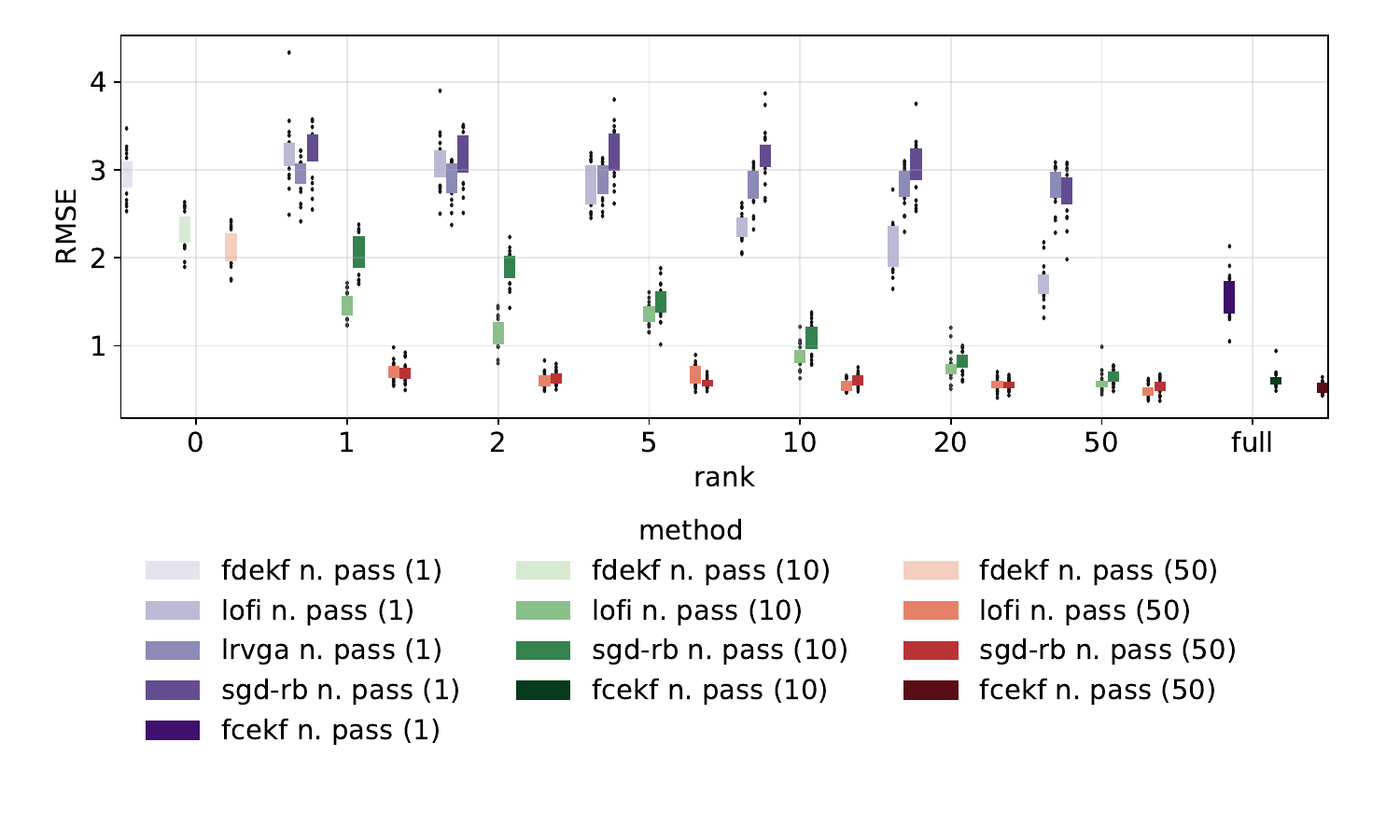}
\label{fig:1d-regression-prediction-final}
\caption{
RMSE boxplot for the energy dataset.
We compare the performance of different estimators
as a function of rank and 
 number of passes over the dataset.
 (Note that VDEKF is very similar to FDEKF so is not shown.)
\showdownfiggen{plots-xval-passes.ipynb}
}
\label{fig:energy-rank-passes}
\end{figure}

In this section, we evaluate various methods
on the UCI tabular regression benchmarks
used in  several other BNN papers
(e.g., \citep{HernandezLobato2015icml,Gal2016,Mishkin2018}).
We use the same splits as in \citep{Gal2016}.
As in these prior works,
we consider an MLP with 1 hidden layer of $H=50$ units
using RELU activation,
so the number of parameters is $\nparams = (D+2)H + 1$,
where $D$ is the number of input features.
In Table \ref{tab:uci-regression-description}, we show the number of features in each dataset, as well
as the number of training and testing examples in each of the 20 partitions.

\eat{
Note that the standard setting uses 100s of epochs (passes over the data), whereas we perform a single pass.
(We can emulate multiple passes by concatenating the data into a single long sequence; this improves performance of all methods,
but is not applicable in the non-stationary setting,
so we do not consider it further.)
}

In Figure \ref{fig:energy-xval}(a) we show the test error vs
number of training observations for different estimators
on the energy dataset.
We see that \LOFI (rank 10) outperforms \LRVGA (rank 10),
and both outperform diagonal EKF and SGD-RB (buffer size 10).
However, full covariance EKF is the most sample efficient learner.
In Figure \ref{fig:energy-xval}(b), we show that increasing the
rank of the \LOFI approximation improves performance;
by $\memory=50$ it has essentially
matched the full rank case, which uses $\nparams=501$ parameters.

\begin{figure}[h!]
\centering
\includegraphics[width=0.8\linewidth]{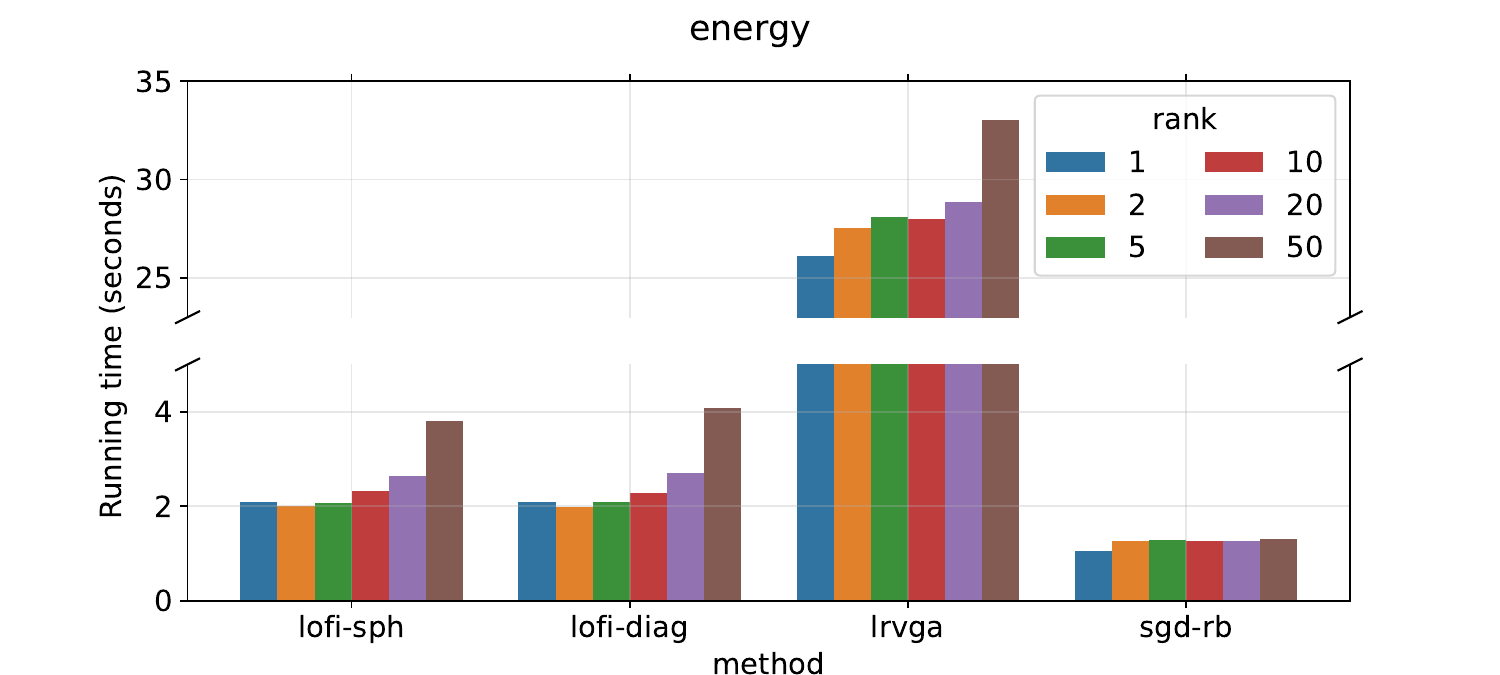}
\caption{
Running time (in seconds)
of a single pass over the Energy dataset for various low-rank methods.
\showdownfiggen{plots-xval-passes.ipynb}
}
\label{fig:energy-running-time}
\end{figure}

Another way to improve performance is to perform multiple passes over the data,
by concatenating the data sequence into a single long stream
(shuffling the order at the end of each epoch).
The benefits of this approach are shown in 
\cref{fig:energy-rank-passes}.
The different colors correspond to 1, 10 and 50 passes over the data.
(Note that we only performed one pass for \LRVGA,
since it is significantly slower than all other methods,
as shown in 
\cref{fig:energy-running-time}.)
We see that multiple passes consistently improves performance.
 However this trick can only be used in the offline setting for static distributions.
 In \cref{fig:energy-rank-passes}, we also see that the error vs rank decreases faster for \lofi than for \LRVGA and \SGDRB, meaning that it makes better use of its increased posterior accuracy to increase the sample efficiency of the learner.

Results for all the UCI regression datasets
for different methods 
are shown in  \cref{tab:uci-regression-rank10}.
\eat{
As a strong baseline, we include the results
from the SLANG paper \citep{Mishkin2018},
which uses a diagonal plus rank 1 approximation to the posterior precision matrix,
together with 120 passes over the data.
}
As in the energy dataset, we find that increasing the rank helps
all low-rank (and memory-based) methods,
and increasing the number of passes also helps.
In general FECKF is the best, with \LOFI usually in second place.
Interestingly we find that spherical \LOFI has comparable
performance to diagonal \LOFI, but is faster
(see \cref{tab:datasets-1pass-time} and
\cref{fig:energy-running-time}
for a running time comparison).
However, we caution against reading too many conclusions
from these results, since the datasets are small,
and the error bars  overlap
a lot between methods.

\begin{table}[h!]
    \centering
    \resizebox{\columnwidth}{!}{%
\begin{tabular}{lll|cccccccc}
\toprule
 &  & dataset & Boston & Concrete & Energy & Kin8nm & Naval & Power & Wine & Yacht \\
\# passes & Rank & Method &  &  &  &  &  &  &  &  \\
\midrule
\multirow[c]{7}{*}{1} & \multirow[c]{2}{*}{0} & fdekf & $5.23 \pm 2.19$ & $8.60 \pm 0.63$ & $2.96 \pm 0.25$ & $0.12 \pm 0.01$ & $0.01 \pm 0.00$ & $4.24 \pm 0.16$ & $0.82 \pm 0.05$ & $5.13 \pm 1.30$ \\
 &  & vdekf & $9.03 \pm 1.18$ & $16.35 \pm 0.82$ & $9.44 \pm 0.47$ & $0.14 \pm 0.01$ & $0.01 \pm 0.00$ & $4.25 \pm 0.16$ & $0.66 \pm 0.05$ & $5.60 \pm 1.29$ \\
  \cline{2-11}
 & \multirow[c]{4}{*}{10} & lofi-s & $5.12 \pm 1.49$ & $7.27 \pm 0.89$ & $2.36 \pm 0.16$ & $0.12 \pm 0.00$ & $0.00 \pm 0.00$ & $4.20 \pm 0.15$ & $0.65 \pm 0.03$ & $4.66 \pm 0.83$ \\
 &  & lofi-d & $4.77 \pm 1.20$ & $7.33 \pm 0.89$ & $2.53 \pm 0.26$ & $0.14 \pm 0.01$ & $0.00 \pm 0.00$ & $4.37 \pm 0.15$ & $0.72 \pm 0.06$ & $4.66 \pm 0.83$ \\
 &  & lrvga & $3.62 \pm 1.02$ & $7.28 \pm 0.73$ & $2.80 \pm 0.22$ & $0.12 \pm 0.00$ & $0.00 \pm 0.00$ & $4.22 \pm 0.15$ & $0.65 \pm 0.04$ & $3.39 \pm 0.79$ \\
 &  & sgd-rb & $4.41 \pm 1.23$ & $8.46 \pm 0.77$ & $3.18 \pm 0.30$ & $0.13 \pm 0.01$ & $0.00 \pm 0.00$ & $4.81 \pm 0.57$ & $0.70 \pm 0.06$ & $7.92 \pm 1.27$ \\
 \cline{2-11}
 & full & fcekf & $4.04 \pm 1.07$ & $6.45 \pm 0.53$ & $1.58 \pm 0.25$ & $0.10 \pm 0.00$ & $0.00 \pm 0.00$ & $4.13 \pm 0.16$ & $0.66 \pm 0.04$ & $3.14 \pm 1.09$ \\
 \hline \hline
\multirow[c]{6}{*}{10} & \multirow[c]{2}{*}{0} & fdekf & $3.20 \pm 0.92$ & $6.68 \pm 0.51$ & $2.32 \pm 0.22$ & $0.10 \pm 0.00$ & $0.01 \pm 0.00$ & $4.18 \pm 0.15$ & $0.82 \pm 0.05$ & $1.18 \pm 0.36$ \\
 &  & vdekf & $9.03 \pm 1.18$ & $16.35 \pm 0.82$ & $10.10 \pm 0.47$ & $0.11 \pm 0.00$ & $0.01 \pm 0.00$ & $4.20 \pm 0.16$ & $0.64 \pm 0.04$ & $2.32 \pm 0.54$ \\
 \cline{2-11}
 & \multirow[c]{3}{*}{10} & lofi-s & $5.38 \pm 1.36$ & $5.63 \pm 0.64$ & $0.88 \pm 0.14$ & $0.10 \pm 0.00$ & $0.00 \pm 0.00$ & $4.14 \pm 0.16$ & $0.64 \pm 0.04$ & $1.51 \pm 0.37$ \\
 &  & lofi-d & $5.08 \pm 1.29$ & $5.86 \pm 0.50$ & $1.36 \pm 0.19$ & $0.09 \pm 0.00$ & $0.00 \pm 0.00$ & $4.13 \pm 0.16$ & $0.64 \pm 0.04$ & $2.26 \pm 0.52$ \\
 &  & sgd-rb & $3.63 \pm 0.84$ & $6.29 \pm 0.68$ & $1.08 \pm 0.18$ & $0.10 \pm 0.01$ & $0.00 \pm 0.00$ & $4.73 \pm 0.38$ & $0.71 \pm 0.05$ & $2.26 \pm 0.56$ \\
 \cline{2-11}
 & full & fcekf & $3.13 \pm 0.89$ & $5.31 \pm 0.48$ & $0.62 \pm 0.09$ & $0.09 \pm 0.00$ & $0.00 \pm 0.00$ & $4.05 \pm 0.17$ & $0.64 \pm 0.05$ & $1.19 \pm 0.27$ \\
 \hline \hline
\multirow[c]{6}{*}{50} & \multirow[c]{2}{*}{0} & fdekf & $2.95 \pm 0.71$ & $6.37 \pm 0.52$ & $2.11 \pm 0.21$ & $0.09 \pm 0.00$ & $0.01 \pm 0.00$ & $4.14 \pm 0.16$ & $0.82 \pm 0.05$ & $0.80 \pm 0.26$ \\
 &  & vdekf & $9.03 \pm 1.18$ & $16.35 \pm 0.82$ & $10.10 \pm 0.47$ & $0.10 \pm 0.00$ & $0.01 \pm 0.00$ & $4.17 \pm 0.16$ & $0.63 \pm 0.04$ & $1.62 \pm 0.37$ \\
 \cline{2-11}
 & \multirow[c]{3}{*}{10} & lofi-s & $5.29 \pm 1.12$ & $5.41 \pm 0.64$ & $0.56 \pm 0.07$ & $0.09 \pm 0.00$ & $0.00 \pm 0.00$ & $4.06 \pm 0.17$ & $0.66 \pm 0.05$ & $0.92 \pm 0.27$ \\
 &  & lofi-d & $4.99 \pm 1.10$ & $5.53 \pm 0.50$ & $0.86 \pm 0.14$ & $0.09 \pm 0.00$ & $0.00 \pm 0.00$ & $4.10 \pm 0.16$ & $0.63 \pm 0.04$ & $1.36 \pm 0.33$ \\
 &  & sgd-rb & $3.52 \pm 0.68$ & $5.78 \pm 0.87$ & $0.60 \pm 0.07$ & $0.10 \pm 0.01$ & $0.00 \pm 0.00$ & $4.74 \pm 0.38$ & $0.79 \pm 0.08$ & $0.81 \pm 0.25$ \\
 \cline{2-11}
 & full & fcekf & $3.62 \pm 1.28$ & $5.12 \pm 0.59$ & $0.52 \pm 0.06$ & $0.09 \pm 0.00$ & $0.00 \pm 0.00$ & $4.00 \pm 0.17$ & $0.68 \pm 0.06$ & $1.12 \pm 0.29$ \\
\bottomrule
\end{tabular}
    }
    \caption{RMSE on UCI regression datasets. We report mean and standard
    error of the mean across 20 splits of the data.
    lofi-s is \lofi spherical, and lofi-d is \lofi diagonal;
    \LOFI and
    \LRVGA use a rank 10 approximation to the posterior
    precision matrix,
    whereas SGD-RB uses a replay buffer with 10 examples.
    }
    \label{tab:uci-regression-rank10}
\end{table}

\begin{table}[h!]
    \centering
\resizebox{\columnwidth}{!}{
\begin{tabular}{llllllllll}
 & dataset & Boston & Concrete & Energy & Kin8nm & Naval & Power & Wine & Yacht \\
rank & variable &  &  &  &  &  &  &  &  \\
    \noalign{\vskip1mm}
    \hline
    \noalign{\vskip-1mm}
    \noalign{\vskip 2mm}
0 & fdekf & $5.23 \pm 2.19$ & $8.60 \pm 0.63$ & $2.96 \pm 0.25$ & $0.12 \pm 0.01$ & $0.01 \pm 0.00$ & $4.24 \pm 0.16$ & $0.82 \pm 0.05$ & $5.13 \pm 1.30$ \\
 \hline
 \hline
\multirow[c]{4}{*}{1} & lofi-sph & $5.08 \pm 1.27$ & $8.84 \pm 1.23$ & $3.21 \pm 0.36$ & $0.14 \pm 0.01$ & $0.01 \pm 0.00$ & $4.36 \pm 0.15$ & $0.67 \pm 0.05$ & $5.76 \pm 1.52$ \\
  \cline{2-10}
 & lofi-diag & $5.08 \pm 1.27$ & $9.12 \pm 1.35$ & $3.50 \pm 0.48$ & $0.14 \pm 0.01$ & $0.01 \pm 0.00$ & $5.01 \pm 0.47$ & $0.69 \pm 0.06$ & $5.91 \pm 1.52$ \\
  \cline{2-10}
 & lrvga & $4.14 \pm 1.03$ & $7.45 \pm 0.75$ & $2.92 \pm 0.22$ & $0.14 \pm 0.01$ & - & $4.25 \pm 0.15$ & $0.65 \pm 0.04$ & $5.06 \pm 1.06$ \\
  \cline{2-10}
 & sgd-rb & $4.44 \pm 1.20$ & $9.62 \pm 0.63$ & $3.19 \pm 0.30$ & $0.16 \pm 0.01$ & $0.01 \pm 0.00$ & $4.41 \pm 0.18$ & $0.66 \pm 0.04$ & $9.84 \pm 1.94$ \\
 \hline
 \hline
\multirow[c]{4}{*}{2} & lofi-sph & $4.38 \pm 1.10$ & $8.17 \pm 0.90$ & $3.07 \pm 0.30$ & $0.26 \pm 0.02$ & $0.00 \pm 0.00$ & $4.33 \pm 0.16$ & $0.66 \pm 0.04$ & $5.98 \pm 1.45$ \\
  \cline{2-10}
 & lofi-diag & $5.00 \pm 1.71$ & $8.54 \pm 1.15$ & $3.34 \pm 0.45$ & $0.15 \pm 0.01$ & $0.00 \pm 0.00$ & $4.59 \pm 0.28$ & $0.73 \pm 0.07$ & $5.65 \pm 1.30$ \\
  \cline{2-10}
 & lrvga & $3.88 \pm 1.03$ & $7.41 \pm 0.90$ & $2.87 \pm 0.22$ & $0.14 \pm 0.01$ & $0.00 \pm 0.00$ & $4.24 \pm 0.14$ & $0.65 \pm 0.04$ & $4.23 \pm 0.91$ \\
  \cline{2-10}
 & sgd-rb & $4.31 \pm 1.20$ & $9.13 \pm 0.63$ & $3.16 \pm 0.30$ & $0.15 \pm 0.01$ & $0.01 \pm 0.00$ & $4.50 \pm 0.24$ & $0.67 \pm 0.05$ & $9.03 \pm 1.64$ \\
 \hline
 \hline
\multirow[c]{4}{*}{5} & lofi-sph & $4.10 \pm 1.13$ & $7.77 \pm 1.02$ & $2.83 \pm 0.26$ & $0.15 \pm 0.01$ & $0.00 \pm 0.00$ & $4.24 \pm 0.15$ & $0.65 \pm 0.04$ & $5.51 \pm 1.22$ \\
  \cline{2-10}
 & lofi-diag & $4.75 \pm 1.30$ & $8.46 \pm 1.37$ & $2.87 \pm 0.34$ & $0.13 \pm 0.01$ & $0.00 \pm 0.00$ & $4.56 \pm 0.20$ & $0.74 \pm 0.07$ & $4.30 \pm 0.88$ \\
  \cline{2-10}
 & lrvga & $3.71 \pm 1.08$ & $6.98 \pm 0.57$ & $2.86 \pm 0.21$ & $0.13 \pm 0.00$ & $0.00 \pm 0.00$ & $4.23 \pm 0.15$ & $0.65 \pm 0.04$ & $3.67 \pm 0.84$ \\
  \cline{2-10}
 & sgd-rb & $4.29 \pm 1.20$ & $8.72 \pm 0.72$ & $3.18 \pm 0.30$ & $0.14 \pm 0.01$ & $0.01 \pm 0.00$ & $4.72 \pm 0.56$ & $0.68 \pm 0.05$ & $8.36 \pm 1.36$ \\
 \hline
 \hline
\multirow[c]{4}{*}{10} & lofi-sph & $5.12 \pm 1.49$ & $7.27 \pm 0.89$ & $2.36 \pm 0.16$ & $0.12 \pm 0.00$ & $0.00 \pm 0.00$ & $4.20 \pm 0.15$ & $0.65 \pm 0.03$ & $4.66 \pm 0.83$ \\
  \cline{2-10}
 & lofi-diag & $4.77 \pm 1.20$ & $7.33 \pm 0.89$ & $2.53 \pm 0.26$ & $0.14 \pm 0.01$ & $0.00 \pm 0.00$ & $4.37 \pm 0.15$ & $0.72 \pm 0.06$ & $4.66 \pm 0.83$ \\
  \cline{2-10}
 & lrvga & $3.62 \pm 1.02$ & $7.28 \pm 0.73$ & $2.80 \pm 0.22$ & $0.12 \pm 0.00$ & $0.00 \pm 0.00$ & $4.22 \pm 0.15$ & $0.65 \pm 0.04$ & $3.39 \pm 0.79$ \\
  \cline{2-10}
 & sgd-rb & $4.41 \pm 1.23$ & $8.46 \pm 0.77$ & $3.18 \pm 0.30$ & $0.13 \pm 0.01$ & $0.00 \pm 0.00$ & $4.81 \pm 0.57$ & $0.70 \pm 0.06$ & $7.92 \pm 1.27$ \\
 \hline
 \hline
\multirow[c]{4}{*}{20} & lofi-sph & $4.88 \pm 1.49$ & $6.92 \pm 0.60$ & $2.11 \pm 0.28$ & $0.11 \pm 0.01$ & $0.00 \pm 0.00$ & $4.23 \pm 0.15$ & $0.65 \pm 0.03$ & $4.73 \pm 0.99$ \\
  \cline{2-10}
 & lofi-diag & $4.88 \pm 1.49$ & $8.03 \pm 1.25$ & $2.16 \pm 0.27$ & $0.14 \pm 0.01$ & $0.00 \pm 0.00$ & $4.41 \pm 0.18$ & $0.66 \pm 0.04$ & $2.37 \pm 0.63$ \\
  \cline{2-10}
 & lrvga & $3.57 \pm 1.07$ & $6.73 \pm 0.60$ & $2.80 \pm 0.22$ & $0.11 \pm 0.00$ & $0.00 \pm 0.00$ & $4.24 \pm 0.16$ & $0.64 \pm 0.04$ & $2.76 \pm 1.08$ \\
  \cline{2-10}
 & sgd-rb & $4.39 \pm 1.18$ & $8.26 \pm 0.95$ & $3.04 \pm 0.31$ & $0.12 \pm 0.01$ & $0.00 \pm 0.00$ & $4.77 \pm 0.32$ & $0.72 \pm 0.06$ & $7.42 \pm 1.24$ \\
 \hline
 \hline
\multirow[c]{4}{*}{50} & lofi-sph & $4.84 \pm 1.39$ & $6.65 \pm 0.54$ & $1.72 \pm 0.20$ & $0.10 \pm 0.00$ & $0.02 \pm 0.00$ & $4.20 \pm 0.14$ & $0.69 \pm 0.05$ & $2.31 \pm 0.54$ \\
  \cline{2-10}
 & lofi-diag & $4.84 \pm 1.39$ & $6.70 \pm 0.50$ & $1.84 \pm 0.29$ & $0.11 \pm 0.00$ & $0.00 \pm 0.00$ & $4.30 \pm 0.15$ & $0.64 \pm 0.04$ & $4.85 \pm 0.98$ \\
  \cline{2-10}
 & lrvga & $3.52 \pm 1.05$ & $6.70 \pm 0.58$ & $2.79 \pm 0.22$ & $0.11 \pm 0.00$ & $0.00 \pm 0.00$ & $4.21 \pm 0.15$ & $0.64 \pm 0.04$ & $3.33 \pm 0.81$ \\
 & sgd-rb & $4.19 \pm 1.18$ & $7.71 \pm 0.88$ & $2.73 \pm 0.28$ & $0.12 \pm 0.01$ & $0.00 \pm 0.00$ & $4.81 \pm 0.24$ & $0.76 \pm 0.05$ & $6.62 \pm 1.19$ \\
 \hline
 \hline
full & fcekf & $4.04 \pm 1.07$ & $6.45 \pm 0.53$ & $1.58 \pm 0.25$ & $0.10 \pm 0.00$ & $0.00 \pm 0.00$ & $4.13 \pm 0.16$ & $0.66 \pm 0.04$ & $3.14 \pm 1.09$ \\
\end{tabular}
}
    \caption{RMSE for datasets as a function of method, rank after a single pass over the dataset.}
    \label{tab:datasets-rank-passes-1}
\end{table}

\begin{table}[h!]
    \centering
\resizebox{\columnwidth}{!}{
\begin{tabular}{llllllllll}
 & dataset & Boston & Concrete & Energy & Kin8nm & Naval & Power & Wine & Yacht \\
rank & variable &  &  &  &  &  &  &  &  \\
    \noalign{\vskip1mm}
    \hline
    \noalign{\vskip-1mm}
    \noalign{\vskip 2mm}
0 & fdekf & $3.20 \pm 0.92$ & $6.68 \pm 0.51$ & $2.32 \pm 0.22$ & $0.10 \pm 0.00$ & $0.01 \pm 0.00$ & $4.18 \pm 0.15$ & $0.82 \pm 0.05$ & $1.18 \pm 0.36$ \\
 \hline
 \hline
\multirow[c]{3}{*}{1} & lofi-sph & $5.60 \pm 1.43$ & $6.35 \pm 0.71$ & $1.47 \pm 0.15$ & $0.10 \pm 0.01$ & $0.00 \pm 0.00$ & $4.27 \pm 0.17$ & $0.66 \pm 0.04$ & $2.12 \pm 0.52$ \\
  \cline{2-10}
 & lofi-diag & $5.21 \pm 1.44$ & $6.24 \pm 0.53$ & $2.22 \pm 0.21$ & $0.11 \pm 0.00$ & $0.01 \pm 0.00$ & $4.17 \pm 0.15$ & $0.64 \pm 0.04$ & $1.76 \pm 0.43$ \\
  \cline{2-10}
 & sgd-rb & $3.47 \pm 0.98$ & $6.57 \pm 0.47$ & $2.04 \pm 0.22$ & $0.09 \pm 0.00$ & $0.00 \pm 0.00$ & $4.23 \pm 0.20$ & $0.65 \pm 0.04$ & $4.82 \pm 0.81$ \\
 \hline
 \hline
\multirow[c]{3}{*}{2} & lofi-sph & $3.51 \pm 0.94$ & $6.23 \pm 0.64$ & $1.16 \pm 0.18$ & $0.31 \pm 0.05$ & $0.00 \pm 0.00$ & $4.20 \pm 0.15$ & $0.65 \pm 0.04$ & $2.49 \pm 0.51$ \\
  \cline{2-10}
 & lofi-diag & $5.08 \pm 1.43$ & $6.19 \pm 0.50$ & $1.97 \pm 0.22$ & $0.10 \pm 0.00$ & $0.00 \pm 0.00$ & $4.17 \pm 0.15$ & $0.63 \pm 0.04$ & $1.75 \pm 0.49$ \\
  \cline{2-10}
 & sgd-rb & $3.50 \pm 0.97$ & $6.41 \pm 0.53$ & $1.86 \pm 0.19$ & $0.09 \pm 0.00$ & $0.00 \pm 0.00$ & $4.27 \pm 0.22$ & $0.65 \pm 0.04$ & $4.31 \pm 0.70$ \\
 \hline
 \hline
\multirow[c]{3}{*}{5} & lofi-sph & $3.47 \pm 1.00$ & $6.02 \pm 0.50$ & $1.36 \pm 0.13$ & $0.14 \pm 0.02$ & $0.00 \pm 0.00$ & $4.17 \pm 0.14$ & $0.65 \pm 0.04$ & $2.44 \pm 0.53$ \\
  \cline{2-10}
 & lofi-diag & $4.95 \pm 1.31$ & $5.74 \pm 0.48$ & $1.57 \pm 0.19$ & $0.10 \pm 0.00$ & $0.00 \pm 0.00$ & $4.14 \pm 0.15$ & $0.63 \pm 0.04$ & $1.40 \pm 0.39$ \\
  \cline{2-10}
 & sgd-rb & $3.60 \pm 0.87$ & $6.28 \pm 0.61$ & $1.51 \pm 0.20$ & $0.10 \pm 0.01$ & $0.00 \pm 0.00$ & $4.45 \pm 0.34$ & $0.68 \pm 0.05$ & $3.40 \pm 0.61$ \\
 \hline
 \hline
\multirow[c]{3}{*}{10} & lofi-sph & $5.38 \pm 1.36$ & $5.63 \pm 0.64$ & $0.88 \pm 0.14$ & $0.10 \pm 0.00$ & $0.00 \pm 0.00$ & $4.14 \pm 0.16$ & $0.64 \pm 0.04$ & $1.51 \pm 0.37$ \\
  \cline{2-10}
 & lofi-diag & $5.08 \pm 1.29$ & $5.86 \pm 0.50$ & $1.36 \pm 0.19$ & $0.09 \pm 0.00$ & $0.00 \pm 0.00$ & $4.13 \pm 0.16$ & $0.64 \pm 0.04$ & $2.26 \pm 0.52$ \\
  \cline{2-10}
 & sgd-rb & $3.63 \pm 0.84$ & $6.29 \pm 0.68$ & $1.08 \pm 0.18$ & $0.10 \pm 0.01$ & $0.00 \pm 0.00$ & $4.73 \pm 0.38$ & $0.71 \pm 0.05$ & $2.26 \pm 0.56$ \\
 \hline
 \hline
\multirow[c]{3}{*}{20} & lofi-sph & $5.14 \pm 1.35$ & $5.47 \pm 0.67$ & $0.75 \pm 0.17$ & $0.09 \pm 0.00$ & $0.00 \pm 0.00$ & $4.18 \pm 0.17$ & $0.64 \pm 0.04$ & $1.75 \pm 0.42$ \\
  \cline{2-10}
 & lofi-diag & $5.17 \pm 1.34$ & $5.54 \pm 0.49$ & $0.92 \pm 0.19$ & $0.09 \pm 0.00$ & $0.00 \pm 0.00$ & $4.10 \pm 0.16$ & $0.63 \pm 0.04$ & $1.23 \pm 0.28$ \\
  \cline{2-10}
 & sgd-rb & $3.60 \pm 0.98$ & $6.08 \pm 0.73$ & $0.83 \pm 0.12$ & $0.10 \pm 0.01$ & $0.00 \pm 0.00$ & $4.80 \pm 0.33$ & $0.77 \pm 0.07$ & $1.32 \pm 0.40$ \\
 \hline
 \hline
\multirow[c]{3}{*}{50} & lofi-sph & $5.18 \pm 1.39$ & $5.35 \pm 0.52$ & $0.59 \pm 0.12$ & $0.09 \pm 0.00$ & $0.02 \pm 0.00$ & $4.12 \pm 0.17$ & $0.66 \pm 0.05$ & $1.03 \pm 0.32$ \\
  \cline{2-10}
 & lofi-diag & $5.20 \pm 1.37$ & $5.54 \pm 0.52$ & $0.70 \pm 0.12$ & $0.09 \pm 0.00$ & $0.00 \pm 0.00$ & $4.08 \pm 0.17$ & $0.64 \pm 0.04$ & $2.30 \pm 0.46$ \\
  \cline{2-10}
 & sgd-rb & $3.70 \pm 1.05$ & $5.76 \pm 0.85$ & $0.64 \pm 0.08$ & $0.11 \pm 0.01$ & $0.00 \pm 0.00$ & $4.96 \pm 0.26$ & $0.83 \pm 0.09$ & $0.88 \pm 0.29$ \\
 \hline
 \hline
full & fcekf & $3.13 \pm 0.89$ & $5.31 \pm 0.48$ & $0.62 \pm 0.09$ & $0.09 \pm 0.00$ & $0.00 \pm 0.00$ & $4.05 \pm 0.17$ & $0.64 \pm 0.05$ & $1.19 \pm 0.27$ \\
\end{tabular}
}
    \caption{RMSE for datasets as a function of method, rank after 10 passes over the dataset.}
    \label{tab:datasets-rank-passes-10}
\end{table}

\begin{table}[h!]
    \centering
\resizebox{\columnwidth}{!}{
\begin{tabular}{llllllllll}
 & dataset & Boston & Concrete & Energy & Kin8nm & Naval & Power & Wine & Yacht \\
rank & variable &  &  &  &  &  &  &  &  \\
    \noalign{\vskip1mm}
    \hline
    \noalign{\vskip-1mm}
    \noalign{\vskip 2mm}
0 & fdekf & $2.95 \pm 0.71$ & $6.37 \pm 0.52$ & $2.11 \pm 0.21$ & $0.09 \pm 0.00$ & $0.01 \pm 0.00$ & $4.14 \pm 0.16$ & $0.82 \pm 0.05$ & $0.80 \pm 0.26$ \\
 \hline
 \hline
\multirow[c]{3}{*}{1} & lofi-sph & $5.70 \pm 1.28$ & $5.89 \pm 0.90$ & $0.71 \pm 0.11$ & $0.09 \pm 0.00$ & $0.00 \pm 0.00$ & $4.15 \pm 0.16$ & $0.66 \pm 0.05$ & $0.96 \pm 0.28$ \\
  \cline{2-10}
 & lofi-diag & $5.48 \pm 1.17$ & $5.88 \pm 0.47$ & $1.96 \pm 0.20$ & $0.10 \pm 0.00$ & $0.00 \pm 0.00$ & $4.15 \pm 0.16$ & $0.63 \pm 0.04$ & $1.19 \pm 0.28$ \\
  \cline{2-10}
 & sgd-rb & $3.28 \pm 0.85$ & $5.70 \pm 0.76$ & $0.69 \pm 0.11$ & $0.08 \pm 0.00$ & $0.00 \pm 0.00$ & $4.13 \pm 0.20$ & $0.66 \pm 0.05$ & $1.33 \pm 0.35$ \\
 \hline
 \hline
\multirow[c]{3}{*}{2} & lofi-sph & $3.26 \pm 0.85$ & $5.75 \pm 0.74$ & $0.61 \pm 0.09$ & $0.29 \pm 0.05$ & $0.00 \pm 0.00$ & $4.13 \pm 0.15$ & $0.66 \pm 0.05$ & $1.06 \pm 0.28$ \\
  \cline{2-10}
 & lofi-diag & $5.13 \pm 1.10$ & $5.81 \pm 0.47$ & $1.68 \pm 0.19$ & $0.09 \pm 0.00$ & $0.00 \pm 0.00$ & $4.15 \pm 0.16$ & $0.63 \pm 0.04$ & $1.30 \pm 0.38$ \\
  \cline{2-10}
 & sgd-rb & $3.27 \pm 0.83$ & $5.74 \pm 0.81$ & $0.64 \pm 0.08$ & $0.09 \pm 0.00$ & $0.00 \pm 0.00$ & $4.20 \pm 0.24$ & $0.69 \pm 0.06$ & $1.13 \pm 0.37$ \\
 \hline
 \hline
\multirow[c]{3}{*}{5} & lofi-sph & $3.10 \pm 0.84$ & $5.82 \pm 0.75$ & $0.67 \pm 0.12$ & $0.19 \pm 0.09$ & $0.00 \pm 0.00$ & $4.13 \pm 0.15$ & $0.65 \pm 0.05$ & $1.05 \pm 0.23$ \\
  \cline{2-10}
 & lofi-diag & $4.92 \pm 1.13$ & $5.44 \pm 0.46$ & $1.17 \pm 0.17$ & $0.10 \pm 0.00$ & $0.00 \pm 0.00$ & $4.10 \pm 0.17$ & $0.63 \pm 0.04$ & $1.08 \pm 0.29$ \\
  \cline{2-10}
 & sgd-rb & $3.38 \pm 0.77$ & $6.04 \pm 0.87$ & $0.58 \pm 0.06$ & $0.09 \pm 0.01$ & $0.00 \pm 0.00$ & $4.36 \pm 0.30$ & $0.73 \pm 0.07$ & $0.95 \pm 0.32$ \\
 \hline
 \hline
\multirow[c]{3}{*}{10} & lofi-sph & $5.29 \pm 1.12$ & $5.41 \pm 0.64$ & $0.56 \pm 0.07$ & $0.09 \pm 0.00$ & $0.00 \pm 0.00$ & $4.06 \pm 0.17$ & $0.66 \pm 0.05$ & $0.92 \pm 0.27$ \\
  \cline{2-10}
 & lofi-diag & $4.99 \pm 1.10$ & $5.53 \pm 0.50$ & $0.86 \pm 0.14$ & $0.09 \pm 0.00$ & $0.00 \pm 0.00$ & $4.10 \pm 0.16$ & $0.63 \pm 0.04$ & $1.36 \pm 0.33$ \\
  \cline{2-10}
 & sgd-rb & $3.52 \pm 0.68$ & $5.78 \pm 0.87$ & $0.60 \pm 0.07$ & $0.10 \pm 0.01$ & $0.00 \pm 0.00$ & $4.74 \pm 0.38$ & $0.79 \pm 0.08$ & $0.81 \pm 0.25$ \\
 \hline
 \hline
\multirow[c]{3}{*}{20} & lofi-sph & $5.01 \pm 1.09$ & $5.14 \pm 0.69$ & $0.56 \pm 0.07$ & $0.09 \pm 0.00$ & $0.00 \pm 0.00$ & $4.12 \pm 0.19$ & $0.67 \pm 0.04$ & $1.17 \pm 0.23$ \\
  \cline{2-10}
 & lofi-diag & $5.01 \pm 1.10$ & $5.35 \pm 0.45$ & $0.67 \pm 0.15$ & $0.09 \pm 0.00$ & $0.00 \pm 0.00$ & $4.06 \pm 0.16$ & $0.63 \pm 0.04$ & $1.03 \pm 0.28$ \\
  \cline{2-10}
 & sgd-rb & $3.76 \pm 0.74$ & $5.86 \pm 0.83$ & $0.56 \pm 0.06$ & $0.10 \pm 0.01$ & $0.00 \pm 0.00$ & $4.89 \pm 0.54$ & $0.85 \pm 0.08$ & $0.78 \pm 0.26$ \\
 \hline
 \hline
\multirow[c]{3}{*}{50} & lofi-sph & $5.00 \pm 1.12$ & $5.09 \pm 0.66$ & $0.48 \pm 0.08$ & $0.08 \pm 0.00$ & $0.02 \pm 0.00$ & $4.05 \pm 0.17$ & $0.68 \pm 0.06$ & $0.93 \pm 0.20$ \\
  \cline{2-10}
 & lofi-diag & $5.01 \pm 1.11$ & $5.27 \pm 0.58$ & $0.57 \pm 0.08$ & $0.09 \pm 0.00$ & $0.00 \pm 0.00$ & $4.05 \pm 0.17$ & $0.65 \pm 0.04$ & $1.38 \pm 0.25$ \\
  \cline{2-10}
 & sgd-rb & $4.05 \pm 1.02$ & $5.81 \pm 0.65$ & $0.53 \pm 0.08$ & $0.10 \pm 0.00$ & $0.00 \pm 0.00$ & $4.73 \pm 0.35$ & $0.94 \pm 0.11$ & $0.71 \pm 0.38$ \\
 \hline
 \hline
full & fcekf & $3.62 \pm 1.28$ & $5.12 \pm 0.59$ & $0.52 \pm 0.06$ & $0.09 \pm 0.00$ & $0.00 \pm 0.00$ & $4.00 \pm 0.17$ & $0.68 \pm 0.06$ & $1.12 \pm 0.29$ \\
\end{tabular}
}
    \caption{RMSE for datasets as a function of method, rank after 50 passes over the dataset.}
    \label{tab:datasets-rank-passes-50}
\end{table}

\begin{table}[h!]
    \centering
\begin{tabular}{lccccccccc}
 & & boston & concrete & energy & kin8nm & naval & power & wine & yacht \\
rank & &  &  &  &  &  &  &  &  \\
    \noalign{\vskip1mm}
    \hline
    \noalign{\vskip-1mm}
    \noalign{\vskip 2mm}
\multirow[c]{4}{*}{1} & lofi-sph & $1.93$ & $2.17$ & $2.10$ & $4.31$ & $5.39$ & $4.60$ & $2.40$ & $1.96$ \\
  \cline{2-10}
 & lofi-diag & $2.12$ & $2.15$ & $2.08$ & $4.31$ & $6.38$ & $4.61$ & $2.40$ & $1.97$ \\
  \cline{2-10}
 & lrvga & $31.31$ & $31.44$ & $26.14$ & $194.52$ & $819.99$ & $125.78$ & $69.74$ & $13.88$ \\
  \cline{2-10}
 & sgd-rb & $1.40$ & $1.04$ & $1.05$ & $1.58$ & $1.86$ & $1.58$ & $1.15$ & $1.02$ \\
 \hline
 \hline
\multirow[c]{4}{*}{2} & lofi-sph & $1.88$ & $2.07$ & $2.00$ & $4.38$ & $5.47$ & $4.70$ & $2.28$ & $1.84$ \\
  \cline{2-10}
 & lofi-diag & $1.91$ & $2.04$ & $1.98$ & $4.34$ & $5.43$ & $4.68$ & $2.27$ & $1.85$ \\
  \cline{2-10}
 & lrvga & $32.83$ & $33.90$ & $27.53$ & $215.47$ & $884.71$ & $145.94$ & $75.78$ & $14.30$ \\
  \cline{2-10}
 & sgd-rb & $1.26$ & $1.31$ & $1.25$ & $2.30$ & $2.77$ & $2.44$ & $2.92$ & $1.19$ \\
 \hline
 \hline
\multirow[c]{4}{*}{5} & lofi-sph & $1.92$ & $2.16$ & $2.07$ & $4.95$ & $6.26$ & $5.34$ & $2.41$ & $3.44$ \\
  \cline{2-10}
 & lofi-diag & $1.91$ & $2.66$ & $2.08$ & $4.95$ & $6.28$ & $5.34$ & $2.43$ & $1.92$ \\
  \cline{2-10}
 & lrvga & $33.06$ & $34.19$ & $28.09$ & $219.50$ & $892.29$ & $149.50$ & $76.85$ & $14.43$ \\
  \cline{2-10}
 & sgd-rb & $1.26$ & $1.28$ & $1.28$ & $2.27$ & $2.81$ & $2.39$ & $1.42$ & $1.20$ \\
 \hline
 \hline
\multirow[c]{4}{*}{10} & lofi-sph & $2.13$ & $2.89$ & $2.31$ & $6.17$ & $8.51$ & $6.65$ & $2.66$ & $2.00$ \\
  \cline{2-10}
 & lofi-diag & $2.13$ & $2.40$ & $2.28$ & $6.40$ & $8.56$ & $7.95$ & $2.67$ & $1.99$ \\
  \cline{2-10}
 & lrvga & $32.99$ & $33.91$ & $28.01$ & $218.10$ & $888.80$ & $151.09$ & $75.00$ & $14.37$ \\
  \cline{2-10}
 & sgd-rb & $1.25$ & $1.26$ & $1.27$ & $2.32$ & $2.93$ & $2.41$ & $2.95$ & $1.19$ \\
 \hline
 \hline
\multirow[c]{4}{*}{20} & lofi-sph & $2.31$ & $2.74$ & $2.64$ & $8.68$ & $12.36$ & $10.29$ & $3.28$ & $2.19$ \\
  \cline{2-10}
 & lofi-diag & $2.31$ & $2.74$ & $2.70$ & $9.34$ & $12.30$ & $10.30$ & $3.25$ & $3.89$ \\
  \cline{2-10}
 & lrvga & $34.19$ & $35.90$ & $28.84$ & $234.75$ & $910.09$ & $169.94$ & $77.88$ & $14.92$ \\
  \cline{2-10}
 & sgd-rb & $1.25$ & $1.27$ & $1.26$ & $2.38$ & $2.89$ & $2.49$ & $1.46$ & $1.22$ \\
 \hline
 \hline
\multirow[c]{4}{*}{50} & lofi-sph & $3.24$ & $4.42$ & $3.81$ & $19.59$ & $39.47$ & $21.43$ & $5.58$ & $2.75$ \\
  \cline{2-10}
 & lofi-diag & $3.44$ & $4.64$ & $4.08$ & $19.56$ & $26.90$ & $21.50$ & $6.03$ & $2.80$ \\
  \cline{2-10}
 & lrvga & $36.65$ & $41.84$ & $33.04$ & $280.21$ & $988.45$ & $222.46$ & $88.81$ & $16.80$ \\
  \cline{2-10}
 & sgd-rb & $1.25$ & $1.35$ & $1.31$ & $2.77$ & $3.52$ & $2.97$ & $1.52$ & $1.24$ \\
 \hline
 \hline
full & fcekf & $1.34$ & $1.69$ & $1.24$ & $2.56$ & $5.98$ & $2.34$ & $1.61$ & $1.15$ \\
\end{tabular}
    \caption{Running time (in seconds) for benchmarked methods after a single pass over the UCI datasets.}
    \label{tab:datasets-1pass-time}
\end{table}

\clearpage

\subsection{Piecewise stationary 1d regression}


\begin{figure}[h!]
\centering
\includegraphics[width=0.9\linewidth]{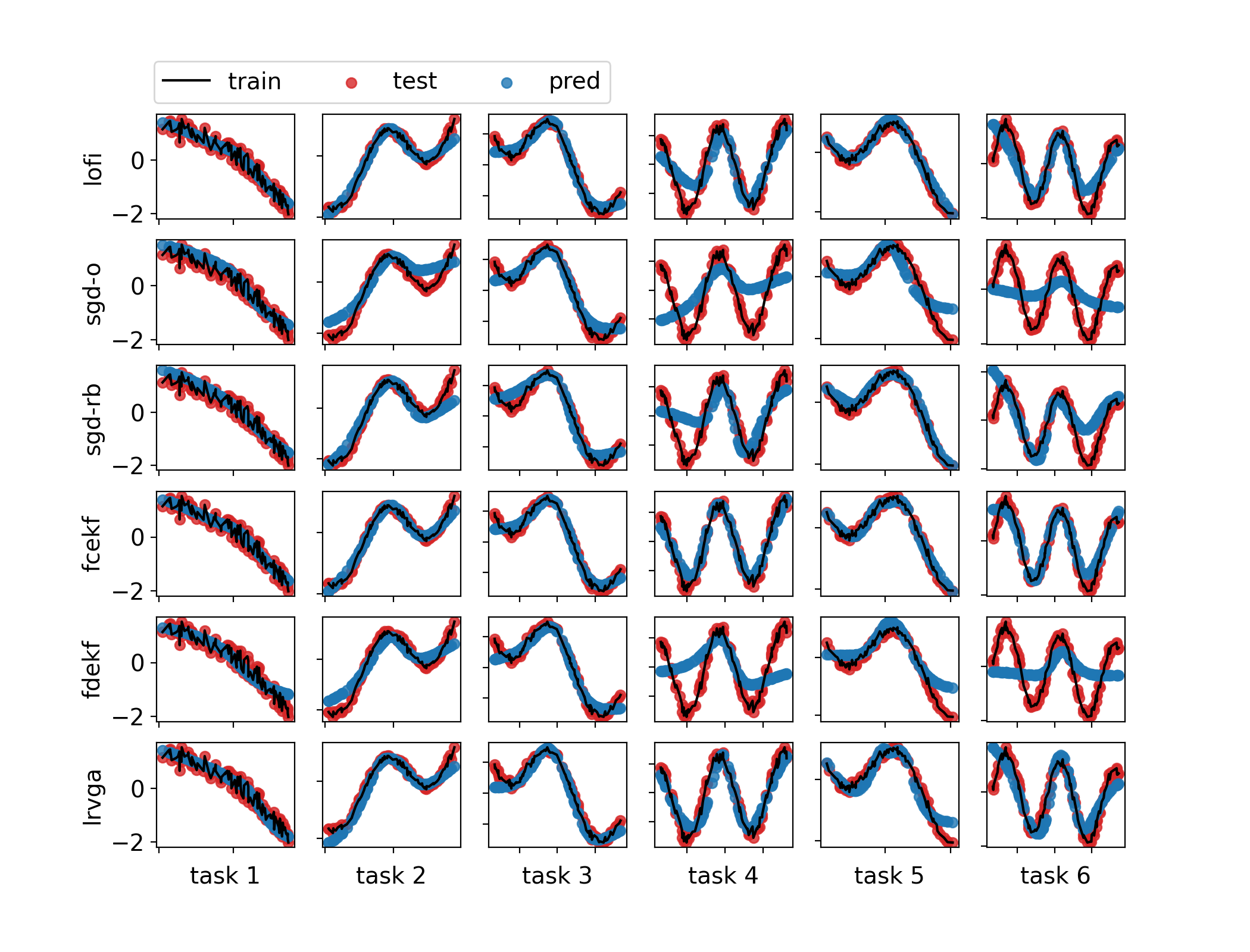}
\caption{
  Results for piecewise stationary 1d regression.
  Red dots are from  the true function for each task,
  and the blue dots are the predictions of the model at the end of
  each task (after training on 200 examples).
\demosfiggen{nonstat-1d-regression.ipynb}
}
\label{fig:regression-1d-functions}
\end{figure}

\begin{figure}[h!]
\centering
\includegraphics[width=0.8\linewidth]{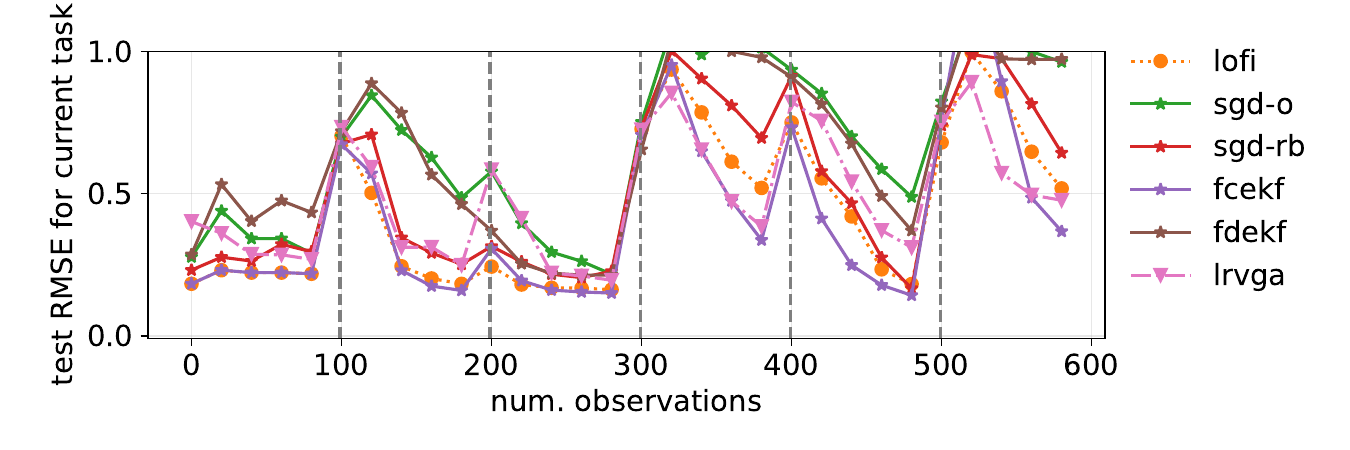}
\caption{
RMSE (rolling average)
on test data from the current task for the 1d regression benchmark
for different estimators.
Vertical lines denote change in the distribution
(unknown to the algorithm).
\demosfiggen{nonstat-1d-regression.ipynb}
}
\label{fig:regression-1d-error-vs-time}
\end{figure}


In this section, we  consider a synthetic 1d nonstationary regression problem
which exhibits ``concept drift''
\citep{Gama2014}.
Specifically we define the data generating process at time $t$
to be $p_t(x,y) = p(x) p_{d(t)}(y|x)$,
where $p(x)=\Unif(-2,2)$ is the input distribution,
 $d(t) \in \{1,\ldots,K\}$ specifies which distribution to use at time $t$,
and $p_k(y|x) = \gauss(y|f_k(x), \sigma^2)$
is the $k$'th such  distribution, for $k=1:K$.
We define $f_k(x) = x + 0.3 \sin(w_k^0 + w_k^1 \pi x)$,
where $\vw_k$ are randomly sampled coefficients
corresponding to the phase and frequency of the sine wave.
We assume $d(t)$ is a staircase function,
so $d(t) = k$ for $T_{k-1} \leq t \leq T_{k}$,
where $T_k - T_{k-1} = 250$ is the number of steps
before  the distribution changes.
We visualize these random functions in
\cref{fig:regression-1d-functions}.

Next we fit a one-layer MLP (with 50 hidden units)
on this data stream. (The algorithms are unaware of the task boundaries,
corresponding to the change in distribution.)
The test error (for the current distribution) vs time is shown in 
\cref{fig:regression-1d-error-vs-time}.
The ``spikes'' in the error rate correspond to times when
the distribution changes. In some cases  the change in distribution is small
(when $f_t$ is similar to $f_{t-1}$),
but in other cases there is a large shift.
The speed with which an estimator can adapt to such changes
is a critical performance metric in many domains.
We see that FCEKF adapts the fastest, followed by \LOFI
and then \LRVGA. 
SGD and the diagonal methods are less sample efficient.
However, after a sufficient number of training examples,
most methods converge to a good fit,
as shown in
\cref{fig:regression-1d-functions}.

\clearpage
\subsection{Stationary image classification}
\label{appx:classification}

In this section we report more results on  stationary classification experiments.

In \cref{fig:fmnist-st-nll}
we plot the plugin NLL  on static fasion MNIST using an MLP
with 2 layers with 500 hidden units each, with $648,010$ parameters.
The trends are similar to the misclassification rate
in  \cref{fig:fmnist-st-clf-miscl}.

In \cref{fig:fmnist-st-clf-nlpd} we plot the NLPD results
using the linearized likelihood
and deterministic probit trick discussed in
\cref{appx:predict-obs}.
We see that in general NLPD outperforms the plugin NLL.
Furthermore, the posterior from LOFI outperforms the posterior
from the (diagonal) Laplace approximation.

Next we use a CNN,
specifically a LeNet-style architecture
with 3 hidden layers and 421,641 parameters.
The results are shown in \cref{fig:fmnist-st-cnn}.
The trends are similar to the MLP case, except the gaps in performance
among the methods are narrower.

In \cref{tab:inflation-ablation} we summarize
the effects of changing the rank of \lofi,
and of different kinds of inflation (discussed in \cref{sec:inflation}),
and of switching from diagonal to spherical covariance (discussed in \cref{sec:spherical})
on the static fashion-MNIST dataset (using the CNN  model) 
after 500 training examples.
Not surprisingly, higher rank improves the results,
as does using a diagonal approximation.
However, inflation seems to have a negligible effect.
In \cref{fig:fmnist-st-cnn-vs-rank},
we visualize these differences as a function of sample size.

\begin{table}[h!]
    \centering
    \resizebox{\columnwidth}{!}{
    \begin{tabular}{lcccc|cccc}
         & \multicolumn{4}{c}{spherical} &\multicolumn{4}{c}{diagonal} \\
         & none & bayesian & hybrid & simple & none & bayesian & hybrid & simple \\
        rank & & & & & & & & \\
        \noalign{\vskip1mm}
        \hline
        \noalign{\vskip-1mm}
        \noalign{\vskip 2mm}
        $1$ & $42.6$ $\pm$ $0.9$ & $42.6$ $\pm$ $0.9$ & $42.6$ $\pm$ $0.9$ & $41.5$ $\pm$ $1.2$ & $41.3$ $\pm$ $1.1$ & $40.1$ $\pm$ $1.1$ & $40.6$ $\pm$ $1.2$ & $40.6$ $\pm$ $1.2$ \\
        $5$ & $37.5$ $\pm$ $1.1$ & $37.8$ $\pm$ $1.1$ & $37.6$ $\pm$ $1.1$ & $38.0$ $\pm$ $1.1$ & $36.6$ $\pm$ $1.3$ & $37.0$ $\pm$ $2.0$ & $37.0$ $\pm$ $2.0$ & $37.0$ $\pm$ $2.0$ \\
        $10$ & $31.8$ $\pm$ $1.0$ & $32.4$ $\pm$ $1.1$ & $30.8$ $\pm$ $0.8$ & $31.2$ $\pm$ $0.8$ & $30.8$ $\pm$ $1.0$ & $32.5$ $\pm$ $1.5$ & $31.7$ $\pm$ $1.1$ & $30.6$ $\pm$ $0.8$ \\
        $20$ & $31.5$ $\pm$ $1.0$ & $35.9$ $\pm$ $1.6$ & $30.1$ $\pm$ $0.9$ & $30.1$ $\pm$ $0.8$ & $28.7$ $\pm$ $0.6$ & $31.1$ $\pm$ $1.2$ & $32.7$ $\pm$ $0.9$ & $32.3$ $\pm$ $1.1$ \\
        $50$ & $28.0$ $\pm$ $0.7$ & $31.7$ $\pm$ $1.3$ & $31.7$ $\pm$ $1.3$ & $31.7$ $\pm$ $1.3$ & $28.6$ $\pm$ $0.8$ & $28.4$ $\pm$ $0.7$ & $29.1$ $\pm$ $0.6$ & $28.4$ $\pm$ $0.7$ \\
    \end{tabular}
    }
    \caption{Stationary fashion-MNIST test set misclassification rates using \lofi of various ranks after $500$ training examples.
    We show results 
    for diagonal vs spherical covariance
    and different forms of inflation (described in \cref{appx:inflation}).
    Means and standard errors computed over $10$ trials.}
    \label{tab:inflation-ablation}
\end{table}

\eat{
\begin{figure}
\centering
\includegraphics[height=2.3in]
{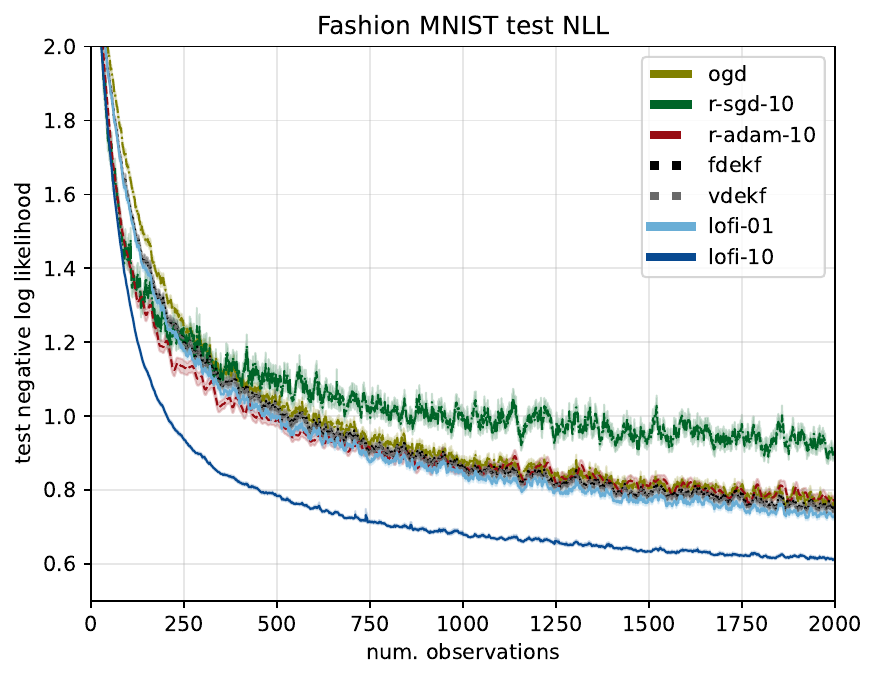}
\caption{
Test set NLL  vs number of observations on the 
fashion-MNIST dataset using MLP. We show the mean and 
standard errors across $100$ random trials.
\collasclffiggen{stationary\_mnist\_clf.py}
}
\end{figure}
}

\begin{figure*}
\centering
\begin{subfigure}[b]{0.47\textwidth}
  \centering
  \includegraphics[height=2.3in]{figs/experiments/stationary-mnist-clf-test-nll.pdf}
  \caption{ }
\label{fig:fmnist-st-nll}
\end{subfigure}
\begin{subfigure}[b]{0.47\textwidth}
  \centering
\includegraphics[height=2.3in]{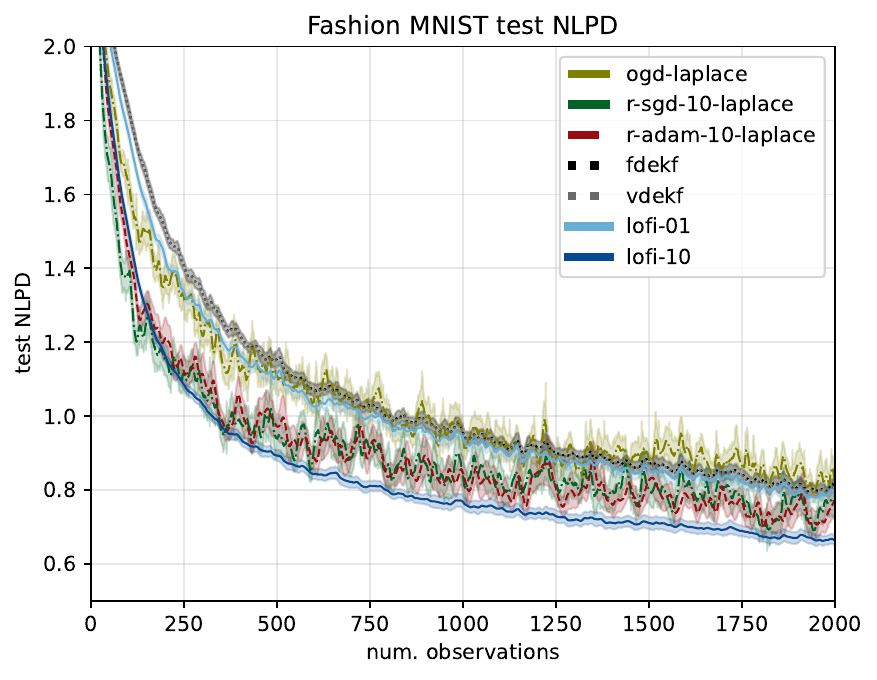}
\caption{ }
\label{fig:fmnist-st-clf-nlpd}
\end{subfigure}
\caption{
  Test set performance vs number of observations on the 
fashion-MNIST dataset using MLP. We show the mean and 
standard errors across random trials.
(a) Negative log likelihood ($100$ random trials).
(b) NLPD under linearized observation model with probit approximation ($20$ random trials).
\collasclffiggen{generate\_stationary\_clf\_plots.ipynb}
}
\end{figure*}

\begin{figure}
\centering
\begin{subfigure}[b]{0.47\textwidth}
    \centering
    \includegraphics[height=2.3in]{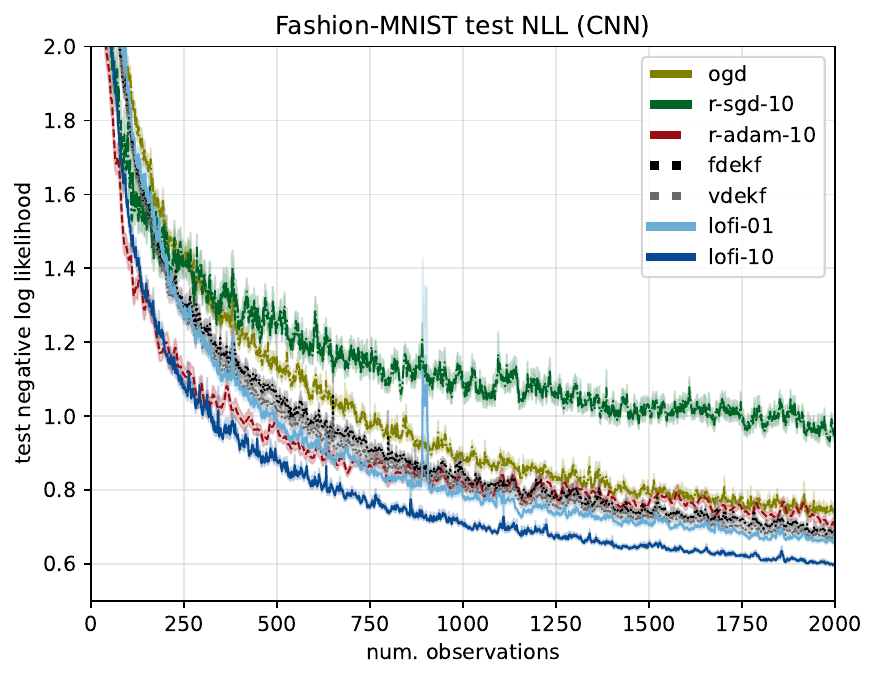}
    \caption{ }
\label{fig:fmnist-st-cnn-nll}
\end{subfigure}
\begin{subfigure}[b]{0.47\textwidth}
    \centering
    \includegraphics[height=2.3in]{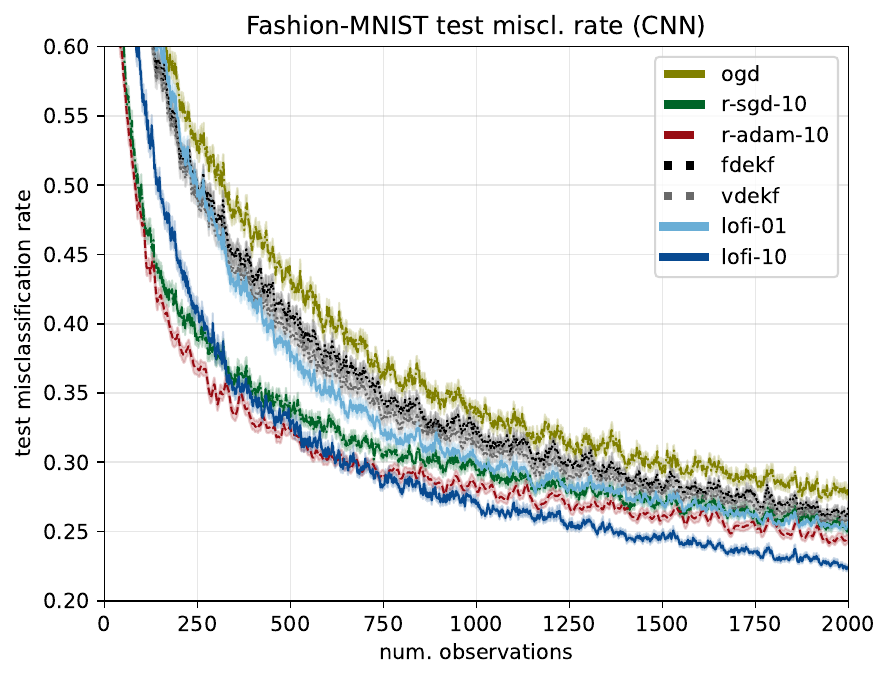}
    \caption{ }
\label{fig:fmnist-st-cnn-miscl}
\end{subfigure}
\caption{
Test set performance vs number of observations on the 
fashion-MNIST dataset using a CNN.
We show the mean and standard errors across 
$100$ random trials.
(a) Negative log-likelihood.
(b) Misclassification rate.
\collasclffiggen{generate\_stationary\_clf\_plots.ipynb}
}
\label{fig:fmnist-st-cnn}
\end{figure}

\begin{figure}
\centering
\begin{subfigure}[b]{0.47\textwidth}
    \centering
    \includegraphics[height=2.3in]{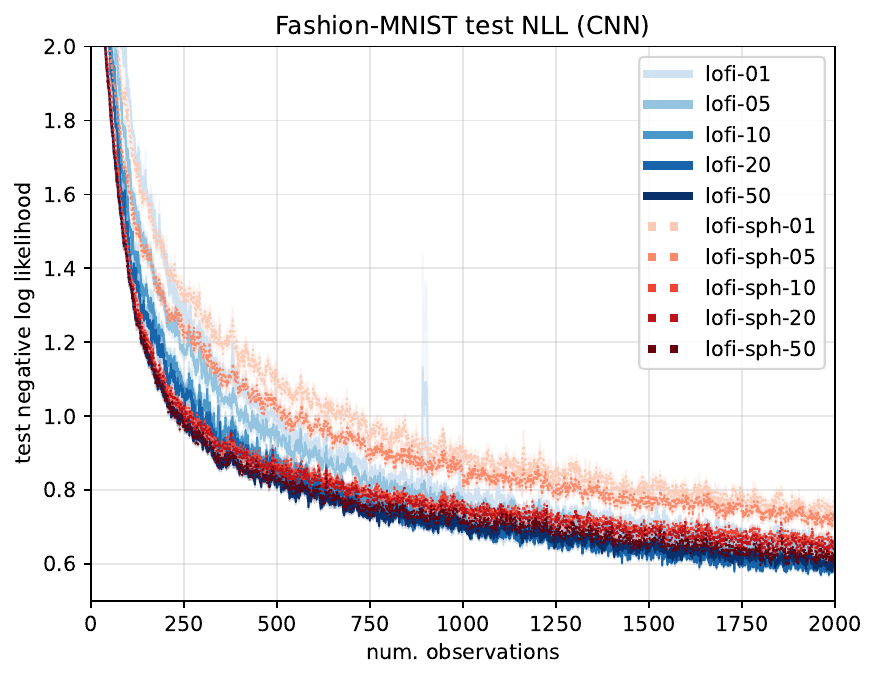}
    \caption{ }
\label{fig:fmnist-st-cnn-nll-lofi}
\end{subfigure}
\begin{subfigure}[b]{0.47\textwidth}
    \centering
    \includegraphics[height=2.3in]{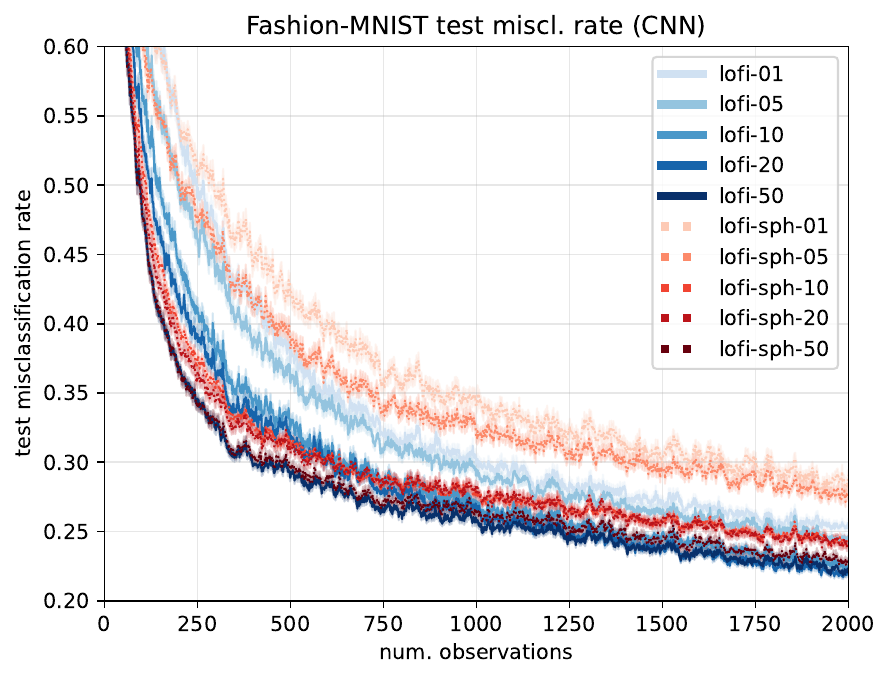}
    \caption{ }
\label{fig:fmnist-st-cnn-miscl-lofi}
\end{subfigure}
\caption{
  Results on fashion-MNIST classification dataset using a CNN.
  We visualize the  effect of changing rank, and using diagonal vs spherical LOFI
      (see \cref{appx:spherical}).
    "lofi-sph-xx" refers to spherical LO-FI of rank xx 
    (a) negative log-likelihood;
  (b) misclassification rate.
\collasclffiggen{generate\_stationary\_clf\_plots.py}
}
\label{fig:fmnist-st-cnn-vs-rank}
\end{figure}

\subsection{Piecewise stationary image classification}
 \label{appx:nonstclassification}

In this section we report more results on piecewise stationary classification experiments.

\paragraph{Permuted Fashion-MNIST}

\begin{figure}
\centering
\includegraphics[height=2.0in]
{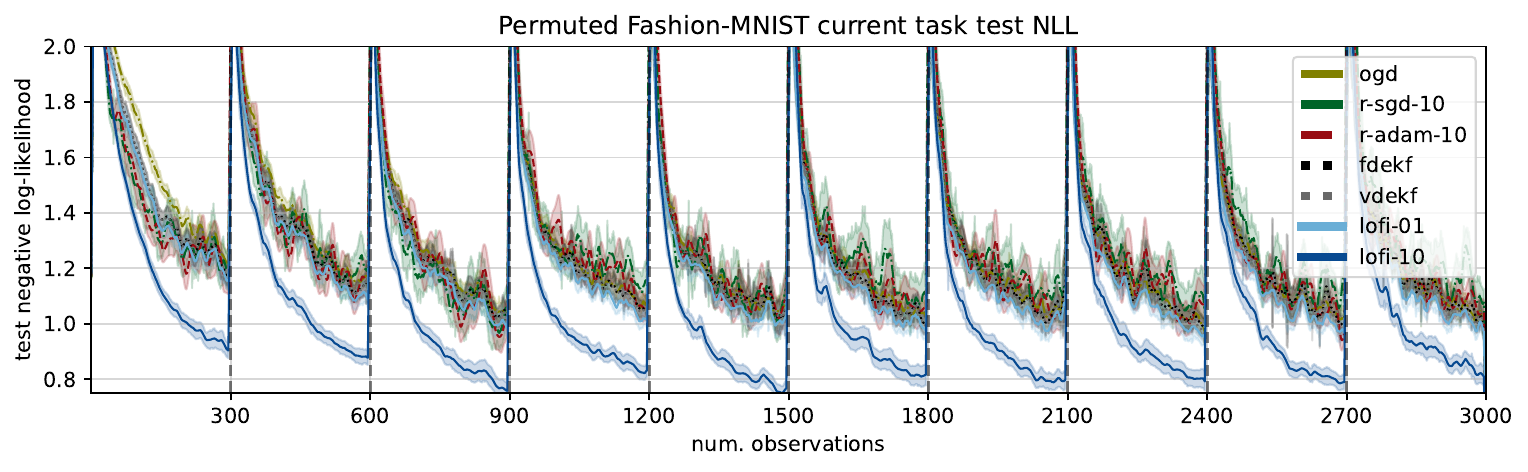}
\caption{
    Non-stationary permuted fashion-MNIST classification.
    The task boundaries are denoted by vertical lines.
    We report NLL
    performance on the current task's test set.  
    \collasclffiggen{generate\_permuted\_clf\_plots.ipynb}
}
\label{fig:ns-pmnist-curr-nll}
\end{figure}

\begin{figure*}
    \centering
    \includegraphics[height=2.0in]
    {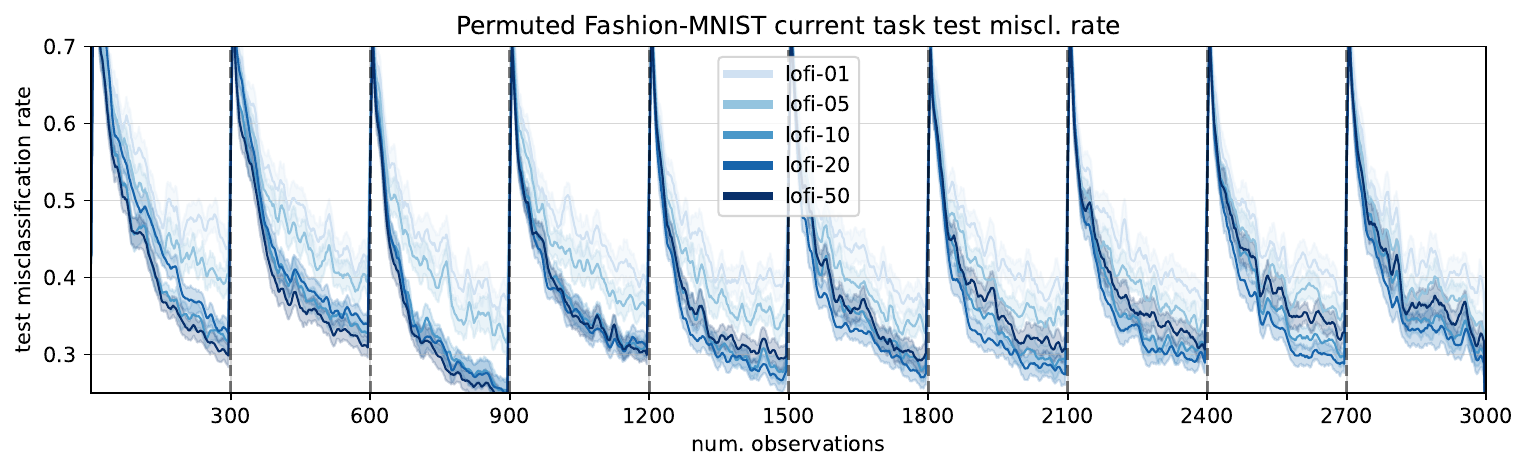}
    \caption{
        Test set misclassification rates vs number of observations on the 
        permuted fashion-MNIST dataset. We compare the performance
        as a function of the rank of LO-FI.
        \collasclffiggen{generate\_permuted\_clf\_plots.ipynb}.
    }
    \label{fig:ns-pmnist-curr-rank}
\end{figure*}

In \cref{fig:ns-pmnist-curr-nll}, we plot the NLL on permuted fashion MNIST.
The results are similar to the misclassification rates in
\cref{fig:ns-pmnist-curr},
except now the gap between LOFI and the other methods
is even larger.
In \cref{fig:ns-pmnist-curr-rank}
we compare the test-set misclassification rates of
LO-FI of various ranks. We see that performance improves with rank
and plateaus at about rank 10.

\begin{figure}
    \centering
    \includegraphics[width=0.75\textwidth]
    {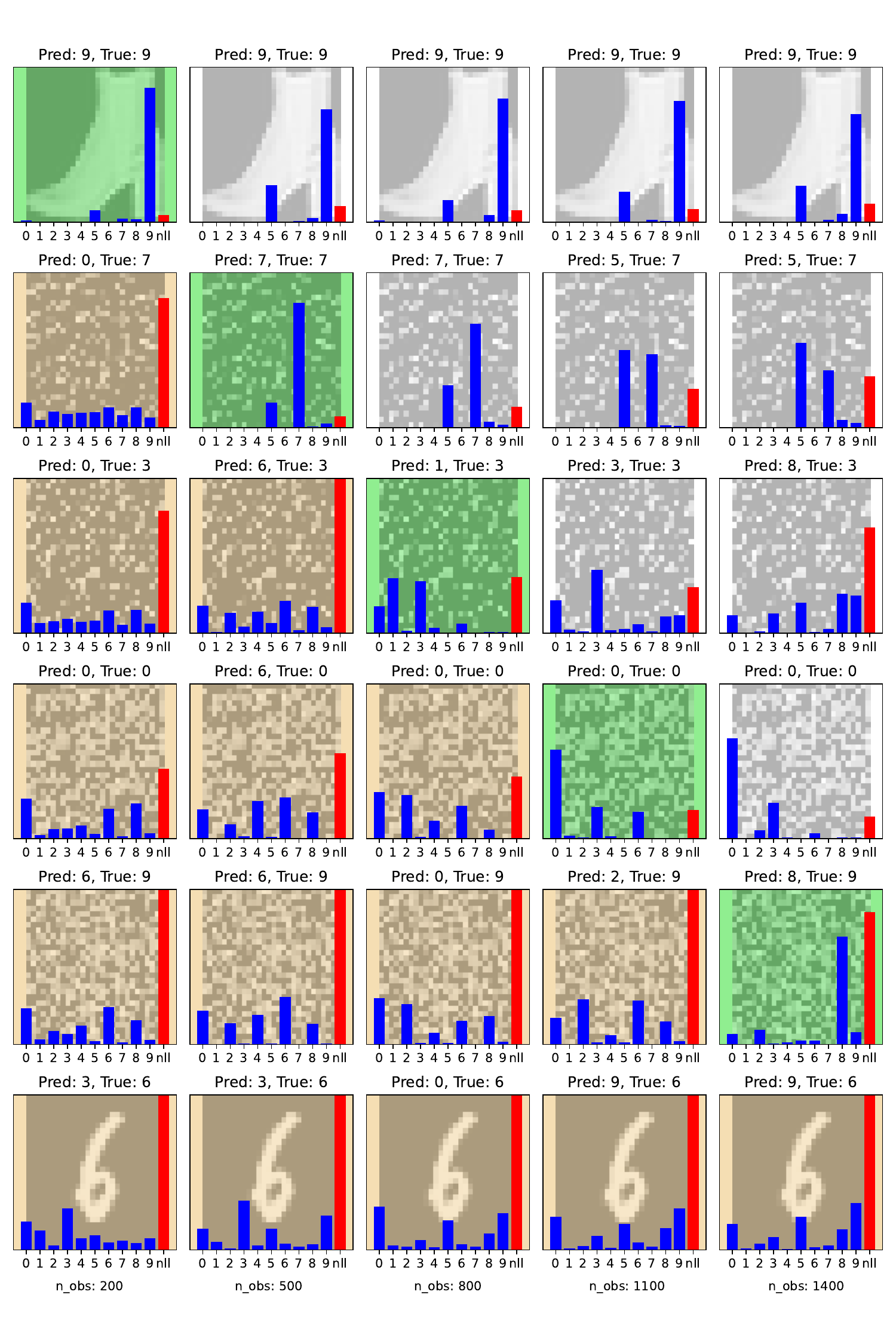}
    \caption{
        Test set predictions for non-stationary permuted fashion-MNIST
        classification problem using \lofi rank 10.
        Rows correspond to different distributions / tasks
        (i.e., different permutations of the data),
        and columns represent snapshots of the posterior predictive
        after every 50 steps  of online learning.
        Thus we can assess the performance of the model after seeing tasks
        $1:t$ by looking at the $t$'th column, and reading
        down across the rows.
        The first task uses the identity permutation.
        The last row 
        corresponds to an out-of-distribution example taken from the MNIST dataset.
        The current task is shown in green; previously seen tasks are shown in gray,
        and future tasks are shown in yellow.
        The blue bars are the predicted class probabilities (using plugin estimate),
        and the red bar is the NLL of the true label.
        in red.
        \collasclffiggen{probe.ipynb}
    }
    \label{fig:ns-pfmnist-probe}
\end{figure}

\begin{figure}
    \centering
    \includegraphics[width=0.75\textwidth]
    {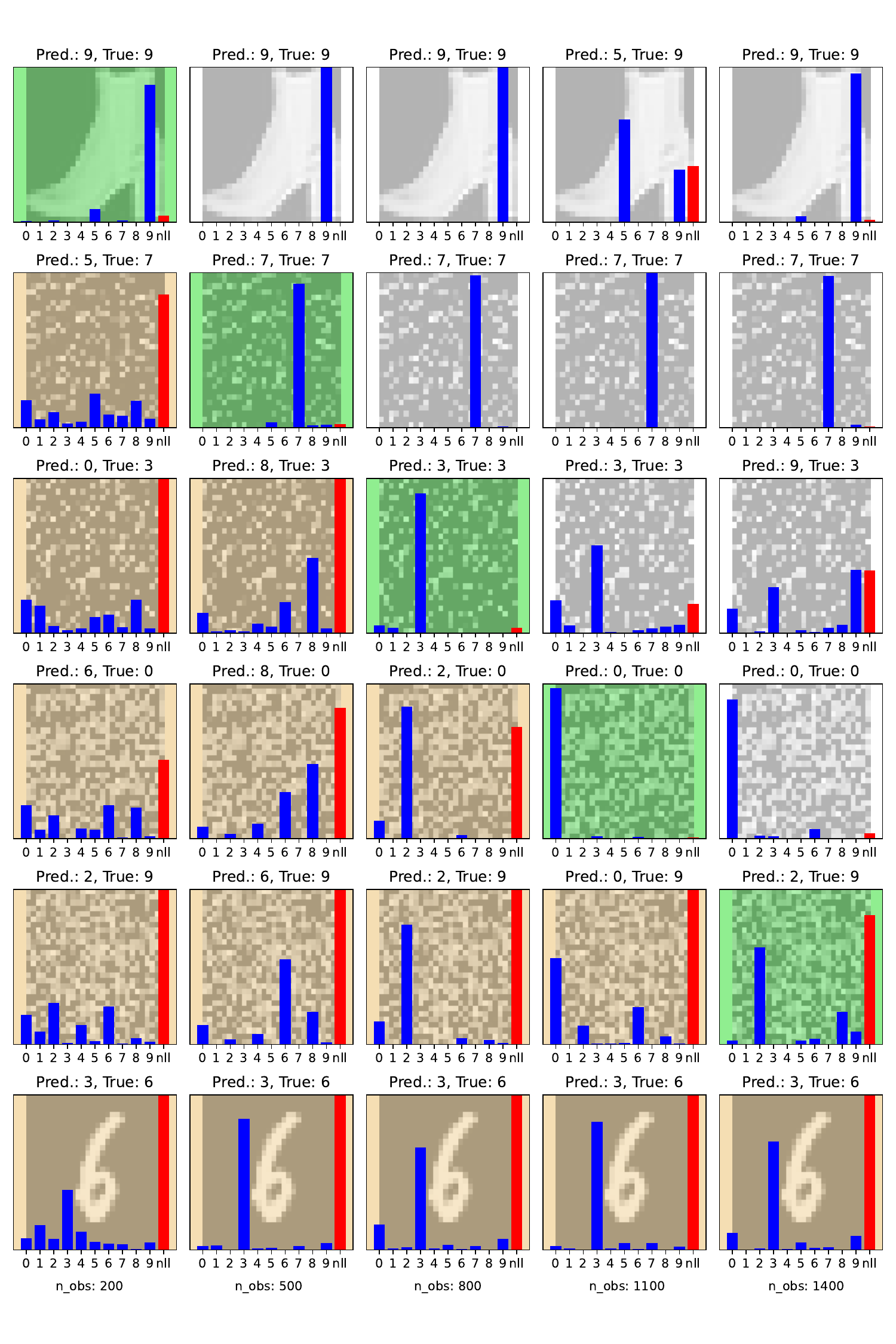}
    \caption{
      Same as \cref{fig:ns-pfmnist-probe}
      except using replay-SGD estimator.
    }
    \label{fig:ns-pfmnist-rsgd-probe}
\end{figure}

In \cref{fig:ns-pfmnist-probe}, we
show the test-set predictions (plugin approxmation)
from a \lofi-10 estimator
on a sample image from each of the first five tasks at various points
during training.
Before the model has seen data from a given distribution  (yellow panels),
its predictions are mostly uniform;
once it encounters data from the distribution, it learns rapidly,
as can be seen by the red NLL bar going down (the model is less
surprised when it sees the true label);
after the distribution shifts, we can still assess its performance
on past tasks (gray panels), and we see that the model is fairly good at remembering the past.
At the bottom of the plot, we show predictions on an OOD dataset that
the model is never trained on; we see that predictions remain close
to uniform, indicating high uncertainty.
In \cref{fig:ns-pfmnist-rsgd-probe}, we
show the same results using RSGD estimator;
we see that it is much less entropic,
even when it should be uncertain (e.g. for OOD).

\paragraph{Split Fashion-MNIST}

\begin{figure*}
\centering
\begin{tabular}{cc}
\includegraphics[height=2.3in]{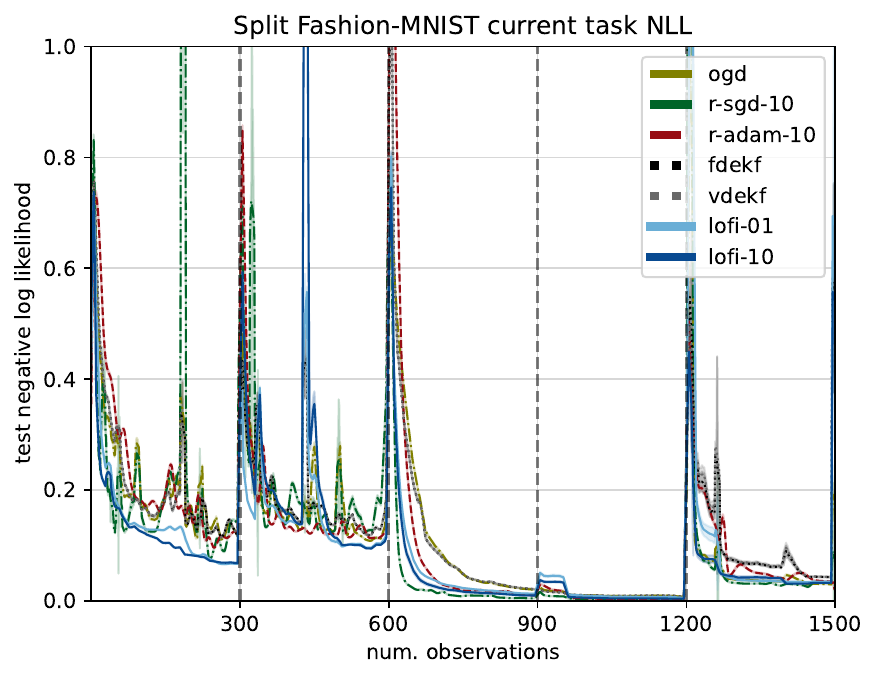}
&
\includegraphics[height=2.3in]{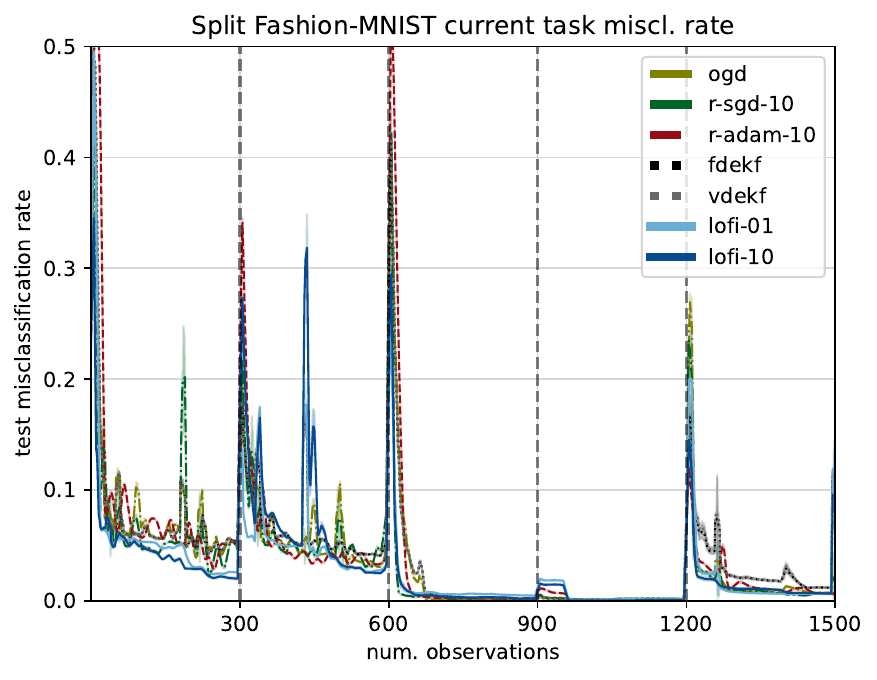}
\end{tabular}
\caption{
    Test set performance vs number of observations on the 
    split fashion-MNIST dataset.
    (a) negative log-likelihood;
    (b) misclassification rate.
     \collasclffiggen{generate\_split\_clf\_plots.ipynb}.
}
\label{fig:smnist}
\end{figure*}

In \cref{fig:smnist}, we evaluate the methods using the split fashion-MNIST dataset.
This task seems so easy that we cannot detect any substantial difference in
test-set performance among the different methods.

\subsection{Slowly changing image classification}

\begin{figure*}
\centering
\begin{subfigure}[b]{0.47\textwidth}
  \centering
  \includegraphics[height=2.3in]{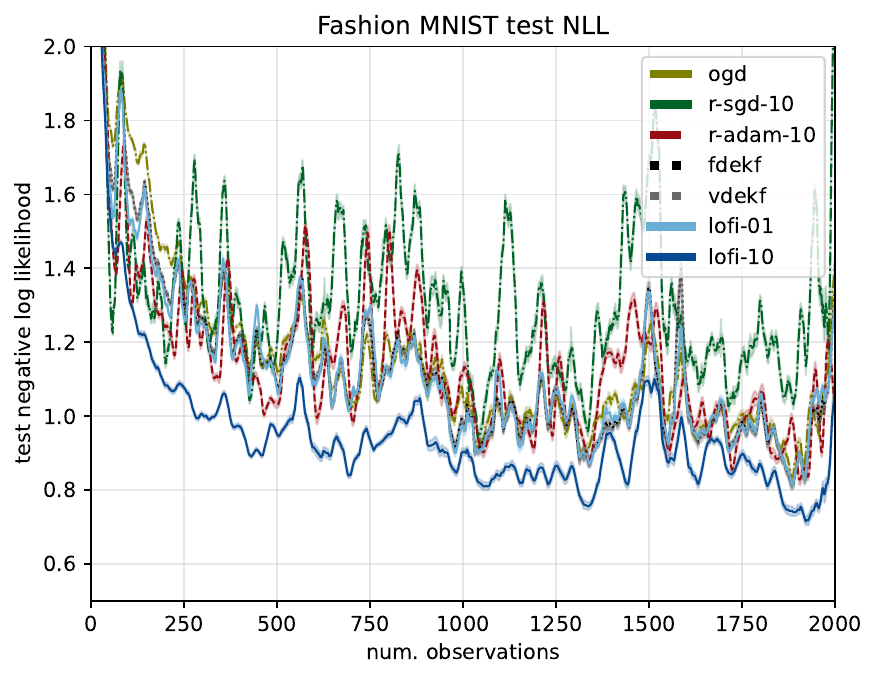}
\caption{ }
\label{fig:gr-nll}
\end{subfigure}
\begin{subfigure}[b]{0.47\textwidth}
  \centering
\includegraphics[height=2.3in]{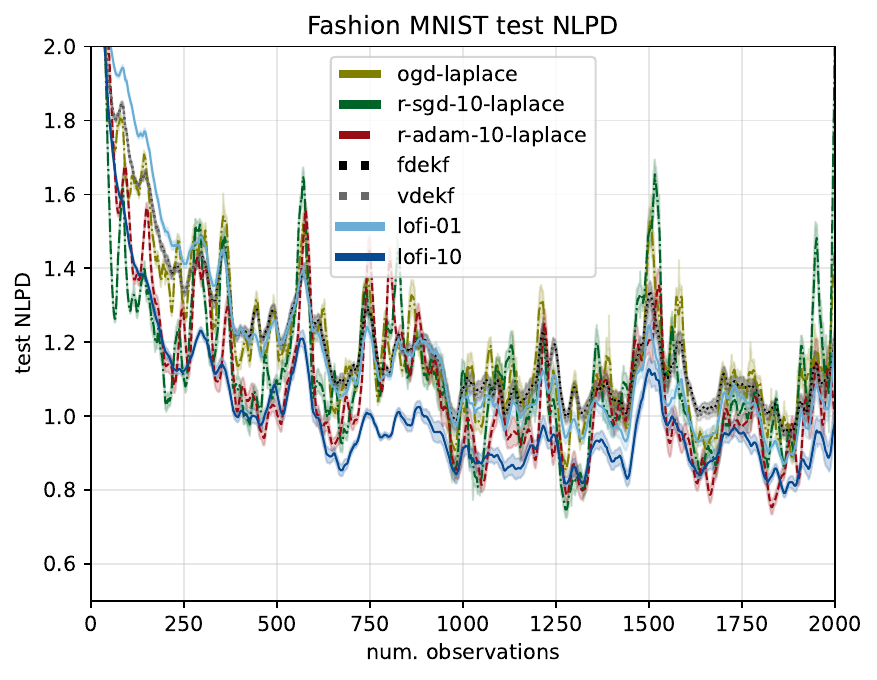}
  \caption{ }
  \label{fig:gr-NLPD}
\end{subfigure}
\caption{
      Gradually rotating fashion-MNIST classification.
      We evaluate the  performance on a test set
      from the current distribution (within a window).
      (a) NLL.
      (b) NLPD under probit approximation.
      \collasclffiggen{generate\_rotated\_clf\_plots.ipynb}.
}
\label{fig:gr-NLL-NLPD}
\end{figure*}

In \cref{fig:gr-NLL-NLPD}
we plot NLL  and NLPD
for the gradually rotating fashion-MNIST experiment.
The difference between the methods is more visible when judged by NLL compared to the misclassification
error in \cref{fig:gr-clf}.
We see that \lofi outperforms other methods.

\eat{
\begin{figure}
\centering
\includegraphics[height=2.3in]{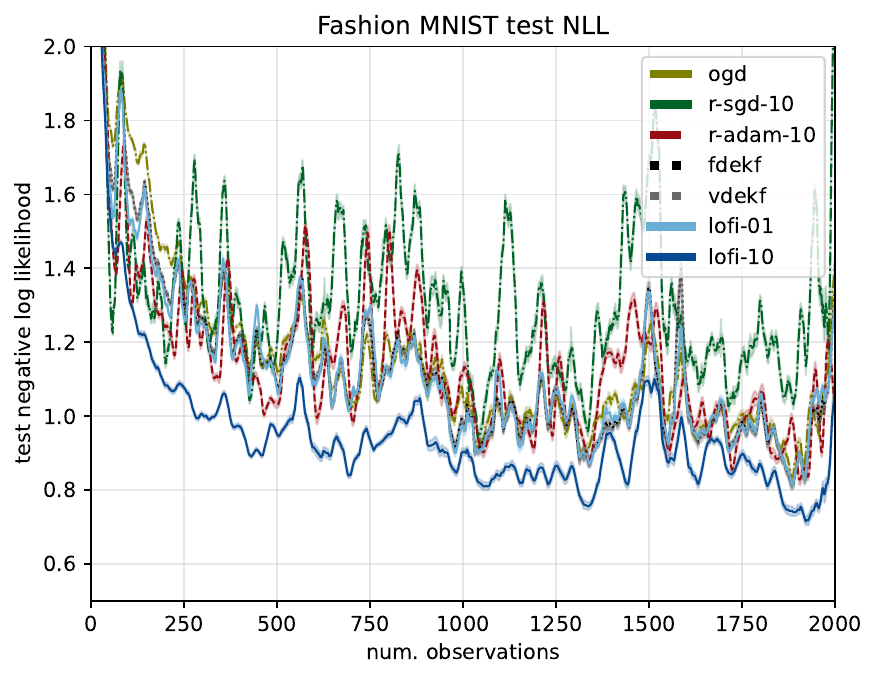}
\caption{
    Gradually rotating fashion-MNIST classification.
    We evaluate the NLL performance on a test set
    from the current distribution (within a window).
    \collasclffiggen{nonstationary\_rmnist\_clf.py}.
}
\label{fig:gr-nll}
\end{figure}
}

\clearpage
\subsection{Stationary image regression}
\label{sec:mnist-reg-extra}

\begin{figure}
\centering
\begin{subfigure}[b]{0.47\textwidth}
\centering
\includegraphics[height=2.3in]{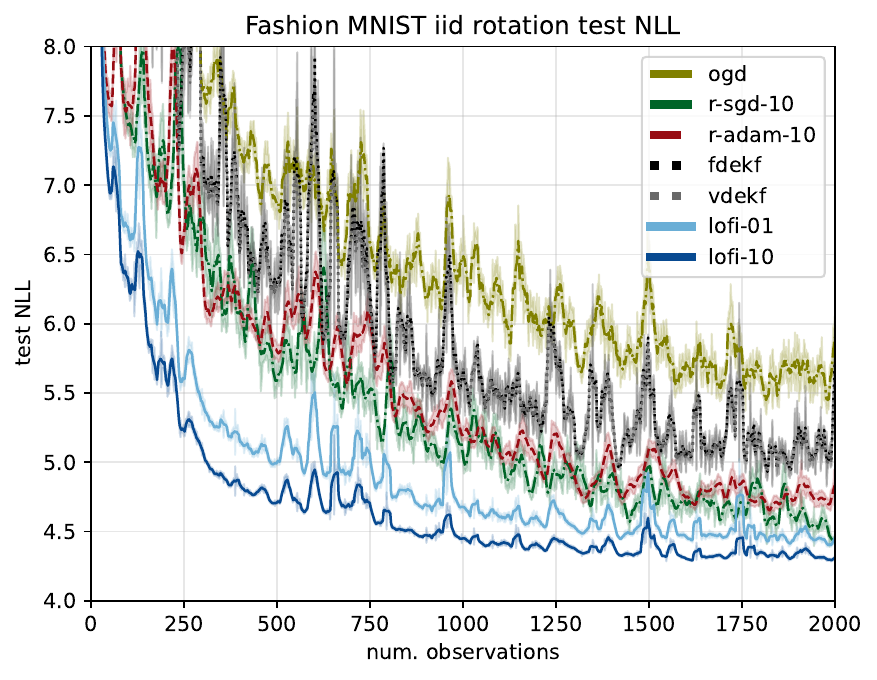}
\caption{ }
\label{fig:rotated-mnist-iid-nll}
\end{subfigure}
~
\begin{subfigure}[b]{0.47\textwidth}
\centering
\includegraphics[height=2.3in]{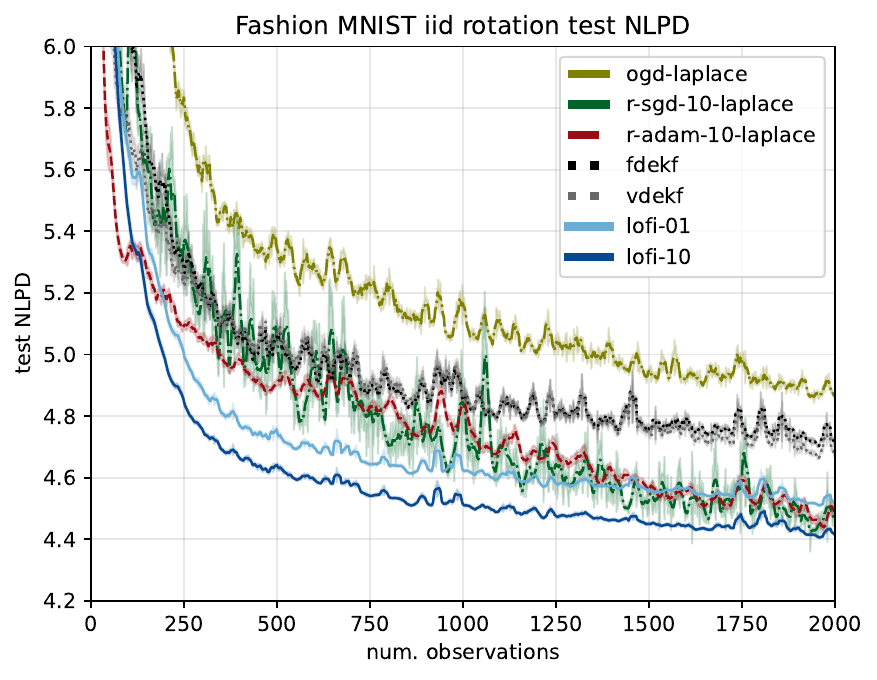}
\caption{ }
\label{fig:rotated-mnist-iid-nlpd}
\end{subfigure}
\caption{
IID rotated fashion-MNIST regression problem.
(a) NLL using  MAP plug-in estimate.
(b) NLPD under linearized observation model.
\collasregfiggen{generate\_iid\_reg\_plots.ipynb}
}
\label{fig:rotated-mnist-iid}
\end{figure}

In \cref{fig:rotated-mnist-iid-nll} 
we show the NLL (per example) for the static fashion-MNIST regression problem.
This has the same shape as the RMSE results in \cref{fig:rotated-mnist-iid-rmse},
since NLL = RMSE + constant,
since we assume the observation noise is fixed.

In \cref{fig:rotated-mnist-iid-nlpd} we show
the NLPD for the same problem,
which is approximated using the posterior predictive distribution
under the linearized observation model (see \cref{appx:predict-obs}).
We see that the NLPD metric of each method outperforms its respective 
NLL metric, and the variance is much lower.
We also see that the posterior from LOFI outperforms the posterior
from (diagonal) Laplace.

\subsection{Piecewise stationary image regression}

\begin{figure}
\centering
\begin{subfigure}[b]{\textwidth}
\centering
\includegraphics[height=2.0in]{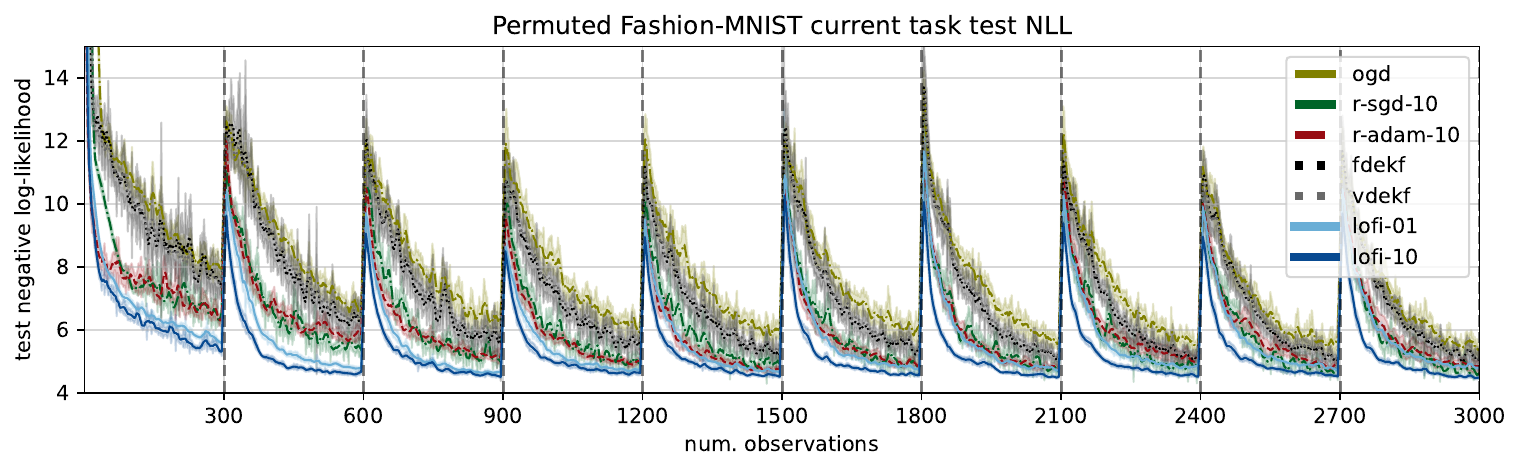}
\caption{ }
\end{subfigure}
~
\begin{subfigure}[b]{\textwidth}
\centering
\includegraphics[height=2.0in]{figs/experiments/nonstationary-rpfmnist-curr-rmse.pdf}
\caption{ }
\end{subfigure}
\caption{
Permuted rotating Fashion MNIST regression problem using MAP plugin prediction.
(a) Negative log-likelihood; (b) RMSE.
\collasregfiggen{generate\_permuted\_reg\_plots.ipynb}
}
\label{fig:prfmnist-reg-app}
\end{figure}

In \cref{fig:prfmnist-reg-app} we show results for a piecewise stationary
distribution created by using permuted fashion MNIST with $300$ samples
per task to create $10$ tasks.
We see that \lofi outperforms RSGD by a large margin.

\subsection{Slowly changing image regression}

\begin{figure}
\centering
\begin{subfigure}[b]{0.47\textwidth}
\centering
\includegraphics[height=2.3in]{figs/gradually-rotating/ou-rotating-window-rmse-all.pdf}
\caption{ }
    \label{fig:rotated-mnist-damped-rmse-all-app}
\end{subfigure}
~
\begin{subfigure}[b]{0.47\textwidth}
\centering
\includegraphics[height=2.3in]{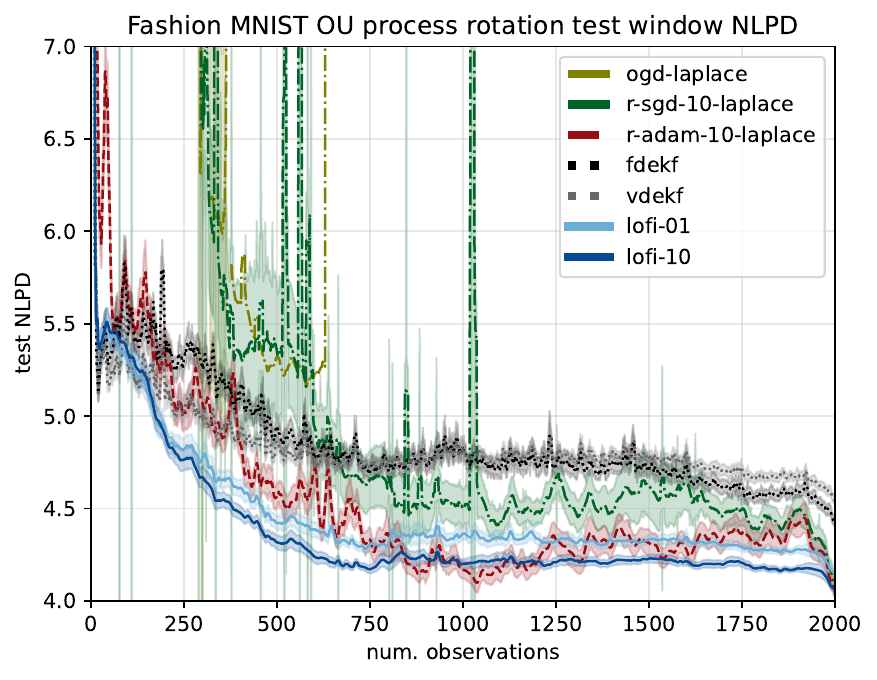}
\caption{ }
\label{fig:rotated-mnist-damped-nlpd}
\end{subfigure}
\caption{
  Slowly drifting MNIST regression problem.
  (a) RMSE using MAP estimate.
(b) NLPD using linearized likelihood.
\collasregfiggen{generate\_rw\_reg\_plots.ipynb}
}
\label{fig:rotated-mnist-damped}
\end{figure}

In \cref{fig:rotated-mnist-damped-nlpd} 
we show the linearized approximation to the NLPD 
on the drifting  MNIST rotation regression problem.
Note that under the nonstationary setting, the GD-based methods
are extremely noisy, whereas \lofi is much more stable.

\clearpage
\subsection{Bandits}
\label{sec:bandits-extra}

\begin{figure}
\centering
\includegraphics[height=2.5in]
{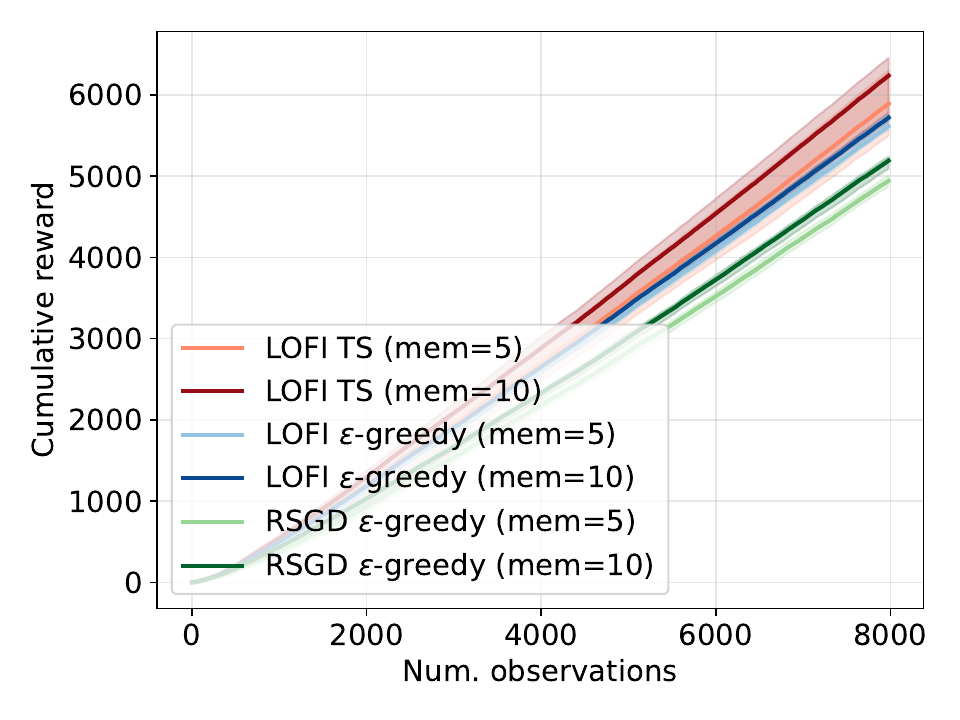}
\caption{
Reward vs time on MNIST bandit problem.
We show results (averaged over 5 trials)
using Thompson sampling or $\epsilon$-greedy with $\epsilon=0.1$.
\banditfiggen{bandit-vs-memory.ipynb}
}
\label{fig:bandits-vs-time-app}
\end{figure}

In \cref{fig:bandits-vs-time-app} we show reward vs time
for different methods on the MNIST bandit problem.
We see that LOFI with Thompson sampling  beats
LOFI with $\epsilon$-greedy,
which beats replay SGD with $\epsilon$-greedy.

\clearpage
\subsection{LRVGA implementation}

\begin{figure}[h!]
\includegraphics[width=0.8\linewidth]{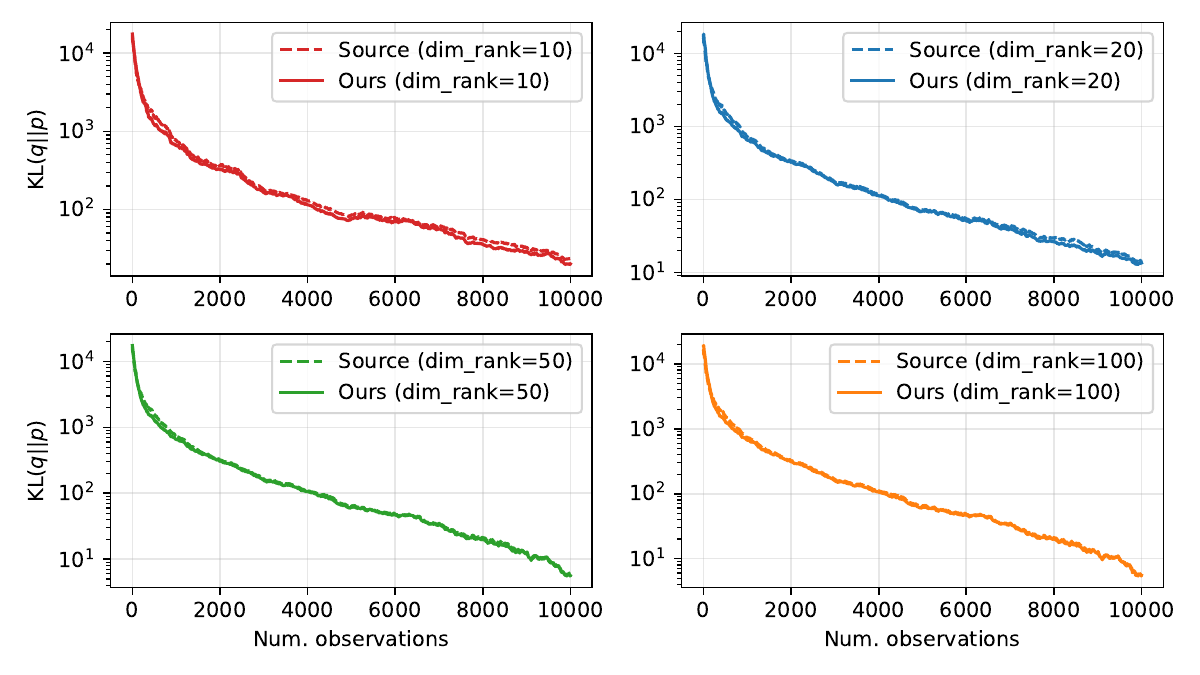}
\centering
\caption{
    KL divergence comparison between the original LRVGA implementation (source) and our implementation.
    \miscfiggen{xp-lrvga-linear-regression.ipynb}
}
\label{fig:lvrga-kl}
\end{figure}

The orignal numpy code for LRVGA code
is at  \url{https://github.com/marc-h-lambert/L-RVGA}.
We reimeplemented it in JAX and
verified that it gives the same results when applied to their
linear regression examples.
Specifically we used their source code with initial hyperparameters $\sigma^2_0 = 1$ and $\epsilon=10^{-3}$.
In \cref{fig:lvrga-kl}, we visually compare the KL between our posterior and theirs,
verifying that our implementation is correct.
By using JAX, we gain speed.
More importantly we can extend the method
to the nonlinear case by using JAX's autodiff framework to compute the relevant gradients.

\eat{
In Table \ref{tab:lrvga-comparisson} we show the KL divergence from the source code and our implementation of LRVGA; the small numerical differences are due to the use of JAX vs numpy.

\begin{table}[h!]
    \centering
    \begin{tabular}{c | c c}
        \hline\hline
        Rank & Source & Ours \\
        \hline
        10 & 23.475 & 20.066 \\
        20 & 14.574 & 13.511 \\
        50 & 5.744 & 5.618 \\
        100 & 5.438 & 5.540
    \end{tabular}
    \caption{LRVGA linear regression}
    \label{tab:lrvga-comparisson}
\end{table}
}

\clearpage

\section{Covariance inflation}
\label{appx:inflation}
\label{sec:inflation}

In this section we derive a modified version of \lofi
where we use a Bayesian version of
the covariance inflation trick of
\citep{Ollivier2018,Alessandri2007,Kurle2020}
to account for errors introduced by approximate
inference, such as linearizing the observation model
(see \citep{Kulhavy1993,Karny2014} for analysis).
In practice this just requires a rescaling
of the terms in the posterior precision matrix
at the end of each update step (or equivalently, just before
doing a predict step). This rescaling only takes $O(\nparams)$ time,
so is negligible extra cost.
However, we have found it does not seem to improve results
(see \cref{tab:inflation-ablation} for results on UCI regression);
thus this section is just for ``historical interest''.

\eat{
We derive a Bayesian version of the covariance inflation trick in \citeauthor{Alessandri2007}'s \citeyearpar{Alessandri2007} ``Modified EKF", which scales up the posterior variance by a factor of $1+\alpha$ after each time step. 
Following \cite{Ollivier2018}, this can be interpreted as exponential discounting of past loglikelihoods.
In our case, this discounting helps to adjust for gradients going stale as we update the parameter predictions, due to drift in the linearized observation model.
}

\Cref{appx:bayesian-inflation} derives our Bayesian inflation method, in which discounting is applied only to the likelihood and not to the prior. This amounts to deflating the entire log posterior and then adding back in the appropriate fraction of the log prior. 
\Cref{appx:simple-inflation} derives a simpler version of inflation that discounts the entire posterior (i.e., likelihood and prior), matching past work \citep{Alessandri2007,Ollivier2018}.
\Cref{appx:hybrid-inflation} derives a hybrid inflation method that uses the covariance update from Bayesian inflation but, like simple inflation, does not change the mean. This turns out to be a special case of the regularized forgetting mechanism of \cite{Kulhavy1993}, which they derive based on uncertainty about the system dynamics rather than drift in the observation model.

We derive all three variations for a general state-space model and then show how they specialize to \lofi.
The results are formulas for going from the parameters of the posterior after step $t-1$ ($\vmu_{t-1},\vUpsilon_{t-1},\vW_{t-1}$) to parameters of an ``inflated" posterior ($\discount{\vmu}_{t-1},\discount{\vUpsilon}_{t-1},\discount{\vW}_{t-1}$). 
Applying inflation then amounts to substituting 
$\discount{\vmu}_{t-1},\discount{\vUpsilon}_{t-1},\discount{\vW}_{t-1}$ 
for 
${\vmu}_{t-1},{\vUpsilon}_{t-1},{\vW}_{t-1}$ 
in \cref{eq:mean-predict-appx,eq:diagonal-predict-appx,eq:lowrank-predict-appx} in \cref{appx:predict-step}.

\subsection{Bayesian inflation}
\label{appx:bayesian-inflation}

Consider first the special case of a static parameter ($\forall_t:\vtheta_t=\vtheta_0$).
The log posterior after step $t-1$ is
\begin{align}
    \log p(\vtheta|\data_{1:t-1}) = \log p(\vtheta) + \sum_{i=1}^{t-1} \log p(\vy_i|\vx_i,\vtheta) + const
\end{align}
We modify this expression by discounting the likelihood of each past observation by $(1+\alpha)^{-k}$, where $k=t-1-i$ is the lag. For Gaussian observations, this is equivalent to scaling up the observation covariance $\vR_i$ by $(1+\alpha)^{-k}$. We indicate this discounting by the subscripted probability $p_{t-1}$, where time $t-1$ is the reference point from which discounting is applied.
\begin{align}
    \log p_{t-1}(\vtheta|\data_{1:t-1}) = \log p(\vtheta) + \sum_{i=1}^{t-1} (1+\alpha)^{-(t-1-i)} \log p(\vy_i|\vx_i,\vtheta) + const
\end{align}
Passing from $p_{t-1}$ to $p_t$ amounts to applying an additional discount factor to the likelihoods, which is equivalent to discounting the entire log posterior and adding back a fraction of the log prior so that it is not discounted:
\begin{align}
    \log p_{t}(\vtheta|\data_{1:t-1}) 
    &= \log p(\vtheta) 
        + \sum_{i=1}^{t-1} (1+\alpha)^{-(t-i)} \log p(\vy_i|\vx_i,\vtheta) + const \\
    &= \log p(\vtheta) 
        + \frac{1}{1+\alpha} \sum_{i=1}^{t-1} (1+\alpha)^{-(t-1-i)} \log p(\vy_i|\vx_i,\vtheta) + const \\
    &= \frac{1}{1+\alpha} \log p_{t-1}(\vtheta|\data_{1:t-1}) 
        + \frac{\alpha}{1+\alpha} \log p(\vtheta)
\end{align}

The same reasoning applies in the general case with state dynamics. 
We expand the log posterior after step $t-1$ as
\begin{align}
    \log p_{t-1}(\vtheta_{t-1}|\data_{1:t-1}) 
    = \log p(\vtheta_{t-1}) 
        + \log p_{t-1}(\data_{1:t-1}|\vtheta_{t-1}) + const
\end{align}
Passing from $p_{t-1}$ to $p_t$ amounts to discounting the data contribution while preserving the latent predictive prior:
\begin{align}
    \log p_{t}(\vtheta_{t-1}|\data_{1:t-1}) 
    &= \log p(\vtheta_{t-1}) 
        + \frac{1}{1+\alpha} \log p_{t-1}(\data_{1:t-1}|\vtheta_{t-1}) + const \\
    &= \frac{1}{1+\alpha} \log p_{t-1}(\vtheta_{t-1}|\data_{1:t-1}) 
        + \frac{\alpha}{1+\alpha} \log p(\vtheta_{t-1}) + const
    \label{eq:inflation-general-log-prior-correction}
\end{align}
A similar result was derived in \citep{Kurle2020}.

We now specialize \cref{eq:inflation-general-log-prior-correction} to \lofi.
Given our initial prior $p(\vtheta_0) = \gauss(\vtheta_0|\vmu_0,\eta_0^{-1}\vI_\nparams)$ and dynamics $p(\vtheta_t|\vtheta_{t-1}) = \gauss(\vtheta_t|\gamma_t\vtheta_{t-1},q_t\vI_\nparams)$, the latent unconditional predictive prior
of the dynamical system at time $t-1$ is
\begin{align}
    p(\vtheta_{t-1}) 
    &= \gauss(\vtheta_{t-1} | \Gamma_{t-1} \vmu_0, \eta_{t-1}^{-1} \vI_\nparams) \\
\eta_t^{-1} &= \gamma_t^2 \eta_{t-1}^{-1} + q_t\\
\Gamma_{t-1} &= \prod_{i=1}^{t-1} \gamma_i
\end{align}
Substituting this and our posterior 
$p_{t-1}(\vtheta_{t-1}|\data_{1:t-1}) = \gauss(\vtheta_{t-1}|\vmu_{t-1},(\vUpsilon_{t-1}+\vW_{t-1}\vW_{t-1}^\trans)^{-1})$ 
into \cref{eq:inflation-general-log-prior-correction} yields
\begin{align}
    p_t(\vtheta_{t-1}|\data_{1:t-1}) = 
    \gauss\left(
       \vtheta_{t-1} \middle| 
        \discount{\vmu}_{t-1},
        \left(
            \discount{\vUpsilon}_{t-1}
            + \discount{\vW}_{t-1}\discount{\vW}_{t-1}^\trans
        \right)^{-1}
        \label{eq:inflation-update-bayesian}
    \right)
\end{align}
with
\begin{align}
    \discount{\vmu}_{t-1} &=
        \vmu_{t-1}
        + \frac{\alpha \eta_{t-1}}{1+\alpha} 
        \left(
            \discount{\vUpsilon}_{t-1}
            + \discount{\vW}_{t-1}\discount{\vW}_{t-1}^\trans
        \right)^{-1}
        (\Gamma_{t-1} \vmu_{0}-\vmu_{t-1}) 
        \label{eq:inflation-bayes-mean-update} \\
    \discount{\vUpsilon}_{t-1} &=
        \frac{1}{1+\alpha} \vUpsilon_{t-1}
        + \frac{\alpha \eta_{t-1}}{1+\alpha} \vI_\nparams
        \label{eq:inflation-bayes-diagonal-update} \\
    \discount{\vW}_{t-1} &=
        \frac{1}{\sqrt{1+\alpha}} \vW_{t-1}
        \label{eq:inflation-bayes-lowrank-update}
\end{align}

\Cref{eq:inflation-bayes-mean-update} implements a form of regularization toward the prior predictive mean $\Gamma_{t-1} \vmu_0$, which originates in the log-prior term in \cref{eq:inflation-general-log-prior-correction}. \Cref{eq:inflation-bayes-diagonal-update,eq:inflation-bayes-lowrank-update} implement inflation of the covariance by a factor of $1+\alpha$, together with the log-prior correction being added to $\discount{\vUpsilon}_{t-1}$.
Together these expressions show how the parameters of the distribution change as we pass from $p_{t-1}(\vtheta_{t-1}|\data_{1:t-1})$ to $p_{t}(\vtheta_{t-1}|\data_{1:t-1})$. Notice that we have incremented the subscript in $p_t$ but the random variable is still $\vtheta_{t-1}$. Thus $\discount{\vmu}_{t-1},\discount{\vUpsilon}_{t-1},\discount{\vW}_{t-1}$ 
define the ``post-inflation'' posterior that is passed to the predict step in \cref{appx:predict-step} to obtain the iterative prior, given by $\vmu_{t|t-1},\vUpsilon_{t|t-1},\vW_{t|t-1}$.

\subsection{Simple inflation}
\label{appx:simple-inflation}

A simpler version of inflation can be obtained by discounting the prior as well as the likelihood. In that case, passing from $p_{t-1}$ to $p_t$ amounts to discounting the entire log posterior. Thus instead of \cref{eq:inflation-general-log-prior-correction} we have
\begin{align}
    \log p_{t}(\vtheta_{t-1}|\data_{1:t-1}) 
    &= \frac{1}{1+\alpha} \log p_{t-1}(\vtheta_{t-1}|\data_{1:t-1}) 
    \label{eq:inflation-general-simple}
\end{align}
Substituting 
$p_{t-1}(\vtheta_{t-1}|\data_{1:t-1}) = \gauss(\vtheta_{t-1}|\vmu_{t-1},(\vUpsilon_{t-1}+\vW_{t-1}\vW_{t-1}^\trans)^{-1})$
yields
\begin{align}
    p_{t} (\vtheta_{t-1} | \data_{1:t-1}) 
    &= \gauss \left(
        \vtheta_{t-1} \middle| 
        \vmu_{t-1}, 
        (1+\alpha) 
        \left(
            \vUpsilon_{t-1}
            + \vW_{t-1} \vW_{t-1}^\trans
        \right)^{-1}
    \right)
    \label{eq:inflation-update-simple}
\end{align}
Thus we merely inflate the covariance by $1+\alpha$, as in \cite{Alessandri2007} and \cite{Ollivier2018}.
This implies the simple inflation equations
\begin{align}
    \discount{\vmu}_{t-1} &=
        \vmu_{t-1}
        \label{eq:inflation-simple-mean-update} \\
    \discount{\vUpsilon}_{t-1} &=
        \frac{1}{1+\alpha} \vUpsilon_{t-1}
        \label{eq:inflation-simple-diagonal-update} \\
    \discount{\vW}_{t-1} &=
        \frac{1}{\sqrt{1+\alpha}} \vW_{t-1}
        \label{eq:inflation-simple-lowrank-update}
\end{align}

\subsection{Hybrid inflation}
\label{appx:hybrid-inflation}

Rather than mixing in the latent predictive prior, as in \cref{eq:inflation-general-log-prior-correction}, we can mix in a distribution that uses the prior predictive variance but the posterior mean:
\begin{align}
    \log p_{t}(\vtheta_{t-1}|\data_{1:t-1}) 
    &= \frac{1}{1+\alpha} \log p_{t-1}(\vtheta_{t-1}|\data_{1:t-1}) 
        + \frac{\alpha}{1+\alpha} \log \gauss(\vtheta_{t-1} | \vmu_{t-1}, \eta_{t-1}^{-1} \vI_\nparams) + const
\end{align}
This approach fits into the more general regularized forgetting framework of \cite{Kulhavy1993} and can be interpreted heuristically as regularizing the covariance but not the mean, which may be preferable since $\vmu_0$ is sampled randomly rather than being an informed prior.
In this case, substituting \lofi's posterior 
$p_{t-1}(\vtheta_{t-1}|\data_{1:t-1}) = \gauss(\vtheta_{t-1}|\vmu_{t-1},(\vUpsilon_{t-1}+\vW_{t-1}\vW_{t-1}^\trans)^{-1})$
yields
\begin{align}
    p_t(\vtheta_{t-1}|\data_{1:t-1}) = 
    \gauss\left(
       \vtheta_{t-1} \middle| 
        \vmu_{t-1},
        (1+\alpha)
        \left(
            \vUpsilon_{t-1}
            + \alpha\eta_{t-1}\vI_\nparams
            + \vW_{t-1}\vW_{t-1}^\trans
        \right)^{-1}
        \label{eq:inflation-update-hybrid}
    \right)
\end{align}
implying
\begin{align}
    \discount{\vmu}_{t-1} &=
        \vmu_{t-1}
        \label{eq:inflation-hybrid-mean-update} \\
    \discount{\vUpsilon}_{t-1} &=
        \frac{1}{1+\alpha} \vUpsilon_{t-1}
        + \frac{\alpha \eta_{t-1}}{1+\alpha} \vI_\nparams
        \label{eq:inflation-hybrid-diagonal-update} \\
    \discount{\vW}_{t-1} &=
        \frac{1}{\sqrt{1+\alpha}} \vW_{t-1}
        \label{eq:inflation-hybrid-lowrank-update}
\end{align}

\eat{
Carrying through the calculations in the previous two subsections, we see that we obtain the mean update from simple inflation and the covariance update from Bayesian inflation:
\begin{align}
    \vmu_{t|t-1} &= \gamma_t \vmu_{t-1} \\
    \vUpsilon_{t\vert t-1}
    &= \left(
        \gamma_{t}^{2} (1+\alpha)
        \left(
            \vUpsilon_{t-1}
            + \alpha \eta_{t-1} \vI_\nparams
        \right)^{-1}
        + q_{t} \vI_{\nparams}
    \right)^{-1} \\
    \vW_{t|t-1}
    &= \gamma_{t} \vD_{t|t-1} \vW_{t-1}
    {\rm chol}\left(
        \left(
            (1+\alpha) \vI_{\memory}
            + q_t \vW_{t-1}^{\trans} \vD_{t|t-1} \vW_{t-1}
        \right)^{-1}
    \right) \\
    \vD_{t|t-1} 
    &= \left(
        \gamma_{t}^{2} \vI_{\nparams}
        + \frac{q_{t}}{1+\alpha} 
        \left(
            \vUpsilon_{t-1}
            + \alpha \eta_{t-1}  \vI_\nparams
        \right)
    \right)^{-1}
\end{align}
}
  
\clearpage
\section{Spherical \lofi}
\label{appx:spherical}
\label{sec:spherical}

Here we describe a restricted version of \lofi in which the diagonal part of the precision is isotropic, $\vUpsilon_t=\eta_t\vI_\nparams$.
We denote this class of spherical plus low-rank models by SPL($\memory$), and refer to this algorithm as spherical \lofi, in contrast to the diagonal \lofi presented in the main text.
 Perhaps surprisingly,
 we find that the spherical restriction can slightly help predictive performance
 (see UCI regression results in \cref{tab:datasets-rank-passes-1}),
which is consistent with the claims in \citep{Tomczak2020}.
 However, the gains are not consistent across datasets.

The spherical restriction also allows a more efficient predict step, taking $O(\nparams)$ instead of $O(\nparams\memory^2)$ as in diagonal \lofi,
although in practice the running times
are indistinguishable
(see \cref{fig:energy-running-time}).
The update step takes $O(\nparams\memoryout^2)$, matching diagonal \lofi, although we present an alternative approximate method in \cref{appx:orth-svd} that takes only $O(\nparams\memory\nout)$.

\subsection{Warmup}

To motivate our approach, consider the case of stationary parameters,
where $p(\vtheta_t|\vtheta_{t-1})=
\delta(\vtheta_t - \vtheta_{t-1})$.
Then $\vSigma_{t|t-1} = \vSigma_{t-1}$ and hence
\cref{eq:EKF-var-update} becomes
    $\vSigma_{t}^{-1} = 
    \vSigma_{t-1}^{-1} + \vH_t^\trans \vR_t^{-1} \vH_t$.
Hence we can unwind \cref{eq:EKF-var-update} to get
\begin{align}
    \vSigma_t^{-1} = \eta_0 \vI_{\nparams} + \sum_{i=1}^t \vG_i \vG_i^\trans \label{eq:EKF-posterior-variance}
\end{align}
where $\vG_t = \vH_t^\trans\vA_t^\trans \in \real^{\nparams \times \nout}$
is the transposed Jacobian of the standardized observation vector $\vA_t \vy_t$.
The data-driven part of \cref{eq:EKF-posterior-variance} is a sum of outer products of gradients, taken over all time steps and (standardized) outcome dimensions. We seek a low-rank approximation of this sum,
\begin{align}
    \vW_t\vW_t^\trans \approx \sum_{i=1}^t \vG_i\vG_i^\trans
\end{align}
with $\vW_t\in\real^{P\times\memory}$. 
\lofi's update step uses incremental SVD after each observation to maintain $\vW_t$ as an approximation of the top $\memory$ non-normalized singular vectors of $[\vG_1,\dots,\vG_t]$. 
\Cref{appx:update-step-spherical} describes two alternative versions of incremental SVD, one matching that of diagonal \lofi (\cref{appx:full-svd}) and the other using a more efficient projection approximation (\cref{appx:orth-svd}).
In both cases we will have $\vW_{t} = \vU_{t} \vLambda_{t}$, where $\vLambda_{t}=\diag(\vlambda_{t})$ is a diagonal $\memory \times \memory$ matrix, and $\vU_{t}^\trans \vU_t = \vI_{\memory}$. Therefore the approximate posterior is written as
\begin{align}
    p(\vtheta_{t}|\data_{1:t}) = \gauss\left(\vtheta_t\middle|\vmu_t,\left(\eta_t\vI_\nparams + \vU_t\vLambda_t^2\vU_t^\trans\right)^{-1}\right)
    \label{eq:SPL-posterior}
\end{align}
Unlike in diagonal \lofi, the spherical part of the precision is data-independent. This is because any data-driven update, like \cref{eq:diagonal-update}, would make it nonspherical. Therefore $\eta$ evolves only due to the dynamics in our generative model, \cref{eq:predict-appx-dynamics}.

\eat{
\subsection{Reparameterization}
\label{appx:reparam}

We typically initialize the parameters using a diagonal
but non-spherical Gaussian, as discussed in \cref{sec:init},
which is incompatible
with our spherical assumptions.
To overcome this difficulty, we can reparameterize the model
by defining 
$\vtheta_t = \vw_t / \vs_0$ (componentwise),
where $\vw_t$ are the raw network weights.
We can then use  a diagonal Gaussian prior, 
$p(\vtheta_0) = \gauss(\vtheta_0|\vmu_0, \eta_0 \vI_\nparams)$ with $\vmu_0$ sampled from $\gauss(0,\vI_\nparams)$.
Thus we set $\vUpsilon_0 = \eta_0 \vI_\nparams$
and $\vW_0 \leftarrow \left[0\right]^{\nparams \times \memory}$.
When working with the reparameterized model,
Jacobians of the observation model ($\vH_t$) and all related quantities are defined with respect to the transformed parameters $\vtheta$ rather than the raw weights $\vw$.
Note also that the assumed dynamics, 
$p(\vtheta_t|\vtheta_{t-1}) = \gauss(\vtheta_{t}|\gamma_t\vtheta_{t-1},q_t\vI)$, translates to 
$p(\vw_t|\vw_{t-1}) = \gauss(\vw_{t}|\gamma_t\vw_{t-1},q_t\vS_0)$.
That is, the system noise takes the same shape as the prior uncertainty.
}

\subsection{Steady-state assumption}

We find it helpful to make the steady-state assumption that 
$\var{\vtheta_t}=\var{\vtheta_0}$ for all $t$,
which is the same as the ``variance preserving'' OU process used in diffusion probabilistic models \citep{song2020score,ho2020denoising}.
Because $\var{\vtheta_0}=\eta_0^{-1}\vI_\nparams$, 
and because $\var{\vtheta_t}$ and $\eta_t^{-1}$ both evolve according to \cref{eq:predict-appx-dynamics}, 
$\eta_t^{-1} = \gamma_t^2 \eta_{t-1}^{-1} + q_t$, 
we have by induction that $\var{\vtheta_t}=\eta_t^{-1}\vI_\nparams$ for all $t$.
Therefore the steady-state assumption is equivalent to $\eta_t=\eta_0$ and implies the following constraint for all $t$:
\begin{align}
    \gamma_t^{2} + q_t \eta_{t-1} = 1 \label{eq:steady-state}
\end{align}

\subsection{Notation}

We use $\sizeparam{\square}$ and $\sizeout{\square}$ to denote objects whose ``focal'' dimension is grown from $\memory$ to $\nparams$ and $\memoryout$, respectively.
For example, 
$\vU_t$ has size $\nparams\times\memory$ while
 $\sizeparam{\vU}_t$ has size $\nparams\times\nparams$ 
 (see \cref{eq:Ubar-orthogonal,eq:Ubar-extension}),
 and  
  $\vLambda_t$ has size $\memory\times\memory$
  while 
 $\sizeout{\vLambda}_t$ has size $\nparams \times \memoryout$ (with $\memoryout$ nonzero entries; see \cref{eq:SVD-update-step,eq:SVD-update-step-spherical-Lambda-topL}).

\subsection{Predict step for the parameters}
\label{appx:spherical-predict-step}

The predict step for the parameters,
$p(\vtheta_{t}|\data_{1:t-1}) = \gauss(\vtheta_{t-1}|\vmu_{t|t-1},\vSigma_{t|t-1})$,
just requires pushing the previous posterior
through the linear-Gaussian dynamics model in \cref{eq:predict-appx-dynamics}:
\begin{align}
    \vmu_{t|t-1} &= \gamma_t\vmu_{t-1} \\
    \vSigma_{t|t-1} 
    &= \gamma_t^2 \vSigma_{t-1} + q_t\vI_\nparams
    \label{eq:spherical-predict-bayes-inflation-variance-unsolved}
\end{align}

To efficiently compute $\vSigma_{t-1}$,
let $\sizeparam{\vU}_{t-1}$ be an orthonormal matrix extending $\vU_{t-1}$ 
from $\nparams \times \memory$ to $\nparams \times \nparams$,
and let $\sizeparam{\vlambda}_{t-1}\in\real^\nparams$ be a vector extending $\vlambda_{t-1}$ with zeros:
\begin{align}
    \sizeparam{\vU}_{t-1} \sizeparam{\vU}_{t-1}^\trans &= \vI_\nparams \label{eq:Ubar-orthogonal} \\
    \sizeparam{\vU}_{t-1}[:,1{:}\memory] &= \vU_{t-1} \label{eq:Ubar-extension} \\
    \sizeparam{\vlambda}_{t-1}[1{:}\memory] &= \vlambda_{t-1} \\
    \sizeparam{\vlambda}_{t-1}[(\memory+1){:}\nparams] &= \bm{0}
\end{align} 
Then we can diagonalize 
using $\sizeparam{\vU}_{t-1}$:
\begin{align}
    \vSigma_{t-1}
    &= \left(
        \eta_{t-1} \vI_\nparams
        + \vU_{t-1} \vLambda_{t-1}^{2} \vU_{t-1}^{\trans}
    \right)^{-1} \\
    &= \left(
        \sizeparam{\vU}_{t-1}
        \diag \left(
            \eta_{t-1} + \sizeparam{\vlambda}_{t-1}^{2}
        \right)
        \sizeparam{\vU}_{t-1}^{\trans}
    \right)^{-1} \\
    &= \sizeparam{\vU}_{t-1}
    \diag \left(
        \eta_{t-1} + \sizeparam{\vlambda}_{t-1}^{2}
    \right)^{-1}
    \sizeparam{\vU}_{t-1}^{\trans}
\end{align}

Substituting into \cref{eq:spherical-predict-bayes-inflation-variance-unsolved} gives an efficient expression for the precision:
\begin{align}
    \vSigma_{t|t-1}^{-1} 
    &= \left(
        \gamma_t^2
        \sizeparam{\vU}_{t-1}
        \diag\left(
            \eta_{t-1} + \sizeparam{\vlambda}_{t-1}^{2}
        \right)^{-1}
        \sizeparam{\vU}_{t-1}^{\trans} 
        + q_t\vI_\nparams
    \right)^{-1} \\
    &= \sizeparam{\vU}_{t-1}
        \diag\left(
            \frac{\eta_{t-1}+\sizeparam{\vlambda}_{t-1}^{2}}
            {
                \gamma_t^2
                + q_t \eta_{t-1}
                + q_t \sizeparam{\vlambda}_{t-1}^{2}
            }
        \right)
        \sizeparam{\vU}_{t-1}^{\trans} \\
    &= \frac{\eta_{t-1}}{\gamma_t^2 + q_t\eta_{t-1}} \vI_\nparams
    + \vU_{t-1}
        \diag\left(
            \frac{\gamma_t^2 \vlambda_{t-1}^{2}}
            {
                (\gamma_t^2 + q_t\eta_{t-1})
                (\gamma_t^2 + q_t \eta_{t-1} + q_t \vlambda_{t-1}^{2})
            }
        \right)
        \vU_{t-1}^{\trans}
\end{align}
This implies the updates
\begin{align}
    \eta_t &= 
        \frac{\eta_{t-1}}
        {\gamma_t^2 + q_t\eta_{t-1}} \\
    \vlambda_{t|t-1}^2 &= 
        \frac{\gamma_t^2 \vlambda_{t-1}^{2}}
            {
                (\gamma_t^2 + q_t\eta_{t-1})
                (\gamma_t^2 + q_t \eta_{t-1} + q_t \vlambda_{t-1}^{2})
            } \\
    \vU_{t|t-1} &= \vU_{t-1}
\end{align}
Under the steady-state assumption, \cref{eq:steady-state}, these reduce to
\begin{align}
    \eta_t &= \eta_{t-1} \\
    \vlambda_{t|t-1}^2 &= 
        \frac{\gamma_t^2\vlambda_{t-1}^{2}}
        {1 + q_t\vlambda_{t-1}^{2}} \\
    \vU_{t|t-1} &= \vU_{t-1}
\end{align}
See \cref{algo:LOFI-spherical-predict} for the pseudocode.


\begin{algorithm}
def $\text{predict}(
\vmu_{t-1}, \vlambda_{t-1}, \vU_{t-1}, \eta_{t-1},
\vx_t,  \gamma_t, q_t)$: \\
$\vmu_{t|t-1} = \gamma \vmu_{t-1}$
\\
$\vlambda_{t|t-1} = 
    \sqrt{
        \frac{\gamma_t^{2} \vlambda_{t-1}^{2}}
        {
            (\gamma_t^2 + q_t \eta_{t-1})
            (\gamma_t^2 + q_t \eta_{t-1} + q_t \vlambda_{t-1}^2)
        }
    }$ 
// componentwise
\\
$\vU_{t|t-1} = \vU_{t-1}$
\\
$\eta_{t} = \frac{\eta_{t-1}}
{\gamma_t^2 + q_t \eta_{t-1}}$
\\
$\hat{\vy}_t = 
h(\vx_t, \vmu_{t|t-1} )$ \\
Return $(\hat{\vy}_t, \vmu_{t|t-1}, \vlambda_{t|t-1}, \vU_{t|t-1}, \eta_{t})$ 
\caption{\lofi predict step (spherical).}
\label{algo:LOFI-spherical-predict}
\end{algorithm}

\subsection{Update step}
\label{appx:update-step-spherical}

\Cref{algo:LOFI-spherical-update} shows the pseudocode for spherical \lofi's update step.
The mean update is the same as for diagonal \lofi, \cref{eq:mean-update}. Substituting the spherical part of the precision, $\vUpsilon_{t|t-1} = \eta_t \vI_\nparams$, yields
\begin{equation}
    \vmu_{t} =
    \vmu_{t|t-1}
    + \eta_t^{-1} \left(
        \vI_\nparams
        - \sizeout{\vW}_{t}
        \left(
             \eta_t \vI_{\memoryout}
            + \sizeout{\vW}_{t}^{\trans} \sizeout{\vW}_{t}
        \right)^{-1}
        \sizeout{\vW}_{t}^{\trans}
    \right)
    \vH_{t}^{\trans} \vR_{t}^{-1} \ve_{t}
\end{equation}

\subsubsection{Precision update: SVD version}
\label{appx:full-svd}

Our primary proposed update step for spherical \lofi is essentially the same as that for diagonal \lofi. We define $\tilde{\vW}_t$ as in \cref{eqn:Wtilde}, calculate its SVD as in \cref{eq:SVD-update-step}, and keep the top $\memory$ singular values and vectors (mirroring \cref{eq:SVD-update-step-topL}):
\begin{align}
    \vlambda_{t|t-1} &= \tilde{\vlambda}_{t|t-1}[1{:}\memory] 
        \label{eq:SVD-update-step-spherical-Lambda-topL}\\
    \vU_{t|t-1} &= \tilde{\vU}_{t|t-1}[:,1{:}\memory] 
\end{align}
To keep the diagonal part of the precision spherical, we do not update it in response to data (cf. \cref{eq:diagonal-update}).

\eat{
The SPL approximation allows a more efficient calculation of the exact posterior mean than \cref{eq:mean-update}, which we use for diagonal \lofi. Using \cref{eq:posterior-precision-exact,eq:SVD-update-step} and the diagonalization trick from \cref{appx:spherical-predict-step}, we can write $\vSigma_t^*$ as
\begin{align}
    \vSigma_t^*
    &= \left( \eta_t \vI_\nparams + \sizeout{\vU}_{t} \sizeout{\vLambda}_{t}^2 \sizeout{\vU}_{t}^\trans \right)^{-1} \\
    &= \sizeparam{\vU}_t 
    \diag \left(
        \eta_t + \sizeparam{\vlambda}_t^2
    \right)^{-1}
    \sizeparam{\vU}_t^\trans \\
    &= \eta_{t}^{-1} \vI_\nparams
    - \sizeout{\vU}_{t} 
    \underbrace {\diag \left(
        \frac{\sizeout{\vlambda}^2_{t}}
        {\eta_{t}^2 + \eta_{t} \sizeout{\vlambda}^2_{t}}
    \right)}_{\vD_{t}}
    \sizeout{\vU}_{t}^\trans 
\label{eq:posterior-variance-explicit}
\end{align}
where here $\overline{\lambda}_t[j] = \sizeout{\lambda}_t[j]$ for $j\le\memoryout$ and is zero otherwise. 
We can hence compute the mean by substituting this expression in \cref{eq:kalman-gain,eq:innovation,eq:EKF-mu-update}:
\begin{align}
    \vmu_t &= \vmu_{t|t-1} + \vK_t \ve_t \\
    \vK_t &= \left(
        \eta_{t}^{-1} \vI_\nparams
        - \sizeout{\vU}_{t} \vD_{t} \sizeout{\vU}_{t}^\trans
    \right) 
    \vH_t^\trans \vR_t^{-1} 
    \label{eqn:kalmanGainSVD}
  \end{align}
Computing $\vK_t$ takes
$O(\nparams \memory^2 + \nparams \nout^2 
+ \nparams \memory \nout)$ time, and multiplying
$\vK_t \ve_t$ takes $O(\nparams \nout)$ time.
}

\begin{algorithm}
def $\text{update}(
\vmu_{t|t-1}, \vlambda_{t|t-1}, \vU_{t|t-1}, \eta_{t},
\vx_t, \vy_t, \hat{\vy}_t, \Var[\vy |\cdot], \memory)$: 
\\
$\ve_t = \vy_t - \hat{\vy}_t$ \\
$\vR_t = \Var[\vy \vert \hat{\vy_t}]$\\
$\vA_t^\trans = \chol(\vR_t^{-1})$ \\
$\tilde{\vW}_{t} = \left[\begin{array}{cc}
\vU_{t | t-1} \diag(\vlambda_{t|t-1}) 
& \vH_{t}^{\trans} \vA_{t}^{\trans}
\end{array}\right]$
\\
$\vmu_t = 
    \vmu_{t|t-1} 
    + \eta_{t}^{-1} \left(
        \vI_{\nparams}  
        - \tilde{\vW}_t
        \left(
            \eta_{t} \vI_{\memoryout} +  \tilde{\vW}_t^\trans \tilde{\vW}_t
        \right)^{-1}
        \tilde{\vW}_t^\trans
    \right)
    \vH_t^\trans \vR_t^{-1} 
    \ve_t$
\\
\eIf{Full-SVD}
{
$(\tilde{\vlambda}_t, \tilde{\vU}_t) = \text{SVD}(\tilde{\vW}_t)$
\\
$(\vlambda_t, \vU_t) = \text{top-\memory}(\tilde{\vU}_t, \tilde{\vlambda}_t)$ 
}
{
$(\vlambda_t, \vU_t) = \text{SVD-orth}(
    \vlambda_{t|t-1}, \vU_{t|t-1}, \vH_t, \vA_t)$ \\
}
Return $(\vmu_t, \vlambda_t, \vU_t)$ 
\caption{\lofi update step (spherical).}
\label{algo:LOFI-spherical-update}
\end{algorithm}

\subsubsection{Precision update: Orthogonal projection version}
\label{appx:orth-svd}

Computing the SVD takes $O(\nparams \memoryout^2)$ time,
which may be expensive. We now present an alternative
that takes $O(\nparams \memory \nout)$ time, but which is less accurate.
The approach is
based on the \orfit method \citep{ORFit},
which uses orthogonal projections to make the SVD fast to compute.
\eat{The overall running time is reduced to $O(\nparams \memoryout)$.}

To explain the method, we start by considering the special case
of a linearized scalar output model of the form
\begin{align}
\gauss(y_t | 
    h(\vx_t,\vmu_{t|t-1}) + \vg_t^\trans (\vtheta_t-\vmu_{t|t-1}),
    \obsVar)
\end{align}
where $\vg_t = \nabla_{\vtheta} h(\vx_t,\vtheta)_{\vmu_{t|t-1}} = \vH_t^\trans$
is the gradient.
So $\tilde{\vW}_{t}$ becomes a $\nparams \times (\memory+1)$ matrix,
given by
$\tilde{\vW}_{t}	=\left[\begin{array}{cc} 
\vU_{t-1} \vLambda_{t-1} & \vg_{t}\end{array}\right]$.
There is no closed-form method for computing the SVD of this new matrix, because the new gradient will generally be oblique to the existing vectors. 
The \orfit method \citep{ORFit}
makes the problem tractable by replacing the gradient
$\vg_t$ by its projection
onto the subspace orthogonal to the current basis set. That is, it replaces $\vg_t$ with 
\begin{align}
    \vv_{t}	&=\left(\vI_{\nparams} - 
    \vU_{t|t-1} \vU_{t|t-1}^{\trans}\right) \vg_t
\end{align}
Computing the SVD of 
$\ring{\vW}_{t}	=\left[\begin{array}{cc} 
\vU_{t|t-1} \vLambda_{t|t-1} & \vv_{t} \end{array}\right]$ is trivial because its columns are orthogonal.
First let $\vlambda_t = \vlambda_{t|t-1}$ and
$\vU_t = \vU_{t|t-1}$.
Now compute
$v = ||\vv_t||$ and let $k = \argmin_{j'} \lambda_{t-1}[j']$.
If $v > \lambda_{t}[k]$,
then we replace $\lambda_t[k]$ with $v$,
and $\vU_t[:,k]$ with $\vv_t/v$.
That is, we discard an old basis vector
if the new observation is more informative,
in the sense of Fisher information with respect to the linearized observation model.

We can generalize to handle $\nout$-dimensional outputs,
to efficiently compute a truncated rank-$\memory$
SVD of 
$\tilde{\vW}_t$ in \cref{eqn:Wtilde},
by incrementally applying the above procedure to each column
of the generalized matrix of gradients,
$\vH_{t}^{\trans} \vA_{t}^{\trans}$.
To reduce the dependence on the order of projection,
we visit the columns in a random order.
We denote this operation by
\begin{align}
    (\vU_t, \vLambda_t) = \text{SVD-orth}(
    \vU_{t|t-1}, \vLambda_{t|t-1}, \vH_t, \vA_t, \memory).
\end{align}
See \cref{algo:SVDorth} for the pseudocode.
This takes $O(\nparams \memory \nout)$ time.

\begin{algorithm}
def $\text{SVD-orth}(\vlambda, \vU, \vH, \vA)$: \\
Sample $\vpi \in \text{perm}(\nout)$ 
\\
 \For{$j \in \vpi$}{
$\vv_j = \left(\vI_{\nparams}-\vU \vU^{\trans}\right) \vH^{\trans} \left[\vA^{\trans}\right]_{\cdot j}$ \\
$v_j = \Vert \vv_j \Vert$ \\
$k = \argmin \vlambda$ \\
\If{$v_j > \lambda_k$}
{
 $\vU[:,k] = \frac{\vv_j}{v_j}$ \\
  $\lambda_k = v_j$
}
}
Return $(\vlambda, \vU)$
\caption{Incremental SVD using orthogonal projection.}
\label{algo:SVDorth}
\end{algorithm}

\subsection{Inflation}
\label{appx:spherical-inflation}

Inflation operates identically in spherical and diagonal \lofi, up to a change in notation. Because spherical \lofi represents the low-rank part of the precision as $\vU_t \vLambda_t$ instead of $\vW_t$, the update to $\vW_{t-1}$ (rescaling by $1/\sqrt{1+\alpha}$ as in \cref{eq:inflation-bayes-lowrank-update,eq:inflation-simple-lowrank-update,eq:inflation-hybrid-lowrank-update}) becomes a rescaling of $\vLambda_{t-1}$, with $\vU_{t-1}$ unchanged. Likewise, because spherical \lofi represents the diagonal part of the precision as $\eta_t \vI_\nparams$ instead of $\vUpsilon_t$, the update to $\vUpsilon_{t-1}$ becomes an update to $\eta_{t-1}$. This update simplifies to $\discount{\eta}_{t-1} = \eta_{t-1}$ for Bayesian and hybrid inflation (see \cref{eq:inflation-bayes-diagonal-update,eq:inflation-hybrid-diagonal-update} with $\vUpsilon_{t-1}=\eta_{t-1}\vI_\nparams$). This simplification arises because, in spherical \lofi, the latent predictive prior exactly coincides with the spherical part of the precision; therefore discounting the likelihood and not the prior amounts to deflating $\vLambda_{t-1}$ and leaving $\eta_{t-1}$ unchanged. Under simple inflation, $\vLambda_{t-1}$ and $\eta_{t-1}$ are both deflated. To implement inflation, the parameters computed here ($\discount{\vmu}_{t-1},\discount{\eta}_{t-1},\discount{\vLambda}_{t-1}$) are substituted for the posterior parameters (${\vmu}_{t-1},{\eta}_{t-1},{\vLambda}_{t-1}$) in the predict step (\cref{appx:spherical-predict-step}).

Bayesian inflation:
\begin{align}
    \discount{\vmu}_{t-1} &=
        \vmu_{t-1}
        + \frac{\alpha \eta_{t-1}}{1+\alpha} 
        \left(
            \discount{\eta}_{t-1} \vI_\nparams
            + \discount{\vU}_{t-1} \discount{\vLambda}_{t-1}^2 \discount{\vU}_{t-1}^\trans
        \right)^{-1}
        (\Gamma_{t-1} \vmu_{0}-\vmu_{t-1}) \\
    \discount{\eta}_{t-1} &= \eta_{t-1} \\
    \discount{\vLambda}_{t-1} &=
        \frac{1}{\sqrt{1+\alpha}} \vLambda_{t-1}
\end{align}

Simple inflation:
\begin{align}
    \discount{\vmu}_{t-1} &= \vmu_{t-1} \\
    \discount{\eta}_{t-1} &= \frac{1}{1+\alpha} \veta_{t-1} \\
    \discount{\vLambda}_{t-1} &= \frac{1}{\sqrt{1+\alpha}} \vLambda_{t-1}
\end{align}

Hybrid inflation:
\begin{align}
    \discount{\vmu}_{t-1} &= \vmu_{t-1} \\
    \discount{\eta}_{t-1} &= \eta_{t-1} \\
    \discount{\vLambda}_{t-1} &= \frac{1}{\sqrt{1+\alpha}} \vLambda_{t-1}
\end{align}

In all three cases, $\discount{\vU}_{t-1} = \vU_{t-1}$.

\end{document}